\documentclass[11pt,a4paper]{article}
\PassOptionsToPackage{table}{xcolor}
\usepackage[left=46pt,right=46pt,top=70pt,bottom=58pt]{geometry}
\usepackage[T1]{fontenc}
\usepackage[utf8]{inputenc}
\usepackage{mathpazo}
\usepackage{natbib}
\usepackage{amsmath,amssymb,amsfonts,amsthm}
\usepackage{booktabs,makecell,multirow,graphicx,caption,subfigure,wrapfig}
\usepackage{enumitem}
\usepackage[normalem]{ulem}
\usepackage{tcolorbox}
\tcbuselibrary{skins,breakable}
\usepackage[framemethod=tikz]{mdframed}
\usepackage{nicefrac,microtype,float,titletoc,titlesec}
\usepackage{needspace}
\usepackage{hyperref,url}
\usepackage[capitalize,nameinlink]{cleveref}

\usepackage{amsmath,amsfonts,bm}

\def\eqref#1{equation~\ref{#1}}

\def\1{\bm{1}}

\DeclareMathAlphabet{\mathsfit}{\encodingdefault}{\sfdefault}{m}{sl}
\SetMathAlphabet{\mathsfit}{bold}{\encodingdefault}{\sfdefault}{bx}{n}

\definecolor{NRCBrown}{RGB}{112,83,39}
\definecolor{NRCCream}{RGB}{250,246,236}
\definecolor{NRCText}{RGB}{41,37,31}
\color{NRCText}
\hypersetup{colorlinks=true,linkcolor=NRCBrown,citecolor=NRCBrown,urlcolor=NRCBrown,linktoc=all}
\titleformat{\section}{\color{NRCBrown}\normalfont\Large\bfseries}{}{0pt}{}
\titleformat{\subsection}{\color{NRCBrown}\normalfont\large\bfseries}{}{0pt}{}
\titleformat{\subsubsection}{\color{NRCBrown}\normalfont\normalsize\bfseries}{}{0pt}{}
\titlespacing*{\section}{0pt}{17pt plus 3pt minus 2pt}{5pt}
\titlespacing*{\subsection}{0pt}{12pt plus 2pt minus 1pt}{4pt}
\titlespacing*{\subsubsection}{0pt}{9pt plus 2pt minus 1pt}{3pt}
\renewenvironment{abstract}{%
  \begin{tcolorbox}[enhanced,breakable,colback=NRCCream,colframe=NRCCream,
    boxrule=0pt,arc=6pt,left=13pt,right=13pt,top=8pt,bottom=10pt,
    before skip=11pt,after skip=15pt]
  \begin{center}\color{NRCBrown}\large\bfseries Abstract\end{center}\smallskip
}{\end{tcolorbox}}
\newcommand{\NRCemail}[1]{\href{mailto:#1}{\textcolor{NRCBrown}{#1}}}
\newcommand{\NRCaffiliations}{%
  \begin{tcolorbox}[enhanced,colback=white,colframe=NRCBrown!65,
    boxrule=.45pt,arc=5pt,left=9pt,right=9pt,top=8pt,bottom=8pt,
    before skip=10pt,after skip=2pt]
  \small
  \setlength{\parskip}{2pt}
  \textsuperscript{1} State Key Laboratory of AI Safety, Institute of Computing Technology,
  Chinese Academy of Sciences.\par
  \textsuperscript{2} School of Computer Science and Technology, University of Chinese
  Academy of Sciences.\par
  \textsuperscript{3} Beijing Academy of Artificial Intelligence.\par
  \textsuperscript{4} Artificial Intelligence Institute of China Electronics Technology
  Group Corporation.\par
  \smallskip
  \textit{*Corresponding authors:} \NRCemail{xuqianqian@ict.ac.cn};
  \NRCemail{qmhuang@ucas.ac.cn}.\par
  \textit{Other emails:} \NRCemail{huacong23z@ict.ac.cn};
  \NRCemail{wangzitai@ict.ac.cn};\\
  \NRCemail{yangzhiyong21@ucas.ac.cn};
  \NRCemail{baoshilong@ict.ac.cn};
  \NRCemail{zhanglinchao@cetc.com.cn}.
  \end{tcolorbox}%
}
\makeatletter
\renewcommand{\maketitle}{%
  \begingroup
  \noindent{\color{NRCBrown}\LARGE\bfseries \@title\par}
  \vspace{7pt}
  \noindent{\normalsize\@author\par}
  \NRCaffiliations
  \endgroup
}
\makeatother
\AtBeginDocument{%
  \setlength{\abovedisplayskip}{7pt plus 2pt minus 3pt}%
  \setlength{\belowdisplayskip}{7pt plus 2pt minus 3pt}%
  \setlength{\abovedisplayshortskip}{2pt plus 1pt}%
  \setlength{\belowdisplayshortskip}{4pt plus 1pt minus 2pt}%
  \setlength{\jot}{3pt}%
}

\newtheorem{theorem}{Theorem}[section]
\newtheorem{proposition}[theorem]{Proposition}

\theoremstyle{definition}

\newtheorem{remark}{Remark}[section]
\definecolor{TheoremBack}{RGB}{235,244,250}
\definecolor{TheoremFrame}{RGB}{142,181,207}
\newmdenv[backgroundcolor=TheoremBack,linecolor=TheoremFrame,
  linewidth=2.2pt,topline=false,rightline=false,bottomline=false,
  innerleftmargin=5pt,innerrightmargin=4pt,
  innertopmargin=4pt,innerbottommargin=4pt,
  skipabove=6pt,skipbelow=6pt]{propositionbox}
\definecolor{RemarkBack}{RGB}{247,247,247}
\definecolor{RemarkFrame}{RGB}{178,178,178}
\definecolor{RemarkText}{RGB}{92,92,92}
\newmdenv[backgroundcolor=RemarkBack,linecolor=RemarkFrame,
  fontcolor=RemarkText,linewidth=2.2pt,
  topline=false,rightline=false,bottomline=false,
  innerleftmargin=5pt,innerrightmargin=4pt,
  innertopmargin=4pt,innerbottommargin=4pt,
  skipabove=6pt,skipbelow=6pt]{remarkbox}

\crefname{equation}{Eq.}{Eqs.}
\Crefname{equation}{Eq.}{Eqs.}
\crefname{section}{Sec.}{Secs.}
\Crefname{section}{Sec.}{Secs.}
\crefname{subsection}{Sec.}{Secs.}
\Crefname{subsection}{Sec.}{Secs.}
\crefname{subsubsection}{Sec.}{Secs.}
\Crefname{subsubsection}{Sec.}{Secs.}
\crefname{appendix}{App.}{Apps.}
\Crefname{appendix}{App.}{Apps.}
\crefname{subappendix}{App.}{Apps.}
\Crefname{subappendix}{App.}{Apps.}
\crefname{subsubappendix}{App.}{Apps.}
\Crefname{subsubappendix}{App.}{Apps.}
\crefname{table}{Tab.}{Tabs.}
\Crefname{table}{Tab.}{Tabs.}
\crefname{figure}{Fig.}{Figs.}
\Crefname{figure}{Fig.}{Figs.}
\crefname{proposition}{Prop.}{Props.}
\Crefname{proposition}{Prop.}{Props.}
\crefname{theorem}{Thm.}{Thms.}
\Crefname{theorem}{Thm.}{Thms.}
\crefname{corollary}{Cor.}{Cors.}
\Crefname{corollary}{Cor.}{Cors.}
\crefname{lemma}{Lem.}{Lems.}
\Crefname{lemma}{Lem.}{Lems.}
\crefname{definition}{Def.}{Defs.}
\Crefname{definition}{Def.}{Defs.}
\crefname{remark}{Remark}{Remarks}
\Crefname{remark}{Remark}{Remarks}

\title{Compare to Correct: Pairwise-Guided Logit\\Adjustment for LLM Test-Time Scaling}
\author{Cong Hua\textsuperscript{1,2}\quad Qianqian Xu\textsuperscript{1,3,*}\quad
  Zitai Wang\textsuperscript{1}\quad Zhiyong Yang\textsuperscript{2}\\
  Shilong Bao\textsuperscript{1}\quad Linchao Zhang\textsuperscript{4}\quad
  Qingming Huang\textsuperscript{1,2,*}}
\hypersetup{pdftitle={Compare to Correct: Pairwise-Guided Logit Adjustment for LLM Test-Time Scaling},
  pdfauthor={Cong Hua, Qianqian Xu, Zitai Wang, Zhiyong Yang, Shilong Bao, Linchao Zhang, Qingming Huang}}

\begin{document}
\maketitle
\begin{abstract}

Sampling-based test-time scaling enhances LLM reasoning by aggregating multiple reasoning paths for the same query. Existing methods use answer frequencies or path confidence to compute candidate scores, which do not explicitly  characterize comparisons between competing answers. Inspired by the Condorcet principle in social choice theory, we introduce pairwise confidence comparisons between paths supporting different answers to capture answer-level comparative advantage. We combine this comparative evidence with voting support through a KL-regularized objective. Theoretically, we show that this comparative-advantage is equivalent to the adjustment to the traditional voting logits. In this way, our proposed \textit{Pairwise-Guided Logit Adjustment} ($\mathsf{PGLA}$)  framework allows comparative evidence to compensate for weaker voting support and correct voting-based ranking errors.
Experiments on \textbf{$10$} challenging benchmarks including mathematical reasoning, knowledge reasoning, and code generation with \textbf{$7$} models ranging from \textbf{1.7B} to \textbf{32B} parameters speak to the effectiveness of this plug-and-play term. Additional evaluations with external reward scores and online sampling further demonstrate its applicability.
\end{abstract}

\section{Introduction}
\label{sec:intro}

Large language models (LLMs) have demonstrated strong reasoning capabilities across mathematical, symbolic, and multi-step decision-making tasks~\citep{wei2022chainofthought,lewkowycz2022solving,yao2023tree}. Beyond increasing model size, allocating additional computation at inference time can further improve reasoning performance, establishing \emph{test-time scaling} as an important paradigm for LLM reasoning~\citep{snell2024scaling,balachandran2025inference}. Among its variants, \emph{sampling-based} approaches generate multiple reasoning paths for the same query and aggregate evidence from these paths (e.g., answer frequencies and path confidence scores) into a collective decision~\citep{cobbe2021training,wang2022self}. In this direction, the central challenge becomes \textbf{how to aggregate path-level evidence into reliable answer-level scores.} Representative approaches include \emph{best-of-$N$ selection} that chooses the answer from the path with the highest confidence score~\citep{cobbe2021training}, \emph{majority voting} based on answer frequencies~\citep{wang2022self}, and \emph{confidence-based voting} that combines answer frequencies with path confidence scores by weighting or filtering path contributions~\citep{kang2025selfcertainty,fu2025deepconf}.

\begin{figure*}[t]
    \centering
    \includegraphics[width=0.9\textwidth]{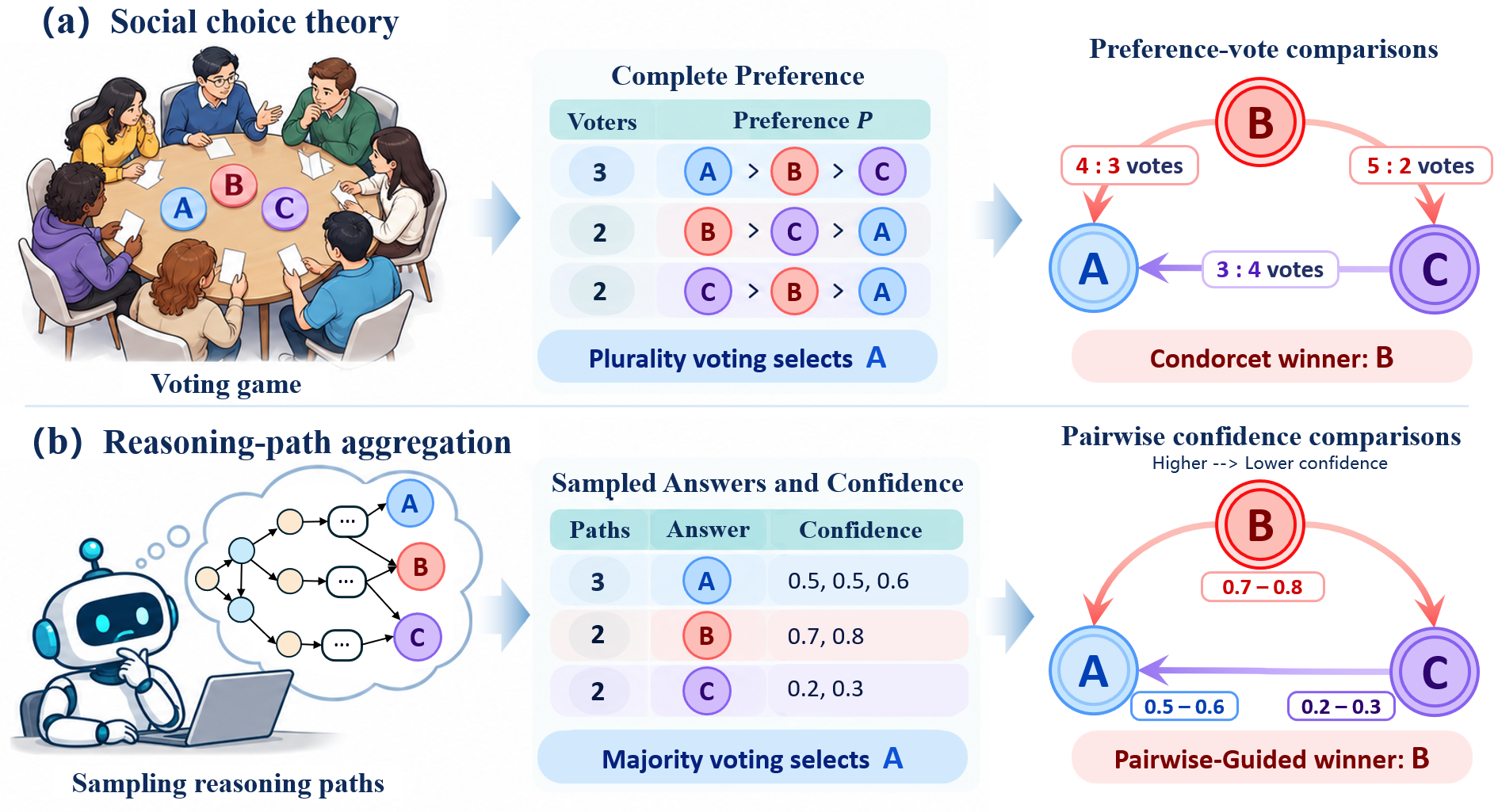}
    \caption{From voting support to pairwise comparative evidence.
    (a) Plurality voting selects $A$, whereas pairwise preference
    comparisons identify $B$ as the Condorcet winner.
    (b) Sampled reasoning paths provide answers and confidence scores.
    Pairwise confidence comparisons complement voting support,
    enabling the proposed aggregation method to select $B$.}
    \label{fig:pipeline_society}
\end{figure*}

Although these methods achieve substantial improvements, we observe that their aggregation rules could overlook some crucial information hidden among sampled paths. To illustrate this, we turn to \emph{social choice theory}, which studies how individual preferences among available options can be combined into a collective decision~\citep{zwicker2016voting}. The well-known \emph{Condorcet principle} highlights that the most frequently selected option does not necessarily rank higher than competing options in direct comparisons~\citep{young1988condorcet,fishburn1977condorcet}. As shown in Figure~\ref{fig:pipeline_society}(a), options $A$, $B$, and $C$ receive $3$, $2$, and $2$ first-choice selections, respectively, so majority voting selects $A$. Under the preference $P$, however, four of seven rankings place $B$ above $A$, and five place $B$ above $C$, so $B$ is selected through pairwise comparisons. Similarly, this principle highlights why prior sampling-based approaches are insufficient: an answer can be assigned the highest score even when its reasoning paths remain at a disadvantage in comparisons with those supporting a competing answer. As shown in Figure~\ref{fig:pipeline_society}(b), answer $A$ is selected by majority voting because it appears most frequently, although every path producing answer $B$ has a higher confidence score than those producing $A$. This motivates us to \textbf{incorporate pairwise information between candidate answers on top of existing voting-based aggregation.}

Ideally, we could apply the Condorcet principle to the reasoning paths if the ranking of candidate answers were available for each path. However, in practice, we only observe each path's first-choice answer and its confidence score, leaving the ordering among the remaining answers unknown. To estimate such missing ordering, we use \textbf{\emph{pairwise confidence comparisons}} between paths producing different answers, as illustrated in Figure~\ref{fig:pipeline_society}(b). These comparisons yield an answer-level comparative advantage, which we combine with voting evidence through a KL-regularized objective. In this way, the objective can balance this advantage against deviations from the base voting distribution, as an alternative to the naive Condorcet approach. Furthermore, we show that its solution is essientially a \emph{logit adjustment} to the base voting distribution, yielding a simple yet effective \emph{Pairwise-Guided Logit Adjustment} ($\mathsf{PGLA}$) approach. This plug-and-play adjustment allows comparative evidence to compensate for weaker evidence from existing voting-based aggregation.

Finally, we conduct extensive experiments on mathematical reasoning, knowledge reasoning, and code generation using both reasoning and non-reasoning models from the \textbf{Qwen}, \textbf{DeepSeek}, and \textbf{InternLM} families, with model sizes ranging from \textbf{1.7B} to \textbf{32B} parameters. With internal confidence scores, $\mathsf{PGLA}$ achieves an average accuracy gain of approximately $2\%$ over the corresponding voting baselines across $33$ model–benchmark settings. Additional experiments show gains with external reward scores. On six mathematical reasoning settings, its online extension also achieves higher mean accuracy with fewer sampled paths than Online SC under the same stopping rule.


\vskip -3pt\relax
\vskip 0pt\relax
\section{Problem Formulation and Representative Aggregators}
\vskip -2pt\relax
\label{sec:preliminaries}

\subsection{Problem Formulation}
\label{sec:problem_formulation}
\vskip -1.5pt\relax

Given an input query, a language model generates $K$ reasoning paths
$z_1,\ldots,z_K$.
Each path produces an answer $a_i=g(z_i)$ and receives a confidence score
$p_i$ from the model or an external verifier.
Together, they form the path-level evidence pool
$\mathcal D=\{(a_i,p_i)\}_{i=1}^{K}$. Let $\mathcal A$ be the set of different answers produced by these paths, which may be predefined for multiple-choice questions or dynamically formed from sampled answers in open-ended mathematical reasoning and code generation.

Based on $\mathcal D$, we construct a \emph{path-answer matrix} $\mathcal{P}$,
with one row per path and one column per candidate answer.
Each row records $p_i$ in the column for its output answer $a_i$ and 
the entries for other answers remain unobserved.
Grouping paths by their output answers produces the block-diagonal pattern
shown in \Cref{fig:matrix}.
Each block contains the confidence scores of paths sharing the same answer.

\newlength{\motivationPairedFigureHeight}
\begin{figure*}[t]
    \centering
    \setlength{\motivationPairedFigureHeight}{0.37\textwidth}
    \begin{minipage}[t]{0.5\textwidth}
        \vspace{0pt}
        \centering
        \includegraphics[height=\motivationPairedFigureHeight]{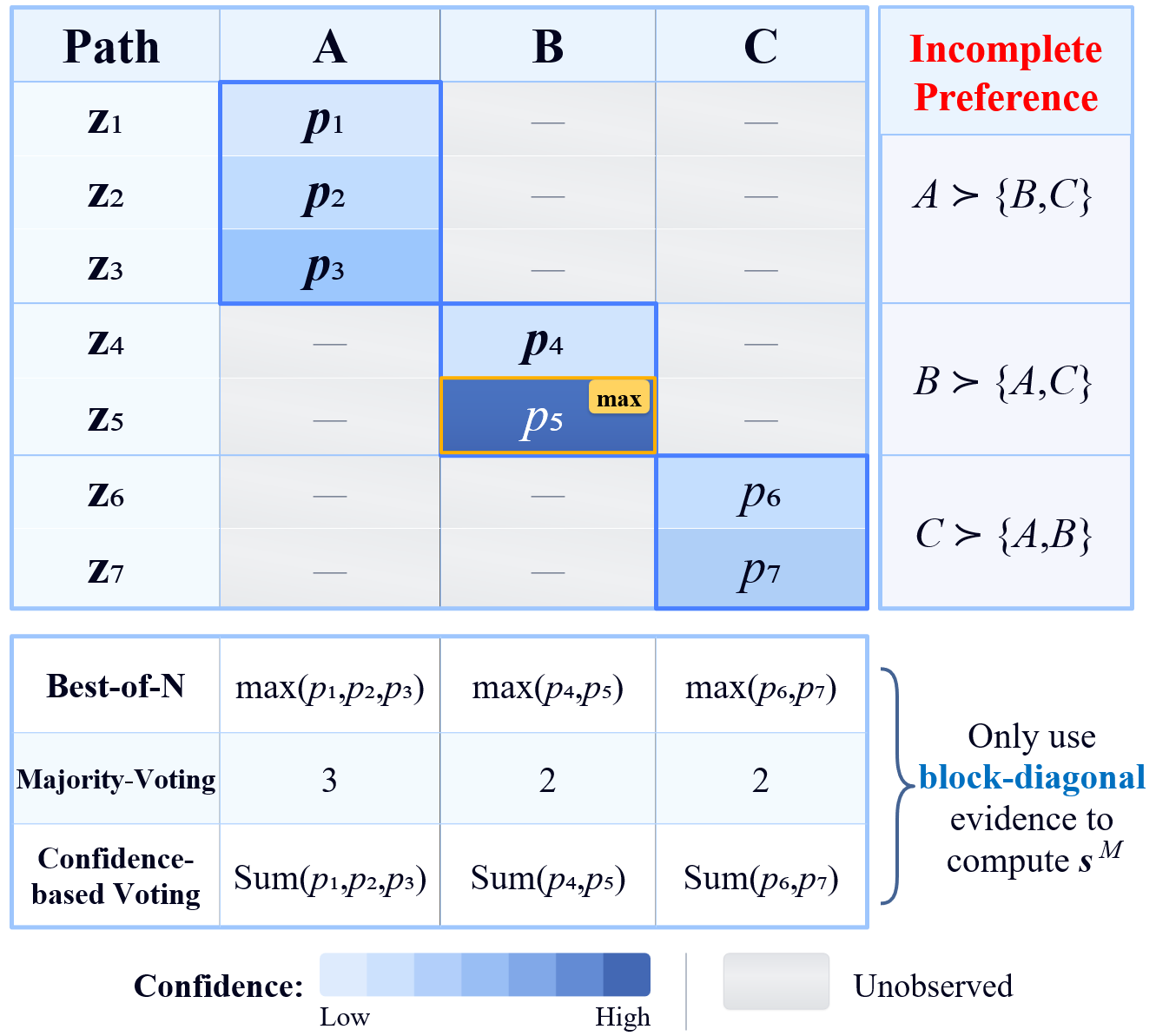}
        \caption{Path-answer evidence and its aggregation. The three methods use each block's maximum confidence, size, or confidence-based contributions.}
        \label{fig:matrix}
    \end{minipage}%
    \hfill
    \begin{minipage}[t]{0.45\textwidth}
        \vspace{0pt}
        \centering
        \includegraphics[width=\linewidth,height=\motivationPairedFigureHeight,keepaspectratio]{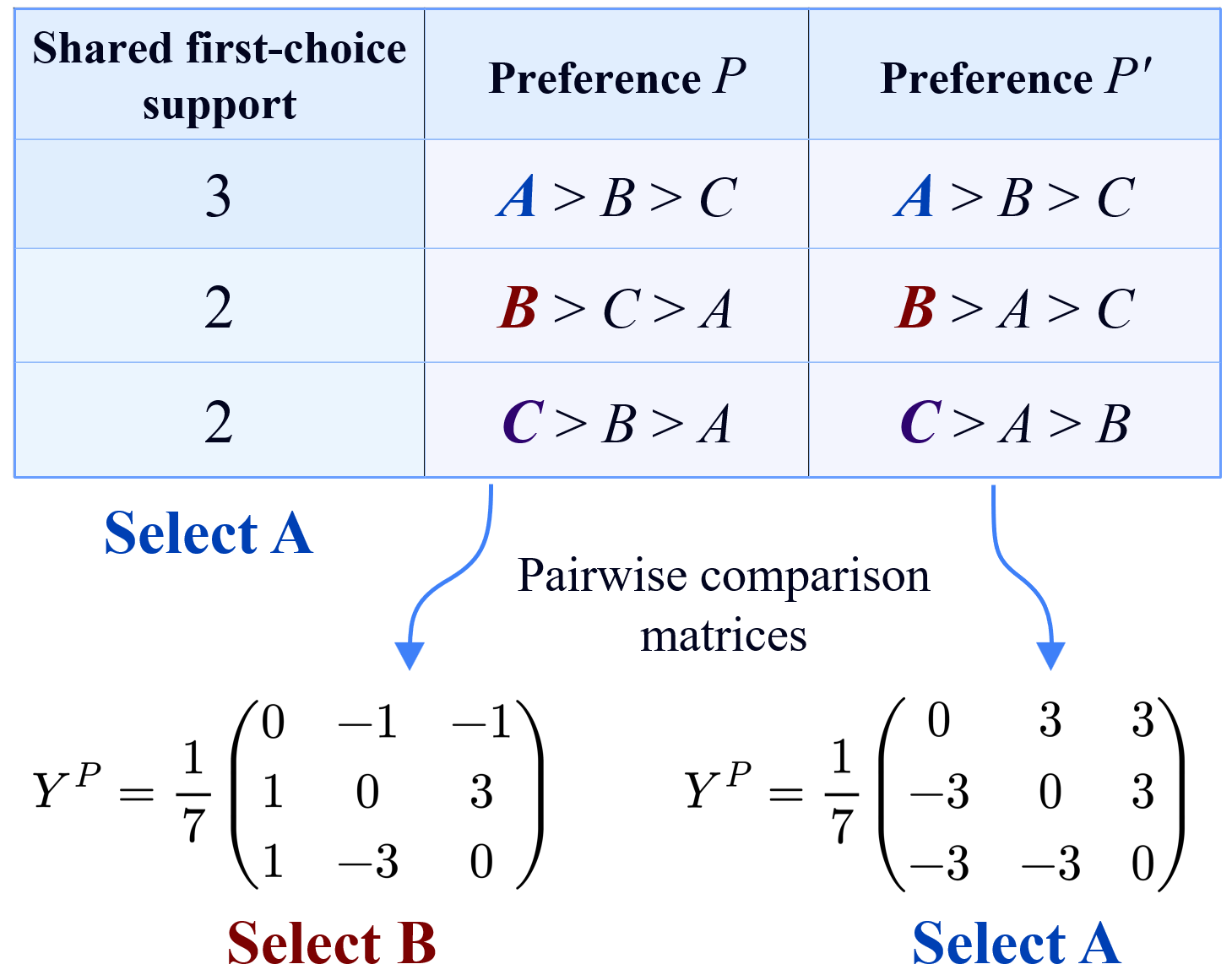}
        \caption{Identical majority-voting scores can correspond to different pairwise comparison matrices and different winners.}
        \label{fig:motivation_theory}
    \end{minipage}
\end{figure*}

The aggregation task is to \textbf{convert the path-level evidence into an answer-level
score for each candidate, followed by selecting the candidate with the highest score}.
Formally, an aggregation method $\mathcal M$ assigns candidate $a$ the score
$s_a^{\mathcal M}$ based on $\mathcal D$:
\begin{equation}
    s_a^{\mathcal M}
    =
    \mathcal M(a;\mathcal D),
    \qquad
    \hat a
    \in
    \arg\max_{a\in\mathcal A}
    s_a^{\mathcal M}.
    \label{eq:answer_level_aggregation}
\end{equation}
$\hat a$ is the final predicted answer selected from the candidates with the highest score. Representative methods use different information from the path-answer matrix to compute $s_a^{\mathcal M}$.

\noindent\textbf{Best-of-$N$ selection.}
Best-of-$N$ uses the highest confidence among the paths sharing
answer $a$~\citep{cobbe2021training}.
This corresponds to taking the largest value
within the block for $a$ in matrix $\mathcal{P}$:
\begin{equation}
    s_a^{\mathrm{BoN}}
    =
    \max_{i:a_i=a}p_i.
    \label{eq:best_of_n_score}
\end{equation}
The confidence $p_i$ may be perplexity~\citep{DBLP:journals/corr/abs-2001-09977}, self-certainty~~\citep{kang2025selfcertainty}, DeepConf confidence~\citep{fu2025deepconf}, or an external verifier's reward~\citep{lightman2023verify}. 

\noindent\textbf{Majority voting.}
Self-consistency (SC) treats each path as one vote for its output answer
and selects the most frequently generated answer~\citep{wang2022self}.
\par\Needspace{4\baselineskip}\noindent
The answer-level score is therefore its total vote count:
\begin{equation}
    s_a^{\mathrm{SC}}
    =
    \sum_{i=1}^{K}\mathbf{1}(a_i=a).
    \label{eq:sc_score}
\end{equation}
where $\mathbf{1}(\cdot)$ is the indicator function. In $\mathcal{P}$, this corresponds to the size of the diagonal block for answer $a$, measured by its number of observed entries. Every supporting path contributes equally, regardless of its confidence.

\noindent\textbf{Confidence-based voting.}
Confidence-based voting methods adjust path contributions through
weighting or filtering~\citep{kang2025selfcertainty,fu2025deepconf}.
Let $h_i\geq0$ be the weighting or filtering coefficient for path $i$,
with $h_i=0$ for filtered paths.
The resulting answer-level score is
\begin{equation}
    s_a^{\mathrm{Conf\text{-}Vote}}
    =
    \sum_{i=1}^{K}
    h_i\cdot p_i\,\mathbf{1}(a_i=a).
    \label{eq:confidence_voting_score}
\end{equation}
This method aggregates confidence-based contributions within each diagonal block in $\mathcal{P}$.

Despite these differences, \textbf{these methods aggregate only the observed
evidence in the diagonal blocks of $\mathcal{P}$}. They do not explicitly model the unobserved preferences among alternative answers. In the next section, we use social choice theory to explain why these preferences matter for answer aggregation.

\vskip -2pt\relax
\vskip 0pt\relax
\subsection{Motivation: A Social Choice Perspective}
\vskip -1.5pt\relax

\label{sec:motivation}

\emph{Social choice theory} studies how individual preferences over available options are aggregated into a collective decision~\citep{zwicker2016voting}. To illustrate this, \Cref{fig:motivation_theory} shows two complete preference profiles, $P$ and $P'$, with the same majority-voting support. To be specific, $A$, $B$, and $C$ receive $3$, $2$, and $2$ first-choice selections, respectively. Majority voting therefore selects $A$ in both. However, the remaining options are ranked differently in $P$ and $P'$, giving different pairwise comparison matrices $Y$. Each entry $Y_{uv}^{\mathrm{P}}$ measures the voting advantage of $u$ over $v$ in a direct comparison under $P$.  A positive value means that more rankings place $u$ above $v$. Then, $Y^{\mathrm{P}}$ can be aggregated into \emph{answer-level comparative advantages} $s_u^{\mathcal M, \mathrm{P}}$ using different rules $\mathcal M$. 
The highest-scoring answer is selected as in \Cref{eq:answer_level_aggregation}.  For example, selection under the Condorcet principle~\citep{young1988condorcet,fishburn1977condorcet} can be represented by
\begin{equation}     s_u^{\mathrm{Condorcet}, \mathrm{P}}     =     \mathbf{1}     \left\{         Y_{uv}^{\mathrm{P}}>0         \ \text{for every }v\in\mathcal A\setminus\{u\}     \right\}.     \label{eq:condorcet_indicator} 
\end{equation} 
This selects the answer that wins every direct comparison, whenever one exists. Otherwise, all scores are zero, and an additional selection rule is needed.  
Following this rule, we select $B$ under $P$ in
\cref{fig:motivation_theory} because it defeats $A$ by $4{:}3$
and $C$ by $5{:}2$.
Under $P'$, $A$ defeats both $B$ and $C$ by $5{:}2$ and is selected.
Thus, we arrive at the following proposition.


\begin{propositionbox}
\begin{proposition}
\label{prop:comparison_structure}
There exist two complete preference profiles $P$ and $P'$ such that:
\begin{enumerate}[
    label=(\roman*),
    leftmargin=1.6em,
    labelsep=0.3em,
    topsep=2pt,
    itemsep=0pt,
    parsep=0pt,
    partopsep=0pt
]
    \item Under $P$, the most frequently selected option is still less preferred than another option in direct pairwise comparisons.
    \item $P$ has the same majority-voting scores as $P'$, but different pairwise comparison matrices and different winners.
\end{enumerate}
\end{proposition}
\end{propositionbox}

\textbf{Connection to the Sampling-based LLM reasoning.} In sampling-based LLM reasoning, we only observe each path's first-choice answer $a_i$ and its confidence score $p_i$, leaving the complete preference among the remaining answers unknown. To approximate the missing ordering, we construct a proxy pairwise matrix $Y$ by comparing path confidence scores across answer groups.  A similar disagreement highlighted in \Cref{prop:comparison_structure} also occurs in this setting:  an answer with the highest aggregation score may still lose to another answer in pairwise comparisons. To examine this phenomenon empirically, we consider SC errors with the following disagreement. An incorrect answer $w$ receives at least as many votes as a correct answer $c$, i.e., $s_w^{\mathrm{SC}}\ge s_c^{\mathrm{SC}}$, yet paths supporting $c$ tend to have higher confidence scores than those supporting $w$ in pairwise comparisons. The formal definition is provided in \cref{app:two_failure_modes}.  
\clearpage
\begin{figure}[t]
    \centering
    \includegraphics[width=0.88\textwidth]{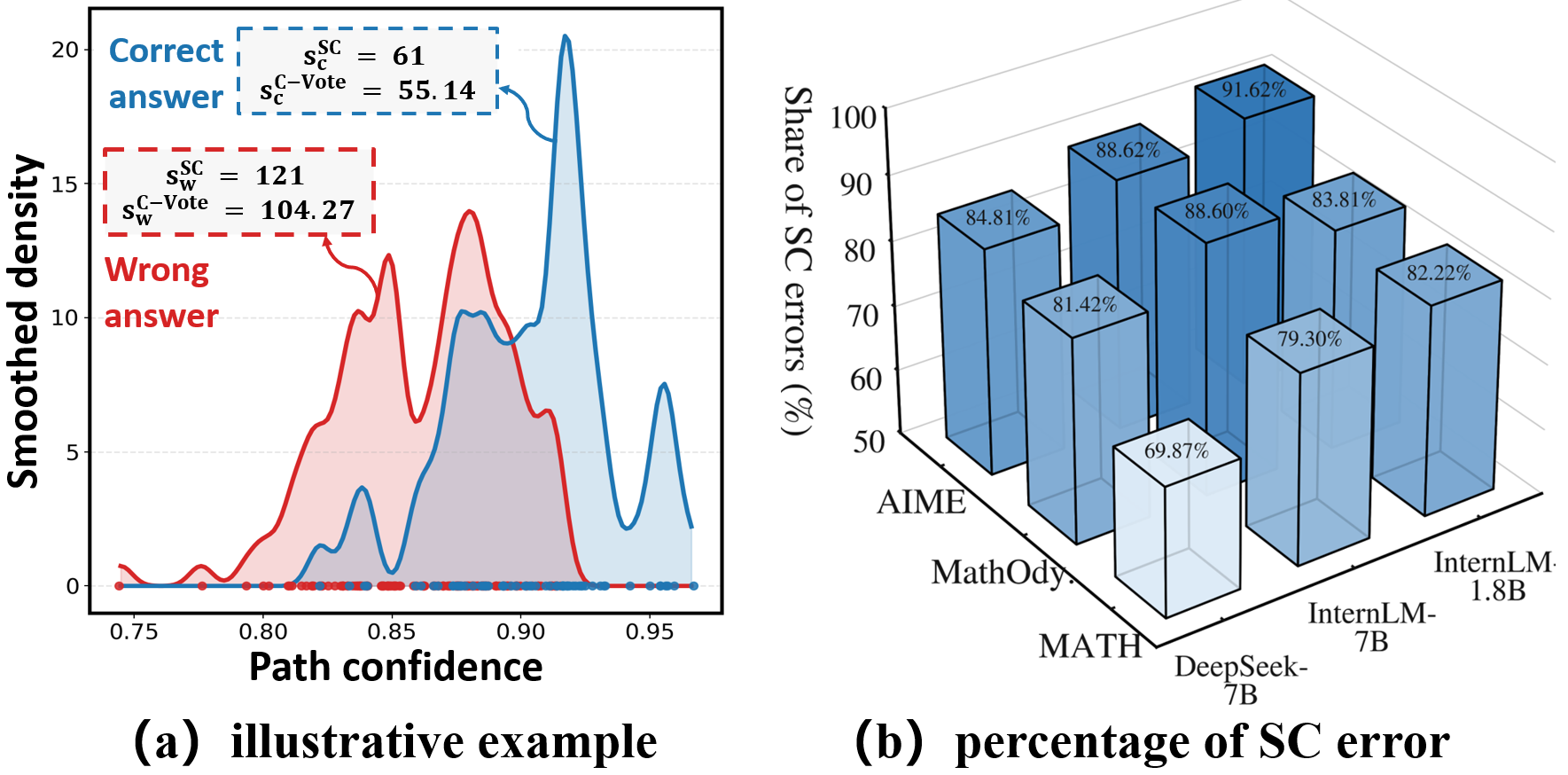}
    \caption{Disagreement between voting and pairwise confidence comparisons: (a) an illustrative example; (b) percentage of recoverable SC errors.}
    \label{fig:sc_error_comparison}
\end{figure}

Using PPL confidence, \textbf{69.87\%--91.62\% of recoverable SC errors} (i.e., candidate pools containing at least one correct answer) satisfy this condition across three models and three mathematical benchmarks as shown in \Cref{fig:sc_error_comparison}(b).

These observations motivate \textbf{enhancing existing voting-based aggregation by incorporating explicit comparisons between answer}.
 We will discuss our solution in the next section.

\vskip -3pt\relax
\vskip 0pt\relax
\section{Pairwise-Guided Path Aggregation}
\vskip -2pt\relax
\label{sec:method}

In this section, we propose \emph{Pairwise-Guided Logit Adjustment} ($\mathsf{PGLA}$) for Path Aggregation. This method aims to construct answer-level comparative evidence through pairwise confidence comparisons and combine it with the voting distribution, as an alternative to the Condorcet approach.

\vskip -2pt\relax
\vskip 0pt\relax
\subsection{Pairwise Confidence Comparisons}
\vskip -1.5pt\relax
\label{sec:ordinal_pairwise_evidence}

Sampled reasoning paths provide answers and confidence scores
rather than complete preference rankings.
We therefore use confidence comparisons between paths supporting
different answers to construct a proxy for answer-level preference comparisons.
For each answer $u\in\mathcal A$, let
$C_u=\{i:a_i=u\}$ denote the indices of its supporting paths.
For two different answers $u,v\in\mathcal A$, we define the corresponding
entry of the confidence-based comparison matrix $Y$ as
\begin{equation}
    Y_{uv}
    =
    \frac{1}{|C_u||C_v|}
    \sum_{i\in C_u}
    \sum_{j\in C_v}
    \operatorname{sgn}(p_i-p_j),
    \qquad u\neq v,
    \label{eq:pairwise_flow}
\end{equation}
Each comparison contributes $+1$, $0$, or $-1$ when the path supporting $u$ has a higher, equal, or lower confidence score.
A positive $Y_{uv}$ means that paths supporting $u$ have higher confidence more often than paths supporting $v$ in these comparisons than the reverse.

We next summarize $Y$ as an answer-level \emph{comparative advantage}.
For a voting aggregator $\mathcal V$, $s_u^{\mathcal V}\geq0$ is the base voting support for the answer $u$ defined in \Cref{eq:sc_score,eq:confidence_voting_score}.
We give greater weight to comparisons against answers with stronger voting support.
Accordingly, the comparative advantage of answer $u$ is 
\begin{equation}
    s_u^{\mathrm{Pair}} = \sum_{v\neq u}\pi_{v\mid u}Y_{uv},  \qquad   \pi_{v\mid u}=\frac{s_v^{\mathcal V}}{\sum_{k\neq u}s_k^{\mathcal V}},
    \qquad v\neq u.
    \label{eq:scpa_correction}
\end{equation}
$s_u^{\mathrm{Pair}}\in[-1,1]$ is a weighted average of comparison scores.
Alternative rules include uniform averaging of pairwise scores;
Copeland, which counts pairwise wins minus
losses~\citep{zwicker2016voting}; and Maximin, which takes the minimum
pairwise score~\citep{fischer2016weighted}.
We use \Cref{eq:scpa_correction} by default. Definitions and matched
comparisons of the alternatives appear in \Cref{app:comparative_rules}.

Regardless of the aggregation rule, $s_u^{\mathrm{Pair}}$ summarizes
confidence comparisons rather than the preference votes used by
the Condorcet principle in \cref{eq:condorcet_indicator}. Since this score does not directly consider answer $u$'s own voting support,
we therefore explore to combine this comparative advantage with the voting information provided by any voting-based aggregation strategy $\mathcal V$.

\vskip -2pt\relax
\vskip 0pt\relax
\subsection{Combining with Voting Information}
\vskip -1.5pt\relax
\label{sec:global_local_reconciliation}

Our goal is to refine the answer-level scores $s_u^{\mathcal V}$
with two complementary objectives:
\textbf{(G1)} favoring answers with stronger comparative advantages
and \textbf{(G2)} exploiting the voting support provided by
the aggregator $\mathcal V$.
We formulate this refinement over answer distributions,
using the normalized base scores to obtain the reference distribution
$\bar{\mathbf s}^{\mathcal V}$, with
$\bar{s}_u^{\mathcal V}
=s_u^{\mathcal V}/\sum_{v\in\mathcal A}s_v^{\mathcal V}$.
We then seek an adjusted distribution through the following
regularized objective:
\begin{equation}
    \bar{\mathbf s}^{\mathsf{PGLA}}
    \in
    \underset{\bar{\mathbf s}}{\arg\max}
    \Biggl[
        \underbrace{
            \sum_{u\in\mathcal A}
            \bar{s}_u s_u^{\mathrm{Pair}}
        }_{\text{(G1)}}
        -
        \underbrace{
            D_{\mathrm{KL}}\!\bigl(
                \bar{\mathbf s}\,\|\,
                \bar{\mathbf s}^{\mathcal V}
            \bigr)
            \vphantom{
                \sum_{u\in\mathcal A}
                \bar{s}_u s_u^{\mathrm{Pair}}
            }
        }_{\text{(G2)}}
    \Biggr].
    \label{eq:variational_scpa}
\end{equation}
Here, $\bar{\mathbf s}$ is the optimization variable,
with $\bar{s}_u\geq0$ and $\sum_{u\in\mathcal A}\bar{s}_u=1$;
$\bar{\mathbf s}^{\mathcal V}$ and $s_u^{\mathrm{Pair}}$
are fixed by the observed candidate pool.

The first term \textbf{rewards favorable answer comparisons},
allowing pairwise confidence evidence to guide answer selection
\textbf{(G1)}.
The KL term \textbf{penalizes departures from the voting distribution},
using the original voting support to constrain this adjustment
\textbf{(G2)}.
The resulting distribution balances comparative evidence with voting support.
Intuitively, reversing a strong voting preference requires a larger
difference in comparative scores, whereas pairwise evidence can
more readily influence selection when answers have similar voting support.
This formulation extends the \textbf{pairwise-comparison perspective
of the Condorcet principle} to confidence-based path aggregation.

At first glance, \Cref{eq:variational_scpa} makes the property of the solution unclear. Fortunately, the following proposition can alleviate this issue, with proof in \Cref{app:variational_proof}.

\begin{propositionbox}
\begin{proposition}
\label{prop:variational_characterization}
If $\bar{s}_u^{\mathcal V}>0$ for every $u\in\mathcal A$,
the objective in \Cref{eq:variational_scpa}
is strictly concave and has the unique solution
\begin{equation}
    \bar{s}_u^{\mathsf{PGLA}}
    =
    \frac{
        \bar{s}_u^{\mathcal V}
        \exp(s_u^{\mathrm{Pair}})
    }{
        \sum_{v\in\mathcal A}
        \bar{s}_v^{\mathcal V}
        \exp(s_v^{\mathrm{Pair}})
    }.
    \label{eq:scpa_distribution}
\end{equation}
\end{proposition}
\end{propositionbox}
\vskip -1.5pt\relax

Since $\bar{s}_u^{\mathcal V}\propto s_u^{\mathcal V}$,
the optimal distribution in \Cref{eq:scpa_distribution}
can be obtained by normalizing the score
$s_u^{\mathsf{PGLA}}
=s_u^{\mathcal V}\exp(s_u^{\mathrm{Pair}})$.
Taking logarithms gives the equivalent formulation in the logit space:
\begin{equation}
\log\!\left(s_u^{\mathsf{PGLA}}\right)
=
\log\!\left(s_u^{\mathcal V}\right)
+
s_u^{\mathrm{Pair}}.
    \label{eq:scpa_score}
\end{equation}
The highest-scoring answer is selected as in
\Cref{eq:answer_level_aggregation}.


\begin{remarkbox}
\begin{remark}[Test-time logit adjustment]
\label{rem:test_time_logit_adjustment}
Thus, we find that the proposed $\mathsf{PGLA}$ can be viewed as a form of \textbf{\textit{Pairwise-guided logit adjustment}}, operating on answer-logit in the observed sampling pool. This additive form connects $\mathsf{PGLA}$ to the classical strategy of Logit Adjustment (LA)~\citep{menon2021long}. In long-tailed classification, post-hoc LA uses offsets based on class priors to reduce the preference for frequent classes. Here, \textbf{this adjustment allows pairwise confidence comparison evidence to compensate for weaker voting support}. What's more, as confidence scores better distinguish correct from incorrect paths, $\mathsf{PGLA}$ can make more reliable pairwise comparisons and further improve aggregation performance. In \Cref{sec:controlled_ranking_analysis}, we test how the reliability of these confidence scores affects $\mathsf{PGLA}$ accuracy.
\end{remark}
\end{remarkbox}





\begingroup
\setlength{\parskip}{4pt}
\vskip -3pt\relax
\vskip 0pt\relax
\section{Experiments}
\vskip -2pt\relax
\label{sec:exp}
In this section, we conduct extensive experiments to evaluate the effectiveness of $\mathsf{PGLA}$. More experimental details and results are provided in \Cref{app:additional_setup} and \Cref{app:additional_results}.

\begin{table*}[t]
    \centering
    \caption{Performance of reasoning models at different scales on three mathematical reasoning benchmarks,
GPQA, and three open-ended code-generation benchmarks, using $K = 32$ sampled responses per
problem. Each voting-based baseline is followed by its \(\mathsf{PGLA}\)-enhanced variant,
denoted by \textbf{+Ours}. The enhanced HumanEval+ and MBPP+ tests for the same sampled pools are reported in \Cref{tab:confidence_transfer_code_plus}.}
    \label{tab:reasoning}
    \fontsize{7}{8.1}\selectfont
    \renewcommand{\arraystretch}{0.96}
    \setlength{\tabcolsep}{2pt}
    \begin{tabular}{@{\hspace{2pt}}p{\dimexpr 0.235\textwidth-4pt\relax}*{7}{>{\centering\arraybackslash}p{\dimexpr 0.1092857\textwidth-4pt\relax}}@{\hspace{2pt}}}
        \toprule
        \textbf{Method} & \textbf{AIME24} & \textbf{AIME25} & \textbf{HMMT25} & \textbf{GPQA-D}
        & \textbf{HE} & \textbf{MBPP} & \textbf{LCB} \\
        \addlinespace[2pt]
        \rowcolor{gray!14} \multicolumn{8}{l}{\strut\hspace{1pt}\textbf{Qwen3-1.7B Reasoning mode}} \\
        \addlinespace[1.5pt]
        PPL Best-of-$N$ & $36.00_{\pm 4.42}$ & $34.00_{\pm 2.91}$ & $23.00_{\pm 3.48}$ & $39.90_{\pm 2.02}$ & $81.10_{\pm 2.06}$ & $75.13_{\pm 0.68}$ & $30.67_{\pm 1.47}$ \\
        DeepConf Best-of-$N$ & $62.00_{\pm 4.52}$ & $42.67_{\pm 2.49}$ & $28.33_{\pm 3.73}$ & $42.93_{\pm 1.52}$ & $84.15_{\pm 2.69}$ & $76.72_{\pm 1.18}$ & $34.29_{\pm 1.62}$ \\
        Self-Certainty Best-of-$N$ & $60.67_{\pm 4.16}$ & $46.67_{\pm 3.33}$ & $30.67_{\pm 4.42}$ & $42.17_{\pm 0.25}$ & $85.06_{\pm 2.70}$ & $81.75_{\pm 1.04}$ & $33.33_{\pm 1.64}$ \\
        \addlinespace[1.5pt]
        SC & $69.50_{\pm 1.98}$ & $47.33_{\pm 3.67}$ & $27.00_{\pm 1.00}$ & $44.44_{\pm 0.51}$ & $83.84_{\pm 1.32}$ & $83.07_{\pm 0.89}$ & $29.90_{\pm 1.33}$ \\
        \rowcolor{gray!5} \hspace{7pt}\textbf{+ Ours (DeepConf)} & $\mathbf{75.00}_{\pm 1.67}$ & $\mathbf{51.33}_{\pm 2.67}$ & $\mathbf{30.00}_{\pm 2.98}$ & $\mathbf{46.72}_{\pm 0.25}$ & $\mathbf{88.11}_{\pm 2.24}$ & $\mathbf{85.45}_{\pm 0.80}$ & $\mathbf{34.29}_{\pm 1.36}$ \\
        \rowcolor{gray!5} \hspace{7pt}\textbf{+ Ours (Self-Certainty)} & $74.00_{\pm 1.33}$ & $50.67_{\pm 3.27}$ & $\mathbf{30.33}_{\pm 2.77}$ & $44.95_{\pm 0.51}$ & $87.80_{\pm 1.61}$ & $\mathbf{85.45}_{\pm 0.80}$ & $\mathbf{33.33}_{\pm 1.35}$ \\
        \addlinespace[1.5pt]
        DeepConf-Weighted SC & $70.00_{\pm 1.49}$ & $48.33_{\pm 4.28}$ & $27.00_{\pm 1.00}$ & $44.19_{\pm 0.76}$ & $85.06_{\pm 1.11}$ & $84.66_{\pm 0.67}$ & $33.71_{\pm 1.30}$ \\
        \rowcolor{gray!5} \hspace{7pt}\textbf{+ Ours} & $\mathbf{75.00}_{\pm 1.67}$ & $\mathbf{51.00}_{\pm 3.00}$ & $\mathbf{31.33}_{\pm 3.40}$ & $\mathbf{46.21}_{\pm 0.76}$ & $\mathbf{87.50}_{\pm 2.22}$ & $\mathbf{85.45}_{\pm 0.80}$ & $\mathbf{34.67}_{\pm 1.63}$ \\
        \addlinespace[1.5pt]
        Self-Certainty-Weighted SC & $70.33_{\pm 1.80}$ & $48.67_{\pm 4.27}$ & $27.00_{\pm 1.00}$ & $44.19_{\pm 0.76}$ & $83.54_{\pm 1.07}$ & $84.66_{\pm 0.57}$ & $32.57_{\pm 1.11}$ \\
        \rowcolor{gray!5} \hspace{7pt}\textbf{+ Ours} & $\mathbf{74.00}_{\pm 1.33}$ & $\mathbf{50.67}_{\pm 3.27}$ & $\mathbf{31.00}_{\pm 3.35}$ & $\mathbf{44.70}_{\pm 0.76}$ & $\mathbf{88.11}_{\pm 1.94}$ & $\mathbf{85.71}_{\pm 0.86}$ & $\mathbf{33.71}_{\pm 1.31}$ \\
        \addlinespace[1.5pt]
        Self-Certainty Borda & $73.33_{\pm 1.49}$ & $50.33_{\pm 3.48}$ & $29.33_{\pm 2.91}$ & $43.69_{\pm 2.27}$ & $86.59_{\pm 1.54}$ & $85.19_{\pm 0.86}$ & $32.95_{\pm 1.28}$ \\
        \rowcolor{gray!5} \hspace{7pt}\textbf{+ Ours} & $\mathbf{74.67}_{\pm 1.63}$ & $\mathbf{50.67}_{\pm 3.27}$ & $\mathbf{34.33}_{\pm 3.96}$ & $\mathbf{44.95}_{\pm 2.02}$ & $\mathbf{88.11}_{\pm 1.83}$ & $\mathbf{85.45}_{\pm 0.77}$ & $\mathbf{33.52}_{\pm 1.43}$ \\
        \addlinespace[1.5pt]
        DeepConf-Filter (90\%) & $70.33_{\pm 1.80}$ & $50.67_{\pm 4.42}$ & $27.00_{\pm 1.80}$ & $44.44_{\pm 1.01}$ & $86.28_{\pm 1.39}$ & $85.19_{\pm 0.71}$ & $33.90_{\pm 1.38}$ \\
        \rowcolor{gray!5} \hspace{7pt}\textbf{+ Ours} & $\mathbf{75.00}_{\pm 1.67}$ & $\mathbf{52.33}_{\pm 3.67}$ & $\mathbf{33.67}_{\pm 2.33}$ & $\mathbf{47.22}_{\pm 0.25}$ & $\mathbf{87.50}_{\pm 2.32}$ & $\mathbf{85.71}_{\pm 0.85}$ & $\mathbf{34.48}_{\pm 1.63}$ \\
        \addlinespace[2pt]
        \rowcolor{gray!14} \multicolumn{8}{l}{\strut\hspace{1pt}\textbf{Qwen3-4B Reasoning mode}} \\
        \addlinespace[1.5pt]
        PPL Best-of-$N$ & $57.00_{\pm 2.33}$ & $64.33_{\pm 3.96}$ & $40.33_{\pm 4.33}$ & $54.71_{\pm 1.41}$ & $92.38_{\pm 0.77}$ & $91.85_{\pm 0.75}$ & $44.00_{\pm 1.26}$ \\
        DeepConf Best-of-$N$ & $77.00_{\pm 2.33}$ & $76.00_{\pm 3.89}$ & $50.00_{\pm 4.94}$ & $58.42_{\pm 2.00}$ & $96.19_{\pm 0.81}$ & $93.44_{\pm 0.28}$ & $47.43_{\pm 0.85}$ \\
        Self-Certainty Best-of-$N$ & $81.67_{\pm 2.24}$ & $71.67_{\pm 3.07}$ & $51.67_{\pm 4.01}$ & $59.93_{\pm 1.12}$ & $96.04_{\pm 0.82}$ & $93.70_{\pm 0.41}$ & $47.81_{\pm 1.16}$ \\
        \addlinespace[1.5pt]
        SC & $80.00_{\pm 0.00}$ & $\mathbf{76.33}_{\pm 0.67}$ & $51.50_{\pm 1.57}$ & $58.25_{\pm 0.45}$ & $93.60_{\pm 1.00}$ & $93.92_{\pm 0.58}$ & $41.71_{\pm 1.26}$ \\
        \rowcolor{gray!5} \hspace{7pt}\textbf{+ Ours (DeepConf)} & $\mathbf{83.00}_{\pm 1.00}$ & $\mathbf{76.33}_{\pm 1.80}$ & $54.00_{\pm 2.91}$ & $\mathbf{60.27}_{\pm 0.88}$ & $\mathbf{95.88}_{\pm 0.95}$ & $94.29_{\pm 0.41}$ & $\mathbf{47.62}_{\pm 0.85}$ \\
        \rowcolor{gray!5} \hspace{7pt}\textbf{+ Ours (Self-Certainty)} & $\mathbf{82.00}_{\pm 1.63}$ & $\mathbf{78.67}_{\pm 1.63}$ & $\mathbf{55.67}_{\pm 1.53}$ & $\mathbf{61.28}_{\pm 0.85}$ & $95.73_{\pm 0.55}$ & $\mathbf{94.31}_{\pm 0.49}$ & $\mathbf{48.00}_{\pm 1.12}$ \\
        \addlinespace[1.5pt]
        DeepConf-Weighted SC & $80.00_{\pm 0.00}$ & $\mathbf{76.33}_{\pm 1.00}$ & $51.33_{\pm 1.63}$ & $58.59_{\pm 0.53}$ & $94.36_{\pm 1.01}$ & $94.21_{\pm 0.48}$ & $47.24_{\pm 0.95}$ \\
        \rowcolor{gray!5} \hspace{7pt}\textbf{+ Ours} & $\mathbf{83.33}_{\pm 1.49}$ & $\mathbf{76.33}_{\pm 2.33}$ & $\mathbf{54.00}_{\pm 2.91}$ & $\mathbf{60.10}_{\pm 0.79}$ & $\mathbf{95.88}_{\pm 0.87}$ & $\mathbf{94.26}_{\pm 0.44}$ & $\mathbf{47.62}_{\pm 0.88}$ \\
        \addlinespace[1.5pt]
        Self-Certainty-Weighted SC & $80.00_{\pm 0.00}$ & $76.33_{\pm 1.00}$ & $52.67_{\pm 2.00}$ & $58.42_{\pm 0.38}$ & $94.21_{\pm 0.86}$ & $\mathbf{94.31}_{\pm 0.49}$ & $47.81_{\pm 1.18}$ \\
        \rowcolor{gray!5} \hspace{7pt}\textbf{+ Ours} & $\mathbf{82.00}_{\pm 1.63}$ & $\mathbf{79.33}_{\pm 1.33}$ & $\mathbf{55.33}_{\pm 1.63}$ & $\mathbf{61.45}_{\pm 0.94}$ & $\mathbf{95.73}_{\pm 0.58}$ & $\mathbf{94.31}_{\pm 0.49}$ & $\mathbf{48.19}_{\pm 1.15}$ \\
        \addlinespace[1.5pt]
        Self-Certainty Borda & $81.00_{\pm 1.53}$ & $\mathbf{79.33}_{\pm 1.33}$ & $55.00_{\pm 1.67}$ & $60.10_{\pm 0.60}$ & $94.97_{\pm 0.90}$ & $94.18_{\pm 0.49}$ & $47.81_{\pm 1.18}$ \\
        \rowcolor{gray!5} \hspace{7pt}\textbf{+ Ours} & $\mathbf{82.33}_{\pm 2.13}$ & $\mathbf{79.33}_{\pm 1.33}$ & $\mathbf{55.33}_{\pm 1.63}$ & $\mathbf{60.94}_{\pm 0.75}$ & $\mathbf{95.73}_{\pm 0.58}$ & $\mathbf{94.26}_{\pm 0.31}$ & $\mathbf{48.00}_{\pm 1.14}$ \\
        \addlinespace[1.5pt]
        DeepConf-Filter (90\%) & $80.00_{\pm 0.00}$ & $78.67_{\pm 2.21}$ & $52.67_{\pm 3.27}$ & $58.59_{\pm 0.72}$ & $94.66_{\pm 0.81}$ & $94.23_{\pm 0.39}$ & $47.24_{\pm 0.95}$ \\
        \rowcolor{gray!5} \hspace{7pt}\textbf{+ Ours} & $\mathbf{83.33}_{\pm 2.11}$ & $\mathbf{79.00}_{\pm 1.53}$ & $\mathbf{55.67}_{\pm 2.60}$ & $\mathbf{61.11}_{\pm 1.10}$ & $\mathbf{96.04}_{\pm 0.71}$ & $\mathbf{94.29}_{\pm 0.36}$ & $\mathbf{47.43}_{\pm 0.81}$ \\
        \addlinespace[2pt]
        \rowcolor{gray!14} \multicolumn{8}{l}{\strut\hspace{1pt}\textbf{Qwen3-8B Reasoning mode}} \\
        \addlinespace[1.5pt]
        PPL Best-of-$N$ & $61.00_{\pm 3.35}$ & $59.33_{\pm 2.00}$ & $35.33_{\pm 3.06}$ & $59.43_{\pm 0.63}$ & $93.70_{\pm 1.24}$ & $93.47_{\pm 0.40}$ & $46.29_{\pm 1.87}$ \\
        DeepConf Best-of-$N$ & $79.33_{\pm 4.16}$ & $76.00_{\pm 2.91}$ & $53.67_{\pm 3.14}$ & $63.64_{\pm 0.41}$ & $96.14_{\pm 0.51}$ & $94.44_{\pm 0.44}$ & $49.71_{\pm 0.00}$ \\
        Self-Certainty Best-of-$N$ & $83.67_{\pm 1.80}$ & $76.33_{\pm 2.77}$ & $52.67_{\pm 2.00}$ & $63.47_{\pm 0.63}$ & $96.54_{\pm 0.66}$ & $95.24_{\pm 0.33}$ & $52.38_{\pm 1.08}$ \\
        \addlinespace[1.5pt]
        SC & $81.50_{\pm 1.57}$ & $75.67_{\pm 2.91}$ & $55.94_{\pm 1.17}$ & $63.22_{\pm 0.95}$ & $93.90_{\pm 0.60}$ & $94.71_{\pm 0.39}$ & $44.57_{\pm 1.68}$ \\
        \rowcolor{gray!5} \hspace{7pt}\textbf{+ Ours (DeepConf)} & $\mathbf{86.67}_{\pm 2.11}$ & $\mathbf{79.67}_{\pm 1.80}$ & $58.67_{\pm 1.63}$ & $\mathbf{64.98}_{\pm 0.24}$ & $96.14_{\pm 0.86}$ & $95.68_{\pm 0.39}$ & $49.90_{\pm 0.27}$ \\
        \rowcolor{gray!5} \hspace{7pt}\textbf{+ Ours (Self-Certainty)} & $86.00_{\pm 2.00}$ & $76.67_{\pm 1.49}$ & $\mathbf{60.00}_{\pm 2.11}$ & $64.31_{\pm 0.86}$ & $\mathbf{96.54}_{\pm 1.14}$ & $\mathbf{96.12}_{\pm 0.52}$ & $\mathbf{52.38}_{\pm 1.08}$ \\
        \addlinespace[1.5pt]
        DeepConf-Weighted SC & $81.67_{\pm 1.67}$ & $76.00_{\pm 3.27}$ & $56.67_{\pm 1.49}$ & $63.30_{\pm 1.19}$ & $94.11_{\pm 0.62}$ & $94.97_{\pm 0.46}$ & $49.14_{\pm 0.47}$ \\
        \rowcolor{gray!5} \hspace{7pt}\textbf{+ Ours} & $\mathbf{87.00}_{\pm 2.33}$ & $\mathbf{79.67}_{\pm 1.80}$ & $\mathbf{59.00}_{\pm 1.53}$ & $\mathbf{64.81}_{\pm 0.24}$ & $\mathbf{96.14}_{\pm 0.72}$ & $\mathbf{95.68}_{\pm 0.37}$ & $\mathbf{49.90}_{\pm 0.27}$ \\
        \addlinespace[1.5pt]
        Self-Certainty-Weighted SC & $81.67_{\pm 1.67}$ & $76.00_{\pm 3.27}$ & $56.67_{\pm 1.49}$ & $63.30_{\pm 1.19}$ & $94.31_{\pm 0.62}$ & $95.15_{\pm 0.41}$ & $51.81_{\pm 0.71}$ \\
        \rowcolor{gray!5} \hspace{7pt}\textbf{+ Ours} & $\mathbf{86.33}_{\pm 1.80}$ & $\mathbf{76.67}_{\pm 1.49}$ & $\mathbf{60.00}_{\pm 1.49}$ & $\mathbf{64.31}_{\pm 0.86}$ & $\mathbf{96.54}_{\pm 1.01}$ & $\mathbf{96.21}_{\pm 0.54}$ & $\mathbf{52.38}_{\pm 1.08}$ \\
        \addlinespace[1.5pt]
        Self-Certainty Borda & $83.67_{\pm 1.80}$ & $\mathbf{76.67}_{\pm 1.49}$ & $59.33_{\pm 2.49}$ & $63.97_{\pm 0.95}$ & $95.33_{\pm 0.49}$ & $95.59_{\pm 0.33}$ & $52.00_{\pm 0.47}$ \\
        \rowcolor{gray!5} \hspace{7pt}\textbf{+ Ours} & $\mathbf{86.33}_{\pm 1.80}$ & $\mathbf{76.67}_{\pm 1.49}$ & $\mathbf{60.00}_{\pm 2.11}$ & $\mathbf{64.31}_{\pm 0.86}$ & $\mathbf{96.34}_{\pm 0.91}$ & $\mathbf{96.12}_{\pm 0.52}$ & $\mathbf{52.38}_{\pm 1.08}$ \\
        \addlinespace[1.5pt]
        DeepConf-Filter (90\%) & $81.67_{\pm 1.67}$ & $78.00_{\pm 2.67}$ & $59.00_{\pm 1.53}$ & $63.30_{\pm 0.95}$ & $94.31_{\pm 0.49}$ & $94.97_{\pm 0.40}$ & $49.14_{\pm 0.47}$ \\
        \rowcolor{gray!5} \hspace{7pt}\textbf{+ Ours} & $\mathbf{87.00}_{\pm 2.33}$ & $\mathbf{79.33}_{\pm 1.33}$ & $\mathbf{61.67}_{\pm 2.24}$ & $\mathbf{64.31}_{\pm 0.63}$ & $\mathbf{96.14}_{\pm 0.51}$ & $\mathbf{95.68}_{\pm 0.39}$ & $\mathbf{49.90}_{\pm 0.27}$ \\
        \bottomrule
    \end{tabular}
\end{table*}

\vskip -2pt\relax
\vskip 0pt\relax
\subsection{Experimental Setup}
\vskip -1.5pt\relax
\label{sec:exp_setup}

\textbf{Competitors.} We compare against \textbf{11} representative sampling-based aggregation baselines using the same pool of sampled reasoning paths. The max-based baselines
follow the Best-of-$N$ paradigm, using path \textbf{perplexity confidence (PPL)}~\citep{chen1998evaluation}, \textbf{DeepConf}~\citep{fu2025deepconf}, and \textbf{Self-Certainty}~\citep{kang2025selfcertainty} to score each path. The voting-based baselines include \textbf{self-consistency (SC)}, three \textbf{confidence-weighted SC} variants based on the above scores, the \textbf{Self-Certainty Borda rule}~\citep{kang2025selfcertainty}, \textbf{DeepConf's offline top-$90\%$ filtered weighted
voting}~\citep{fu2025deepconf}, \textbf{adaptive self-consistency}~\citep{aggarwal2023adaptive}, and \textbf{RPC}~\citep{zhou2025rpc}. Beyond internal confidence scores, we evaluate
three baselines based on scores from an external process reward model
(PRM): \textbf{PRM Best-of-$N$}, \textbf{PRM-Weighted SC}, and \textbf{PRM-RPC}.
We use Qwen2.5-Math-PRM-7B as the external
scorer~\citep{lightman2023verify,zhang2025prm}.
PRM-Weighted SC assigns each path a voting weight equal to the product
of its step-level PRM scores and sums these weights within each answer
group. \textbf{\uline{We also apply the same PGLA adjustment to all voting-based aggregators}}.

\textbf{LLM Backbones.} To test generality across model families and scales, we run our pipeline with \textbf{Qwen3-1.7B}, \textbf{Qwen3-4B},
\textbf{Qwen3-8B}, and \textbf{Qwen3-32B}~\citep{yang2025qwen3},
all in reasoning mode. We also include three open math-specialized
LLM backbones: \textbf{DeepSeek-Math-RL-7B}~\citep{shao2024deepseekmath},
\textbf{InternLM2-Math-Plus-1.8B}, and
\textbf{InternLM2-Math-Plus-7B}~\citep{ying2024internlmmath}.

\textbf{Datasets.} We evaluate on ten challenging datasets, \textbf{AIME 2024}~\citep{aimo2024aime}, \textbf{AIME 2025}~\citep{opencompass2025aime}, \textbf{HMMT 2025}~\citep{balunovic2025matharena}, \textbf{AIME}~\citep{zamil2024aime}, \textbf{MATH}~\citep{hendrycks2021math}, \textbf{MathOdyssey}~\citep{fang2024mathodyssey}, and \textbf{GPQA-Diamond (GPQA-D)}, together with the code benchmarks \textbf{HumanEval}~\citep{chen2021evaluatingcode}, \textbf{MBPP}~\citep{austin2021program}, and \textbf{LiveCodeBench V6}~\citep{jain2024livecodebench}. The first four are high-difficulty mathematical competition problems, while GPQA comprises graduate-level STEM reasoning tasks. Most of  benchmarks are widely
adopted in recent evaluations of top reasoning LLMs. 
Additionally, we also report \textbf{HumanEval+} and \textbf{MBPP+} using the strengthened EvalPlus tests~\citep{liu2023evalplus}.

\begin{table*}[t]
    \centering
    \caption{Performance comparison of different mathematics-specialized models in terms of accuracy and expected calibration error (ECE). Following the RPC setting, perplexity-based confidence (PPL) is used by default. Each voting-based baseline is followed by its \(\mathsf{PGLA}\)-enhanced variant, denoted by \textbf{+Ours}.}
    \label{tab:main_results}
    \vskip -3pt\relax
    \fontsize{7.8}{8.3}\selectfont
    \renewcommand{\arraystretch}{0.99}
    \setlength{\tabcolsep}{2pt}
    \begin{tabular}{@{\hspace{2pt}}p{\dimexpr\textwidth-324pt\relax}*{6}{>{\centering\arraybackslash}p{40.5pt}}*{2}{>{\centering\arraybackslash}p{22.5pt}}@{\hspace{2pt}}}
        \toprule
        \multirow{2}{*}{\textbf{Method}} & \multicolumn{2}{c}{\textbf{MATH}} & \multicolumn{2}{c}{\textbf{MathOdyssey}} & \multicolumn{2}{c}{\textbf{AIME}} & \multicolumn{2}{c}{\textbf{Average}} \\
        \cmidrule(lr){2-3} \cmidrule(lr){4-5} \cmidrule(lr){6-7} \cmidrule(lr){8-9}
        & Acc. $\uparrow$ & ECE $\downarrow$ & Acc. $\uparrow$ & ECE $\downarrow$ & Acc. $\uparrow$ & ECE $\downarrow$ & Acc. $\uparrow$ & ECE $\downarrow$ \\
        \addlinespace[2pt]
        \rowcolor{gray!14} \multicolumn{9}{l}{\strut\hspace{1pt}\textbf{InternLM2-Math-Plus-1.8B}} \\
        \addlinespace[1.5pt]
        PPL & $33.23_{\pm 0.25}$ & $61.58_{\pm 0.25}$ & $16.63_{\pm 0.83}$ & $77.20_{\pm 0.82}$ & $1.66_{\pm 0.15}$ & $92.28_{\pm 0.16}$ & $17.17$ & $77.02$ \\
        \addlinespace[1.5pt]
        SC & $36.48_{\pm 0.14}$ & $6.66_{\pm 0.14}$ & $14.57_{\pm 0.45}$ & $18.51_{\pm 0.41}$ & $2.66_{\pm 0.20}$ & $16.84_{\pm 0.16}$ & $17.90$ & $14.00$ \\
        \rowcolor{gray!5} \hspace{7pt}\textbf{+ Ours} & $\mathbf{39.77}_{\pm 0.17}$ & $\mathbf{5.63}_{\pm 0.13}$ & $\mathbf{18.28}_{\pm 0.56}$ & $\mathbf{13.97}_{\pm 0.59}$ & $\mathbf{3.37}_{\pm 0.17}$ & $\mathbf{15.51}_{\pm 0.15}$ & $\mathbf{20.47}$ & $\mathbf{11.70}$ \\
        \addlinespace[1.5pt]
        Weighted SC & $36.72_{\pm 0.12}$ & $6.52_{\pm 0.14}$ & $14.86_{\pm 0.47}$ & $18.05_{\pm 0.50}$ & $2.75_{\pm 0.23}$ & $16.65_{\pm 0.19}$ & $18.11$ & $13.74$ \\
        \rowcolor{gray!5} \hspace{7pt}\textbf{+ Ours} & $\mathbf{39.91}_{\pm 0.22}$ & $\mathbf{5.55}_{\pm 0.17}$ & $\mathbf{18.56}_{\pm 0.56}$ & $\mathbf{13.47}_{\pm 0.56}$ & $\mathbf{3.37}_{\pm 0.17}$ & $\mathbf{15.34}_{\pm 0.15}$ & $\mathbf{20.61}$ & $\mathbf{11.45}$ \\
        \addlinespace[1.5pt]
        Adaptive SC & $36.48_{\pm 0.16}$ & $10.61_{\pm 0.23}$ & $14.43_{\pm 0.51}$ & $22.69_{\pm 0.73}$ & $2.65_{\pm 0.20}$ & $19.25_{\pm 0.25}$ & $17.85$ & $17.52$ \\
        \rowcolor{gray!5} \hspace{7pt}\textbf{+ Ours} & $\mathbf{39.05}_{\pm 0.23}$ & $\mathbf{9.28}_{\pm 0.27}$ & $\mathbf{17.47}_{\pm 0.63}$ & $\mathbf{17.96}_{\pm 0.60}$ & $\mathbf{3.23}_{\pm 0.21}$ & $\mathbf{17.53}_{\pm 0.41}$ & $\mathbf{19.92}$ & $\mathbf{14.92}$ \\
        \addlinespace[1.5pt]
        RPC & $37.87_{\pm 0.14}$ & $6.45_{\pm 0.15}$ & $16.43_{\pm 0.60}$ & $16.25_{\pm 0.76}$ & $3.28_{\pm 0.23}$ & $16.32_{\pm 0.20}$ & $19.19$ & $13.01$ \\
        \rowcolor{gray!5} \hspace{7pt}\textbf{+ Ours} & $\mathbf{40.29}_{\pm 0.27}$ & $\mathbf{5.50}_{\pm 0.22}$ & $\mathbf{20.28}_{\pm 0.48}$ & $\mathbf{10.78}_{\pm 0.60}$ & $\mathbf{3.57}_{\pm 0.21}$ & $\mathbf{14.67}_{\pm 0.23}$ & $\mathbf{21.38}$ & $\mathbf{10.32}$ \\
        \addlinespace[2pt]
        \rowcolor{gray!14} \multicolumn{9}{l}{\strut\hspace{1pt}\textbf{InternLM2-Math-Plus-7B}} \\
        \addlinespace[1.5pt]
        PPL & $47.00_{\pm 0.20}$ & $48.98_{\pm 0.20}$ & $27.51_{\pm 1.18}$ & $67.54_{\pm 1.18}$ & $5.96_{\pm 0.48}$ & $88.98_{\pm 0.49}$ & $26.82$ & $68.50$ \\
        \addlinespace[1.5pt]
        SC & $50.57_{\pm 0.18}$ & $6.72_{\pm 0.19}$ & $28.30_{\pm 0.61}$ & $12.18_{\pm 0.57}$ & $9.40_{\pm 0.21}$ & $14.35_{\pm 0.23}$ & $29.42$ & $11.08$ \\
        \rowcolor{gray!5} \hspace{7pt}\textbf{+ Ours} & $\mathbf{53.99}_{\pm 0.18}$ & $\mathbf{6.10}_{\pm 0.20}$ & $\mathbf{35.30}_{\pm 0.55}$ & $\mathbf{9.69}_{\pm 1.02}$ & $\mathbf{9.95}_{\pm 0.24}$ & $\mathbf{13.90}_{\pm 0.23}$ & $\mathbf{33.08}$ & $\mathbf{9.90}$ \\
        \addlinespace[1.5pt]
        Weighted SC & $50.81_{\pm 0.20}$ & $6.56_{\pm 0.21}$ & $28.82_{\pm 0.67}$ & $11.57_{\pm 0.68}$ & $9.45_{\pm 0.23}$ & $14.27_{\pm 0.25}$ & $29.69$ & $10.80$ \\
        \rowcolor{gray!5} \hspace{7pt}\textbf{+ Ours} & $\mathbf{53.99}_{\pm 0.13}$ & $\mathbf{6.15}_{\pm 0.17}$ & $\mathbf{35.78}_{\pm 0.51}$ & $\mathbf{9.37}_{\pm 1.16}$ & $\mathbf{9.88}_{\pm 0.28}$ & $\mathbf{13.86}_{\pm 0.27}$ & $\mathbf{33.22}$ & $\mathbf{9.79}$ \\
        \addlinespace[1.5pt]
        Adaptive SC & $50.54_{\pm 0.22}$ & $11.02_{\pm 0.23}$ & $28.20_{\pm 0.68}$ & $18.17_{\pm 1.02}$ & $9.38_{\pm 0.23}$ & $17.39_{\pm 0.41}$ & $29.37$ & $15.53$ \\
        \rowcolor{gray!5} \hspace{7pt}\textbf{+ Ours} & $\mathbf{52.66}_{\pm 0.26}$ & $\mathbf{10.53}_{\pm 0.35}$ & $\mathbf{33.14}_{\pm 0.73}$ & $\mathbf{13.22}_{\pm 1.52}$ & $\mathbf{9.83}_{\pm 0.21}$ & $\mathbf{16.45}_{\pm 0.33}$ & $\mathbf{31.87}$ & $\mathbf{13.40}$ \\
        \addlinespace[1.5pt]
        RPC & $51.94_{\pm 0.14}$ & $\mathbf{6.41}_{\pm 0.16}$ & $31.70_{\pm 0.71}$ & $9.88_{\pm 0.80}$ & $9.74_{\pm 0.23}$ & $14.31_{\pm 0.21}$ & $31.13$ & $10.20$ \\
        \rowcolor{gray!5} \hspace{7pt}\textbf{+ Ours} & $\mathbf{53.81}_{\pm 0.12}$ & $6.72_{\pm 0.15}$ & $\mathbf{36.53}_{\pm 1.06}$ & $\mathbf{7.90}_{\pm 1.09}$ & $\mathbf{10.08}_{\pm 0.24}$ & $\mathbf{13.31}_{\pm 0.22}$ & $\mathbf{33.47}$ & $\mathbf{9.31}$ \\
        \addlinespace[2pt]
        \rowcolor{gray!14} \multicolumn{9}{l}{\strut\hspace{1pt}\textbf{Deepseek-Math-RL-7B}} \\
        \addlinespace[1.5pt]
        PPL & $42.50_{\pm 0.23}$ & $53.84_{\pm 0.23}$ & $22.34_{\pm 1.00}$ & $73.85_{\pm 0.99}$ & $3.37_{\pm 0.46}$ & $93.38_{\pm 0.46}$ & $22.74$ & $73.69$ \\
        \addlinespace[1.5pt]
        SC & $53.33_{\pm 0.08}$ & $\mathbf{6.43}_{\pm 0.11}$ & $36.57_{\pm 0.56}$ & $9.40_{\pm 1.21}$ & $9.42_{\pm 0.23}$ & $12.12_{\pm 0.26}$ & $33.11$ & $9.32$ \\
        \rowcolor{gray!5} \hspace{7pt}\textbf{+ Ours} & $\mathbf{53.52}_{\pm 0.10}$ & $6.86_{\pm 0.09}$ & $\mathbf{37.87}_{\pm 0.55}$ & $\mathbf{9.31}_{\pm 0.95}$ & $\mathbf{10.02}_{\pm 0.23}$ & $\mathbf{10.88}_{\pm 0.25}$ & $\mathbf{33.80}$ & $\mathbf{9.02}$ \\
        \addlinespace[1.5pt]
        Weighted SC & $53.35_{\pm 0.09}$ & $\mathbf{6.43}_{\pm 0.10}$ & $36.68_{\pm 0.61}$ & $9.55_{\pm 1.31}$ & $9.43_{\pm 0.31}$ & $12.10_{\pm 0.34}$ & $33.16$ & $9.36$ \\
        \rowcolor{gray!5} \hspace{7pt}\textbf{+ Ours} & $\mathbf{53.53}_{\pm 0.10}$ & $6.84_{\pm 0.09}$ & $\mathbf{37.94}_{\pm 0.58}$ & $\mathbf{9.53}_{\pm 0.79}$ & $\mathbf{10.00}_{\pm 0.24}$ & $\mathbf{10.84}_{\pm 0.25}$ & $\mathbf{33.82}$ & $\mathbf{9.07}$ \\
        \addlinespace[1.5pt]
        Adaptive SC & $53.28_{\pm 0.09}$ & $\mathbf{10.01}_{\pm 0.14}$ & $\mathbf{36.30}_{\pm 0.58}$ & $13.25_{\pm 1.14}$ & $9.36_{\pm 0.22}$ & $14.88_{\pm 0.41}$ & $32.98$ & $12.71$ \\
        \rowcolor{gray!5} \hspace{7pt}\textbf{+ Ours} & $\mathbf{53.44}_{\pm 0.16}$ & $10.10_{\pm 0.15}$ & $36.27_{\pm 0.80}$ & $\mathbf{12.33}_{\pm 0.87}$ & $\mathbf{9.92}_{\pm 0.23}$ & $\mathbf{13.31}_{\pm 0.49}$ & $\mathbf{33.21}$ & $\mathbf{11.91}$ \\
        \addlinespace[1.5pt]
        RPC & $53.36_{\pm 0.11}$ & $\mathbf{6.47}_{\pm 0.12}$ & $37.12_{\pm 0.61}$ & $9.54_{\pm 1.54}$ & $9.51_{\pm 0.28}$ & $11.94_{\pm 0.32}$ & $33.33$ & $9.32$ \\
        \rowcolor{gray!5} \hspace{7pt}\textbf{+ Ours} & $\mathbf{53.52}_{\pm 0.11}$ & $6.85_{\pm 0.10}$ & $\mathbf{38.10}_{\pm 0.61}$ & $\mathbf{8.89}_{\pm 0.76}$ & $\mathbf{10.01}_{\pm 0.26}$ & $\mathbf{10.44}_{\pm 0.29}$ & $\mathbf{33.88}$ & $\mathbf{8.73}$ \\
        \addlinespace[1.5pt]
        \bottomrule
    \end{tabular}
\end{table*}

\begin{table*}[!t]
    \centering
    \caption{Accuracy (\%) using Qwen2.5-Math-PRM-7B scores at sampling budgets $K/2$ and $K$ ($K=64$ for MATH; $128$ otherwise). \textbf{+ Ours} denotes the $\mathsf{PGLA}$-enhanced voting rule.}
    \label{tab:prm_main_results}
    \vskip -3pt\relax
    \fontsize{7.8}{8.3}\selectfont
    \renewcommand{\arraystretch}{0.99}
    \setlength{\tabcolsep}{2pt}
    \begin{tabular}{@{\hspace{2pt}}p{\dimexpr\textwidth-324pt\relax}*{6}{>{\centering\arraybackslash}p{40.5pt}}*{2}{>{\centering\arraybackslash}p{22.5pt}}@{\hspace{2pt}}}
        \toprule
        \multirow{2}{*}{\textbf{Method}} & \multicolumn{2}{c}{\textbf{MATH}} & \multicolumn{2}{c}{\textbf{MathOdyssey}} & \multicolumn{2}{c}{\textbf{AIME}} & \multicolumn{2}{c}{\textbf{Average}} \\
        \cmidrule(lr){2-3} \cmidrule(lr){4-5} \cmidrule(lr){6-7} \cmidrule(lr){8-9}
        & $K/2$ $\uparrow$ & $K$ $\uparrow$ & $K/2$ $\uparrow$ & $K$ $\uparrow$ & $K/2$ $\uparrow$ & $K$ $\uparrow$ & $K/2$ $\uparrow$ & $K$ $\uparrow$ \\
        \addlinespace[2pt]
        \rowcolor{gray!14} \multicolumn{9}{l}{\strut\hspace{1pt}\textbf{InternLM2-Math-Plus-7B}} \\
        \addlinespace[1pt]
        PRM-Weighted SC & $57.56_{\pm 0.31}$ & $58.67_{\pm 0.10}$ & $33.57_{\pm 1.13}$ & $34.34_{\pm 0.58}$ & $15.49_{\pm 0.36}$ & $16.23_{\pm 0.48}$ & $35.54$ & $36.42$ \\
        \rowcolor{gray!5} \hspace{7pt}\textbf{+ Ours} & $\mathbf{58.86}_{\pm 0.29}$ & $\mathbf{60.40}_{\pm 0.24}$ & $\mathbf{37.63}_{\pm 1.18}$ & $\mathbf{38.79}_{\pm 0.62}$ & $\mathbf{15.55}_{\pm 0.42}$ & $\mathbf{16.57}_{\pm 0.35}$ & $\mathbf{37.35}$ & $\mathbf{38.59}$ \\
        \addlinespace[1pt]
        PRM-RPC & $57.74_{\pm 0.28}$ & $58.90_{\pm 0.13}$ & $34.83_{\pm 1.18}$ & $35.48_{\pm 0.40}$ & $\mathbf{15.57}_{\pm 0.42}$ & $16.31_{\pm 0.44}$ & $36.05$ & $36.90$ \\
        \rowcolor{gray!5} \hspace{7pt}\textbf{+ Ours} & $\mathbf{58.88}_{\pm 0.24}$ & $\mathbf{60.41}_{\pm 0.25}$ & $\mathbf{37.63}_{\pm 1.24}$ & $\mathbf{38.87}_{\pm 0.60}$ & $15.53_{\pm 0.37}$ & $\mathbf{16.50}_{\pm 0.28}$ & $\mathbf{37.35}$ & $\mathbf{38.59}$ \\
        \addlinespace[2pt]
        \rowcolor{gray!14} \multicolumn{9}{l}{\strut\hspace{1pt}\textbf{DeepSeek-Math-RL-7B}} \\
        \addlinespace[1pt]
        PRM-Weighted SC & $60.76_{\pm 0.25}$ & $62.04_{\pm 0.20}$ & $39.69_{\pm 0.59}$ & $40.28_{\pm 0.53}$ & $15.77_{\pm 0.27}$ & $16.93_{\pm 0.43}$ & $38.74$ & $39.75$ \\
        \rowcolor{gray!5} \hspace{7pt}\textbf{+ Ours} & $\mathbf{61.17}_{\pm 0.24}$ & $\mathbf{62.41}_{\pm 0.20}$ & $\mathbf{40.18}_{\pm 0.64}$ & $\mathbf{40.69}_{\pm 0.53}$ & $\mathbf{16.13}_{\pm 0.48}$ & $\mathbf{17.49}_{\pm 0.57}$ & $\mathbf{39.16}$ & $\mathbf{40.20}$ \\
        \addlinespace[1pt]
        PRM-RPC & $60.91_{\pm 0.27}$ & $62.26_{\pm 0.22}$ & $\mathbf{40.05}_{\pm 0.83}$ & $40.67_{\pm 0.51}$ & $15.85_{\pm 0.32}$ & $17.05_{\pm 0.36}$ & $38.94$ & $39.99$ \\
        \rowcolor{gray!5} \hspace{7pt}\textbf{+ Ours} & $\mathbf{61.19}_{\pm 0.23}$ & $\mathbf{62.55}_{\pm 0.19}$ & $39.85_{\pm 0.75}$ & $\mathbf{40.69}_{\pm 0.42}$ & $\mathbf{16.14}_{\pm 0.51}$ & $\mathbf{17.52}_{\pm 0.66}$ & $\mathbf{39.06}$ & $\mathbf{40.26}$ \\
        \addlinespace[1pt]
        \bottomrule
    \end{tabular}
\end{table*}


\vskip -2pt\relax
\vskip 0pt\relax
\subsection{Overall performance}
\vskip -1.5pt\relax

\textbf{Performance using Advanced Models.} \Cref{tab:reasoning}
compares the overall performance of different aggregation methods on Qwen3
reasoning models at three scales. We observe that our method
matches or improves upon the corresponding voting baselines
across the evaluated reasoning and code-generation tasks. In particular, averaged over the six corresponding voting baselines on
Qwen3-8B, \(\mathsf{PGLA}\) improves accuracy by $4.61\%$, $1.78\%$, and $2.63\%$
on AIME24, AIME25, and HMMT25, respectively. The gains also extend to the
larger Qwen3-32B model on mathematical reasoning tasks shown in \Cref{tab:qwen32_extension}. These consistent gains demonstrate that
\(\mathsf{PGLA}\) generalizes well across
model scales and task domains.

\textbf{Performance with the Reliability of Confidence Estimation.}
\Cref{tab:main_results} compares the aggregation methods under internal
confidence estimates from three mathematics-specialized backbones. We observe
that our method consistently improves accuracy while reducing ECE across the
four voting baselines, achieving an average accuracy improvement of $1.96\%$
and an average ECE reduction of $1.42\%$. This demonstrates
the robustness of our method to confidence estimates of varying reliability.

\textbf{Performance with the External Reward Setting.}
\Cref{tab:prm_main_results} and \Cref{tab:aggregation_performance} compares the methods using
product-aggregated rewards from an external process reward model. We observe
that our method improves PRM-Weighted SC and PRM-RPC in most settings. To be specific, at the full budget, \(\mathsf{PGLA}\) raises
their average accuracy from 33.98 to 35.51 and from 34.30 to 35.55,
respectively. This further demonstrates that our method can effectively use
externally provided reward signals.

\begin{figure*}[t]
    \centering
    \begin{minipage}[t]{0.235\textwidth}
    \centering
    \subfigure[AIME24, 1.7B]{
        \includegraphics[width=0.94\textwidth]{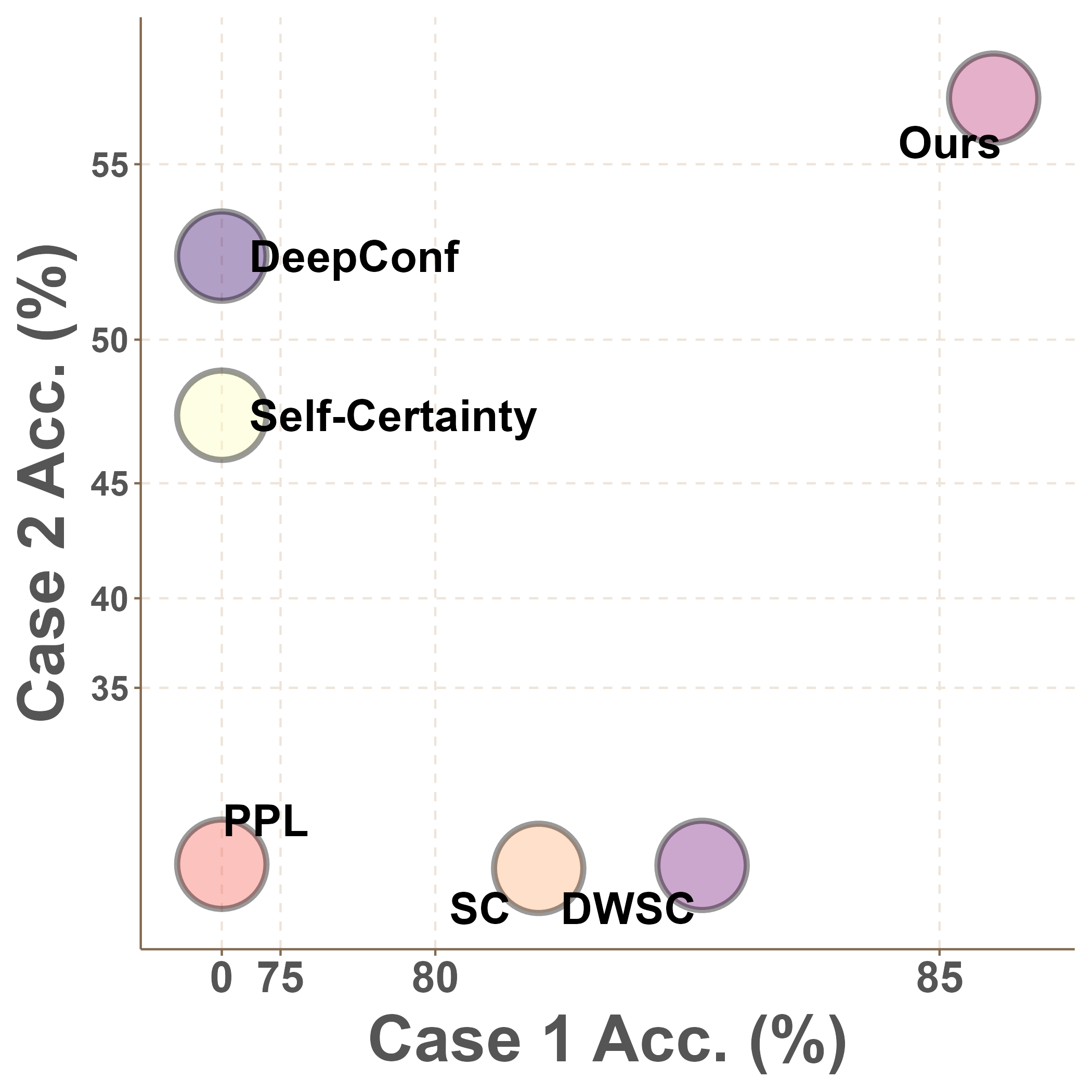}}
    \end{minipage}
    \hfill
    \begin{minipage}[t]{0.235\textwidth}
    \centering
    \subfigure[AIME24, 8B]{
        \includegraphics[width=0.94\textwidth]{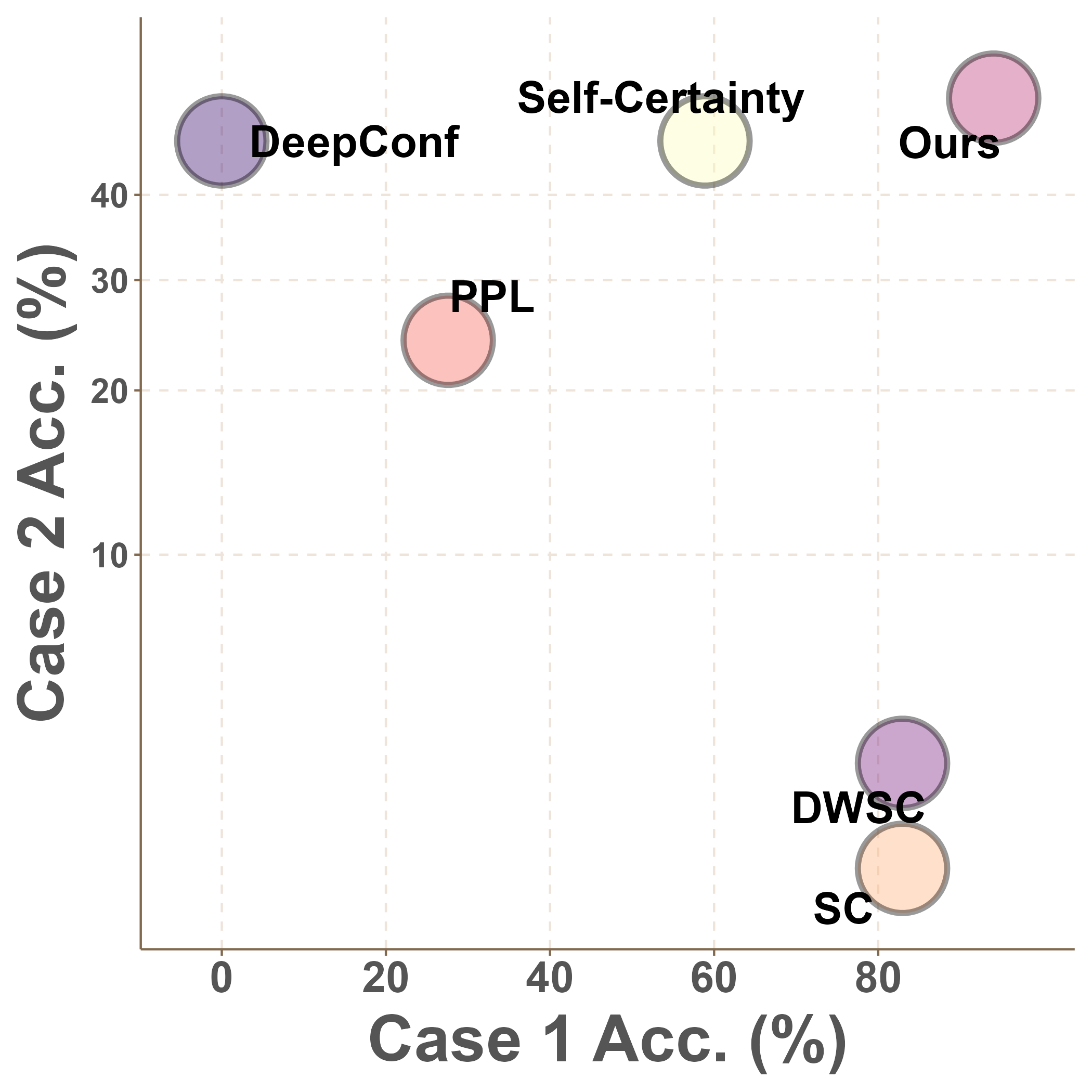}}
    \end{minipage}
    \hfill
    \begin{minipage}[t]{0.235\textwidth}
    \centering
    \subfigure[AIME25, 1.7B]{
        \includegraphics[width=0.94\textwidth]{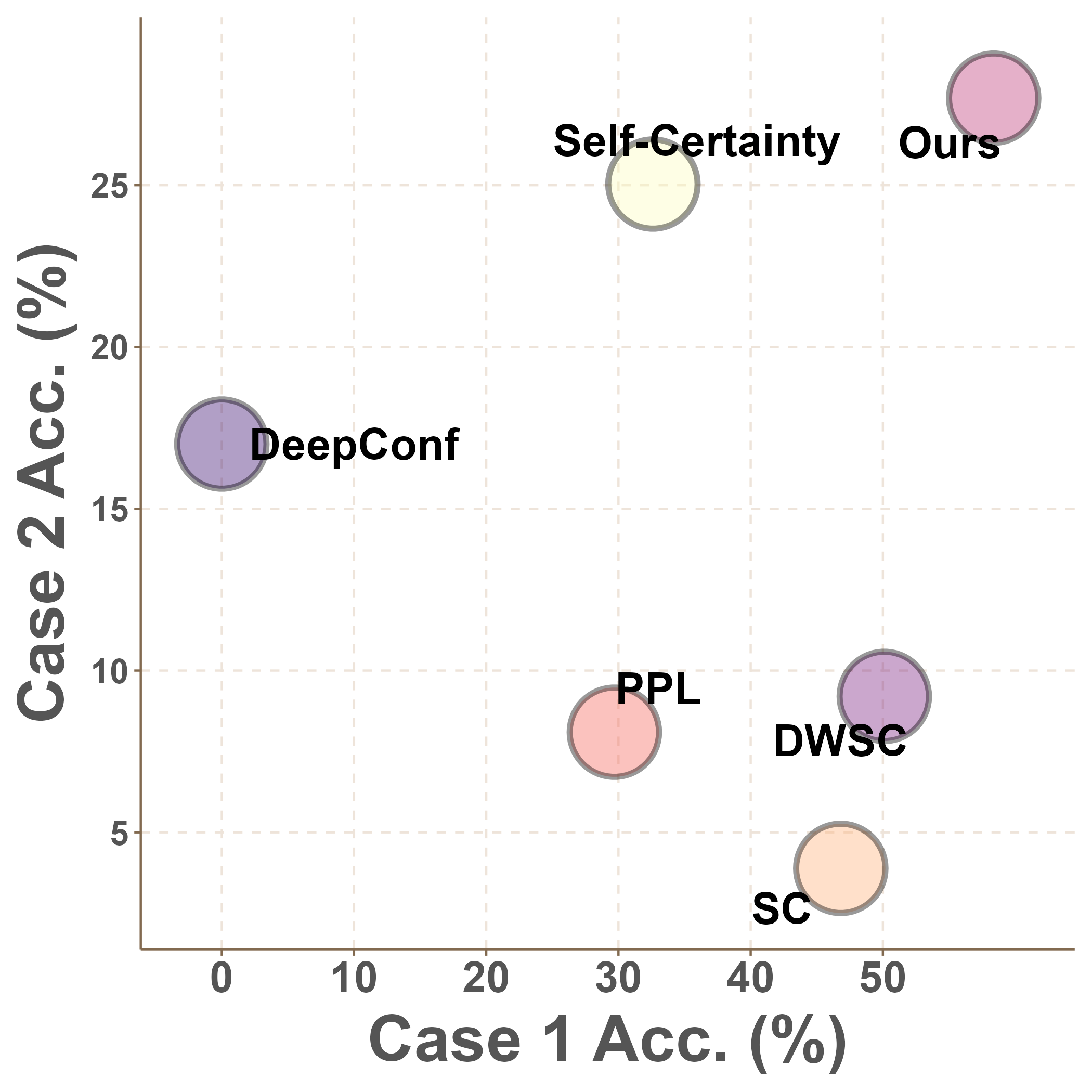}}
    \end{minipage}
    \hfill
    \begin{minipage}[t]{0.235\textwidth}
    \centering
    \subfigure[AIME25, 8B]{
        \includegraphics[width=0.94\textwidth]{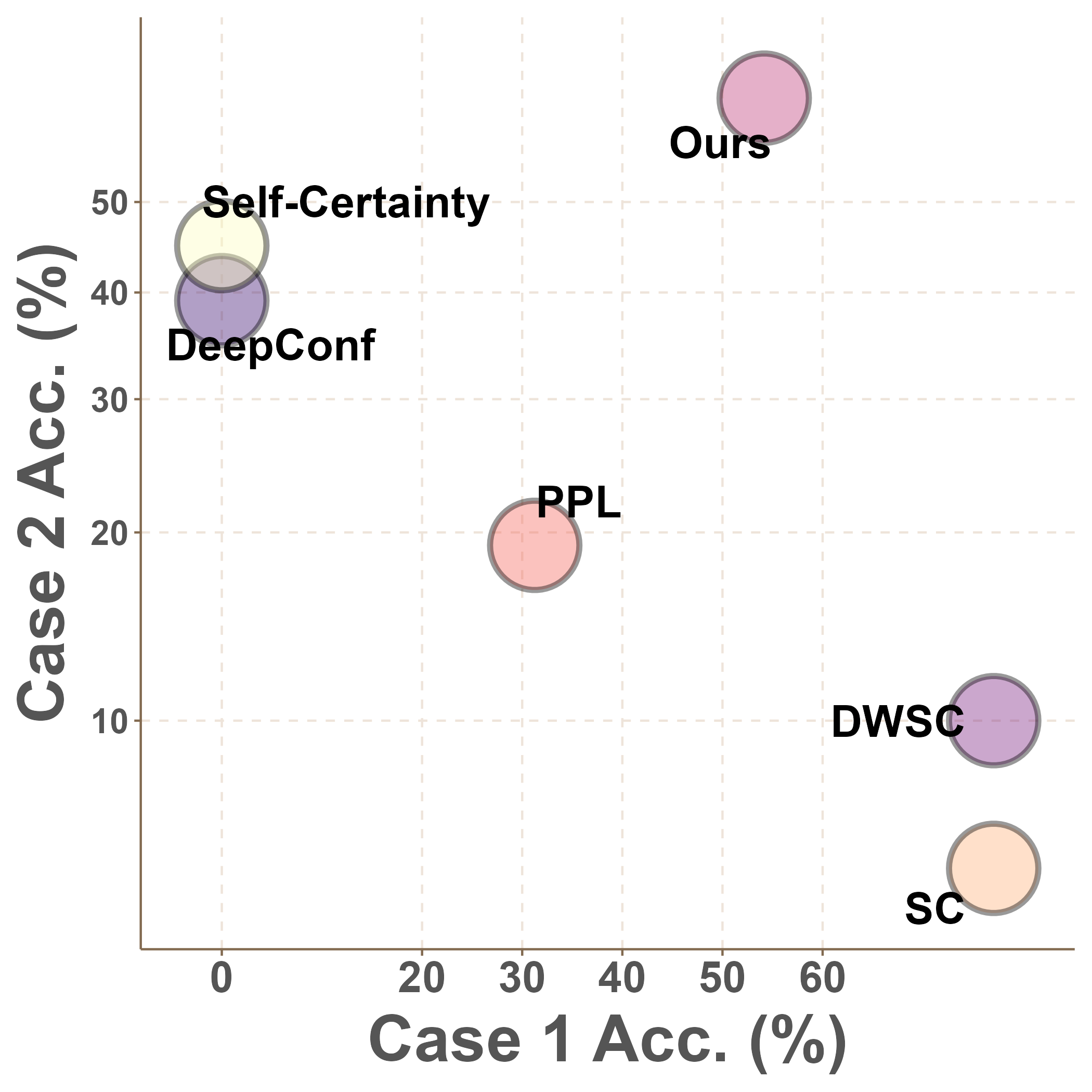}}
    \end{minipage}
    \vskip -3pt\relax
    \caption{Tradeoff between Case I and Case II on AIME24 and AIME25 using Qwen3-1.7B and Qwen3-8B. DWSC denotes the DeepConf-Weighted SC method. }
    \label{fig:case_tradeoff}
    \vskip -4pt\relax
\end{figure*}

\subsection{Case analysis}

We examine two cases that highlight different weaknesses of answer aggregation.
In Case~I, an incorrect answer has higher maximum path confidence, while a correct answer has stronger voting support.
In Case~II, an incorrect answer receives equal or stronger voting support, while pairwise confidence comparisons favor a correct answer.
Best-of-$N$ often fails in Case~I, while voting-based methods typically struggle in Case~II.
As shown in \Cref{fig:case_tradeoff} and \Cref{fig:case_tradeoff_full}, $\mathsf{PGLA}$ better balances the two cases by combining voting support with pairwise comparisons.
See \Cref{app:two_failure_modes} for detailed definitions and examples.

\vskip -2pt\relax
\vskip 0pt\relax
\subsection{Fine-grained Analysis}
\vskip -1.5pt\relax
\label{sec:controlled_ranking_analysis}

\textbf{Effect of Confidence Reliability.} Larger $\lambda$ indicates more reliable PRM scores for pairwise comparisons. As shown in \Cref{fig:rank_comparison}, more reliable scores lead to higher $\mathsf{PGLA}$ accuracy on both models. Base denotes the PRM-Weighted SC baseline. Experimental details are provided in \Cref{app:controlled_rank_details}.

\textbf{Scaling with Varying $K$ Values} We examine how the number of sampled reasoning paths affects aggregation performance on MATH and MathOdyssey using \textbf{InternLM2-Math-Plus-7B}. \Cref{fig:ablation_k} varies $K$ over $\{1,2,4,8,16,32,40,48,64,80,96,112,128\}$ . Increasing $K$ generally improves performance by providing richer candidate sets, and $\mathsf{PGLA}$ remains consistently strong across most of the sampling range. 



\textbf{Performance using High Sampling Temperatures.} 
To further examine robustness under more diverse sampled reasoning paths, we conduct additional experiments on MathOdyssey and AIME using InternLM2-Math-Plus-1.8B, InternLM2-Math-Plus-7B, and DeepSeek-Math-RL-7B. At higher sampling temperatures $T=1.1$ and $T=1.3$, $\mathsf{PGLA}$ generally outperforms the comparison methods as shown in \Cref{tab:temperature_sensitivity}. This supports its robustness to more diverse reasoning paths.

\textbf{Effect of Various Comparative Advantage Functions.}
We compare the default voting-weighted rule with uniform averaging, Copeland, and Maximin in \Cref{app:comparative_rules}. These rules generally improve accuracy over the voting baselines, with the default achieving strong overall performance. This shows that $\mathsf{PGLA}$ benefits from pairwise evidence under different comparative advantage rules.

\par\noindent\begin{minipage}{\textwidth}
\begin{minipage}[t]{\dimexpr0.5\textwidth-\columnsep\relax}
\vskip 0pt\relax
\textbf{Performance in the Online Setting.}
As shown in \Cref{tab:online_sc_PGLA_18b}, Online SC+$\mathsf{PGLA}$
improves accuracy over full-budget SC by $0.28$--$3.63\%$
while reducing the sampling cost by 50.8\%--73.1\% on
InternLM2-Math-Plus-1.8B. Full comparisons with Adaptive SC and Online SC,
along with results on InternLM2-Math-Plus-7B, are provided in
\Cref{app:online_sampling_efficiency}.
\end{minipage}\hfill
\begin{minipage}[t]{0.5\textwidth}
\vskip 0pt\relax
\centering
\captionsetup{justification=raggedright,singlelinecheck=false}
\captionof{table}{Online setting on InternLM2-1.8B.}
\label{tab:online_sc_PGLA_18b}
\vskip -3pt\relax
\small
\setlength{\tabcolsep}{2.6pt}
\renewcommand{\arraystretch}{0.95}
\resizebox{\linewidth}{!}{%
\begin{tabular}{@{}llccc@{}}
\toprule
\textbf{Dataset} & \textbf{Method} & \textbf{Acc.} & $\overline K$ & \textbf{Saving} \\
\midrule
\multirow{2}{*}{MATH}
 & SC (Full Budget) & $36.48_{\pm0.14}$ & 64.0 & -- \\
 & \textbf{Online SC+$\mathsf{PGLA}$} & $\mathbf{39.32}_{\pm0.31}$ & \textbf{31.5} & \textbf{50.8\%} \\
\addlinespace[1.5pt]
\multirow{2}{*}{MathOdyssey}
 & SC (Full Budget) & $14.57_{\pm0.45}$ & 128.0 & -- \\
 & \textbf{Online SC+$\mathsf{PGLA}$} & $\mathbf{18.20}_{\pm0.90}$ & \textbf{34.5} & \textbf{73.1\%} \\
\addlinespace[1.5pt]
\multirow{2}{*}{AIME}
 & SC (Full Budget) & $2.66_{\pm0.20}$ & 128.0 & -- \\
 & \textbf{Online SC+$\mathsf{PGLA}$} & $\mathbf{2.94}_{\pm0.29}$ & \textbf{38.3} & \textbf{70.1\%} \\
\bottomrule
\end{tabular}%
}
\end{minipage}
\end{minipage}

\begin{figure*}[]
    \centering
    \begin{minipage}[t]{0.48\textwidth}
        \centering
        \subfigure[MATH]{
            \includegraphics[width=0.46\linewidth]{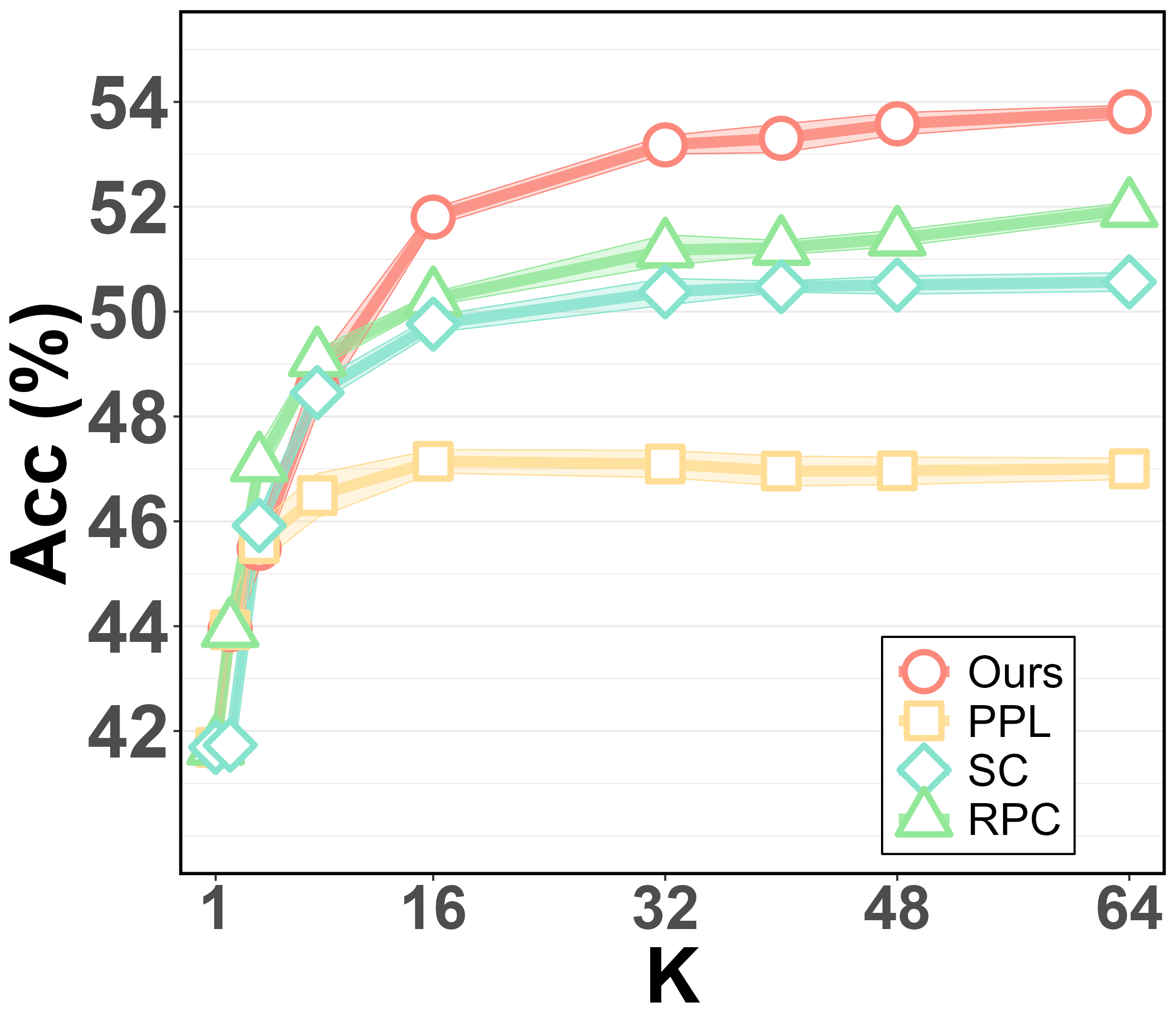}
        }
        \hfill
        \subfigure[MathOdyssey]{
            \includegraphics[width=0.46\linewidth]{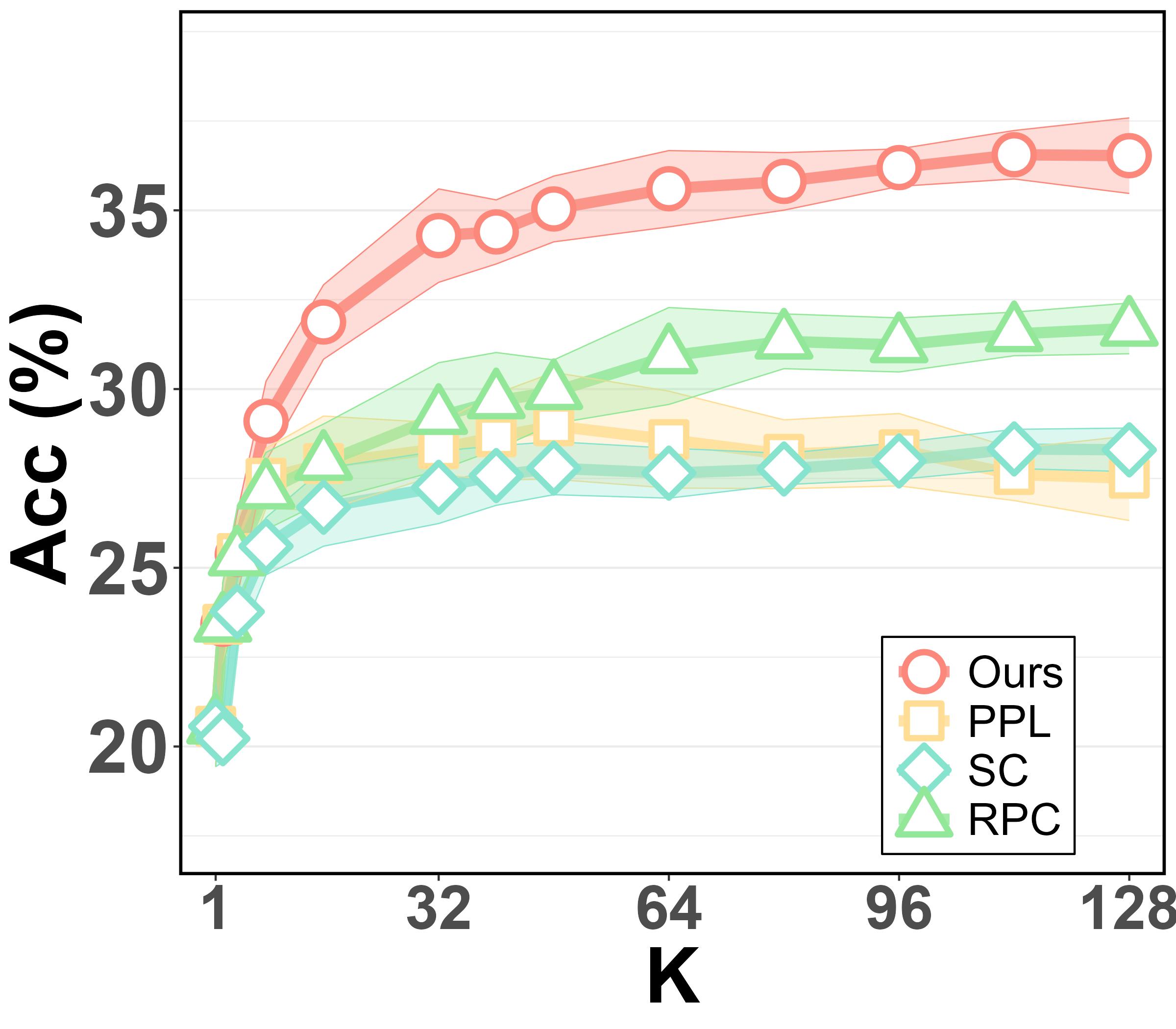}
        }
        \vskip -3pt\relax
        \caption{Scaling behavior under varying numbers of sampled reasoning paths $K$ on MATH and MathOdy. using InternLM2-Math-Plus-7B.}
        \label{fig:ablation_k}
    \end{minipage}
    \hfill
    \begin{minipage}[t]{0.48\textwidth}
        \centering
        \subfigure[DeepSeek]{
            \includegraphics[width=0.46\linewidth]{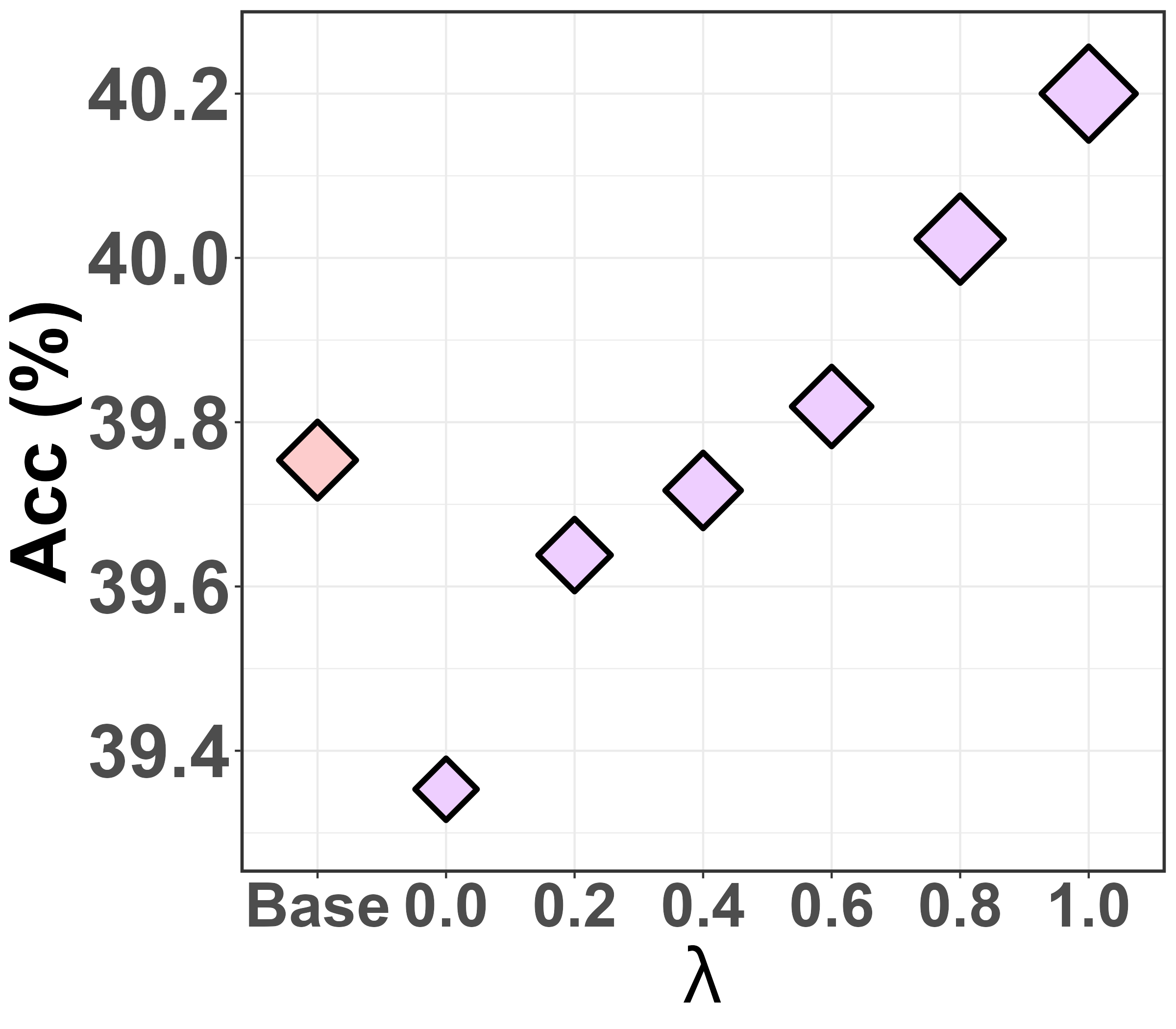}
        }
        \hfill
        \subfigure[InternLM2]{
            \includegraphics[width=0.46\linewidth]{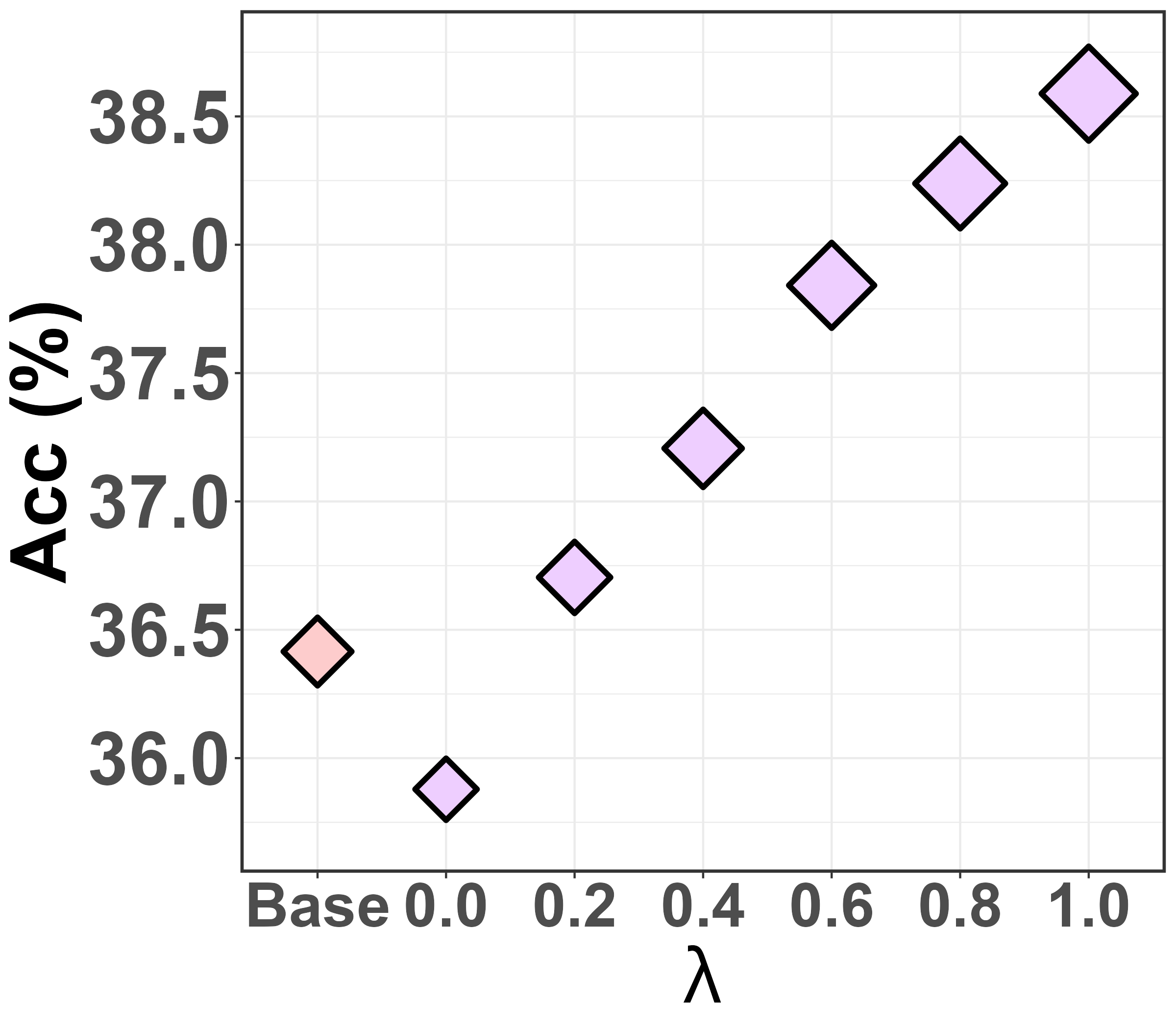}
        }
        \vskip -3pt\relax
        \caption{Effect of confidence reliability in pairwise comparisons on accuracy, averaged over three math datasets.}
        \label{fig:rank_comparison}
    \end{minipage}
    \vskip -4pt\relax
\end{figure*}

\vskip -3pt\relax
\vskip 0pt\relax
\section{Conclusion}
\vskip -2pt\relax
\label{sec:con}

Inspired by social choice theory and the Condorcet principle, we propose \emph{Pairwise-Guided Logit Adjustment} ($\mathsf{PGLA}$), a plug-and-play framework for sampling-based LLM reasoning.
Through pairwise confidence comparisons between sampled reasoning paths, $\mathsf{PGLA}$ estimates the implicit preference structure among candidate answers and captures their comparative advantages.
A KL-regularized objective combines this comparative evidence with voting support, yielding a simple adjustment to the base voting logits.
The framework accommodates different voting methods, confidence sources, and comparative advantage rules.
Experiments on mathematical reasoning, knowledge reasoning, and code generation demonstrate overall gains over voting baselines, while the online extension improves sampling efficiency.

\endgroup
\clearpage
\bibliographystyle{iclr2027_conference}
\bibliography{main}
\clearpage
\appendix
\startcontents[appendices]
\section*{Appendix Contents}
\printcontents[appendices]{}{1}{\setcounter{tocdepth}{3}}
\clearpage
\section{Related Work}
\label{app:related_work}

\noindent\textbf{Sampling-based reasoning and confidence aggregation.}
Sampling-based test-time scaling improves LLM reasoning by generating multiple
reasoning paths and selecting an answer from their collective evidence.
Self-Consistency (SC) aggregates paths through majority voting over their final
answers~\citep{wang2022self}, whereas Best-of-$N$ selects a single candidate using
a quality score, such as a learned verifier or process reward
model~\citep{cobbe2021training,lightman2023verify}. For outputs that cannot be
grouped by exact answers, Universal Self-Consistency uses an LLM to assess
consistency~\citep{chen2023usc}, while Latent Self-Consistency learns
representations for semantic agreement~\citep{oh2025lsc}.

Confidence-aware methods further exploit differences in path quality.
Soft Self-Consistency scores candidate actions using model likelihoods in
interactive tasks~\citep{wang2024soft}. RPC combines internal probability with
answer consistency through Perplexity Consistency and Reasoning
Pruning~\citep{zhou2025rpc}. CISC weights votes by model confidence and shows
that distinguishing correct from incorrect responses \emph{within the same
question} is more relevant to aggregation than cross-question calibration
alone~\citep{taubenfeld2025cisc}. Self-Certainty derives confidence from the
token distribution~\citep{kang2025selfcertainty}, while DeepConf uses local
confidence to filter traces and weight their votes~\citep{fu2025deepconf}.
More recently, PiCSAR combines reasoning and answer likelihoods for candidate
selection~\citep{leang2025picsar}. 
Self-Certainty's Borda variant ranks sampled paths by confidence and accumulates
rank-dependent votes for their answers~\citep{kang2025selfcertainty}.
Ranked Voting SC instead elicits ordered candidate answers within each response
and aggregates these lists using ranked voting rules~\citep{wang2025rankedvoting}.
RISC learns an answer ranker from frequency, semantic, and reasoning-consistency
features~\citep{marina2026risc}. 

Our work studies preference reconstruction from
ordinary single-answer samples. Each path provides one answer and a confidence
score, leaving its ordering over alternatives unobserved. We seek to reconstruct
an \emph{answer-level preference structure} through pairwise confidence
comparisons across answer groups, and reconcile this proxy for the missing
preferences with voting support through KL-regularized aggregation.

\medskip
\noindent\textbf{Efficient and adaptive reasoning.}
The benefits of additional inference computation depend on how it is allocated
across problems and reasoning paths~\citep{snell2024scaling}.
Adaptive-Consistency and Early-Stopping Self-Consistency reduce the number of
completed samples by monitoring answer agreement and stopping when sufficient
evidence has accumulated~\citep{aggarwal2023adaptive,li2024esc}.
DeepConf additionally monitors confidence during generation, terminates
low-confidence traces, and uses answer consensus to stop further
sampling~\citep{fu2025deepconf}. 

Recent work refines both allocation and stopping. Blend-ASC analyzes SC through
mode estimation and voting theory and dynamically distributes a sampling
budget across questions~\citep{feng2025optimal}. ReASC incorporates
response-level confidence into its evidence accumulation and stopping
decisions~\citep{kim2026reasc}. MARS probes intermediate answers during parallel
generation and stops when a conservative bound on future vote changes protects
the current leader; its theoretical target is agreement with the full-budget
vote~\citep{chen2026mars}. Our primary focus is the preference structure used
for answer aggregation. The online extension updates this structure as paths
arrive and stops when the resulting decision remains stable, allowing us to
examine whether preference-aware aggregation also improves sample efficiency.

\clearpage

\section{Representative Methods}
\label{sec:representative_methods}

We follow the notation and scoring rules in \Cref{sec:problem_formulation}.
The $K$ sampled paths $z_i$ produce answers $a_i=g(z_i)$ and confidence
scores $p_i$, forming $\mathcal D=\{(a_i,p_i)\}_{i=1}^{K}$.
For each candidate answer $a\in\mathcal A$, an aggregation method
$\mathcal M$ computes $s_a^{\mathcal M}=\mathcal M(a;\mathcal D)$;
the predicted answer $\hat a$ is selected by
\Cref{eq:answer_level_aggregation}. We first specify possible confidence
scores $p_i$ and then apply the same selection and voting formulas.

\subsection{Path-Confidence Measures and Selection}
\label{sec:confidence_based_selection}

Let $T_i$ be the number of tokens in path $z_i$, and let $z_{i,t}$ be its
$t$-th token. For a vocabulary token $v$, $P_{i,t}(v)$ denotes the model's
predicted probability of $v$ at position $t$, conditioned on the input query
and preceding tokens of $z_i$. The vocabulary is denoted by
$\mathcal V_{\mathrm{tok}}$. The following superscripts distinguish
alternative choices for the confidence $p_i$ in
\Cref{sec:problem_formulation}.

\emph{Perplexity confidence} uses inverse perplexity so that higher values
indicate higher confidence~\citep{chen1998evaluation}:
\begin{equation}
    p_i^{\mathrm{PPL}}
    =\exp\!\left(\frac{1}{T_i}\sum_{t=1}^{T_i}
      \log P_{i,t}(z_{i,t})\right)
    =\operatorname{PPL}(z_i)^{-1}.
    \label{eq:ppl_confidence}
\end{equation}
Here, $\operatorname{PPL}(z_i)$ is the perplexity of the sampled path.
Its negative log-perplexity is $\log p_i^{\mathrm{PPL}}$; exponentiation
preserves its ordering and ties, so either scale gives the same Best-of-$N$
selection and pairwise confidence comparisons. The positive scale above
also supplies the original perplexity-derived voting contribution directly
as $p_i$ in \Cref{eq:confidence_voting_score}.

\emph{Self-Certainty}~\citep{kang2025selfcertainty} measures deviation from
the uniform distribution over the vocabulary:
\begin{equation}
    p_i^{\mathrm{SCert}}
    =-\frac{1}{T_i|\mathcal V_{\mathrm{tok}}|}
      \sum_{t=1}^{T_i}\sum_{v\in\mathcal V_{\mathrm{tok}}}
      \log\!\left(|\mathcal V_{\mathrm{tok}}|P_{i,t}(v)\right).
    \label{eq:self_certainty_score}
\end{equation}
The notation $|\mathcal V_{\mathrm{tok}}|$ denotes the vocabulary size.
Unlike perplexity, this score uses the entire predicted token distribution
at each position, rather than only the probability of the sampled token.

\emph{DeepConf}~\citep{fu2025deepconf} computes token confidences and
aggregates them over local windows. Let $k\in\{1,\ldots,|\mathcal V_{\mathrm{tok}}|\}$
be the number of highest-probability
tokens used at each position, and let $\operatorname{Top}_k(P_{i,t})$ be
that set of vocabulary tokens. Its bottom-$10\%$ variant is
\begin{equation}
    \begin{aligned}
    c_{i,t}&=-\frac{1}{k}
      \sum_{v\in\operatorname{Top}_k(P_{i,t})}\log P_{i,t}(v),
    &G_{i,\ell}&=\frac{1}{|W_{i,\ell}|}
      \sum_{t\in W_{i,\ell}}c_{i,t},\\
    p_i^{\mathrm{DeepConf}}&=\frac{1}{|\mathcal L_i|}
      \sum_{\ell\in\mathcal L_i}G_{i,\ell}.
    \end{aligned}
    \label{eq:deepconf_score}
\end{equation}
Here, $c_{i,t}$ is the token confidence; $\ell$ indexes a local window;
$W_{i,\ell}\subseteq\{1,\ldots,T_i\}$ is its nonempty set of token
positions; and $G_{i,\ell}$ is the average confidence in that window.
The nonempty index set $\mathcal L_i$ selects the lowest-$10\%$ window
confidences, with at least one window retained. Vertical bars denote set
size. Tail and minimum-window variants use the same local-confidence
principle. We use $k=20$, following DeepConf. This score averages negative log-probabilities over the top-$k$ candidate tokens and differs from the negative log-likelihood of the sampled token. Larger values are treated as higher confidence. The resulting token scores and their window averages are nonnegative, allowing direct use as voting weights.

For any chosen confidence $p_i$, Best-of-$N$ uses exactly the answer-level
score in \Cref{eq:best_of_n_score}:
\begin{equation}
    s_a^{\mathrm{BoN}}=\max_{i:a_i=a}p_i,
    \qquad a\in\mathcal A.
    \label{app:best_of_n}
\end{equation}
The index condition $a_i=a$ restricts the maximum to paths producing
answer $a$. Best-of-$N$ is the conventional method name; its sampling
budget here is the $K$ paths defined in \Cref{sec:problem_formulation}.

\subsection{Answer-Level Voting}
\label{sec:majority_based_aggregation}

Self-Consistency (SC)~\citep{wang2022self} assigns one vote to each sampled
path, as in \Cref{eq:sc_score}:
\begin{equation}
    s_a^{\mathrm{SC}}=\sum_{i=1}^{K}\mathbf 1(a_i=a).
    \label{eq:self_consistency}
\end{equation}
The indicator $\mathbf 1(\cdot)$ is one when its condition holds and zero
otherwise. Thus $s_a^{\mathrm{SC}}$ counts paths supporting answer $a$
without using their confidence values.

Confidence-based voting uses the same coefficient $h_i\geq0$ as
\Cref{eq:confidence_voting_score}, with $h_i=0$ for a filtered path:
\begin{equation}
    s_a^{\mathrm{Conf\text{-}Vote}}
    =\sum_{i=1}^{K}h_i\cdot p_i\,\mathbf 1(a_i=a).
    \label{eq:weighted_self_consistency}
\end{equation}
Unfiltered Weighted SC sets $h_i=1$ for every path. PPL-Weighted SC uses
$p_i=p_i^{\mathrm{PPL}}$ from \Cref{eq:ppl_confidence}; DeepConf-Weighted SC
and Self-Certainty-Weighted SC use $p_i^{\mathrm{DeepConf}}$ and
$p_i^{\mathrm{SCert}}$, respectively. Confidence filtering instead sets
$h_i=0$ for discarded paths and $h_i=1$ for retained paths.

RPC~\citep{zhou2025rpc} is another instance of this confidence-based voting
rule. Perplexity Consistency aggregates probability-derived contributions
from distinct path--confidence pairs, while Reasoning Pruning filters
low-confidence paths under its original query-specific mixture rule.
With $p_i=p_i^{\mathrm{PPL}}$, set $h_i=1$ for one representative of each
retained distinct pair $(z_i,p_i)$, and $h_i=0$ for pruned paths and repeated
copies of the same pair.
Its score is therefore
\begin{equation}
    s_a^{\mathrm{RPC}}
    =\sum_{i=1}^{K}h_i\cdot p_i\,\mathbf 1(a_i=a),
    \label{eq:rpc_score}
\end{equation}
where the superscript $\mathrm{RPC}$ identifies this particular choice of
confidence and filtering coefficients in the Conf-Vote family.
Further baseline-specific rules, including rank-weighted voting, are
described in \Cref{app:comparison_methods}.

\section{Case Analysis}
\label{app:two_failure_modes}

We use the Best-of-$N$, SC, and confidence-based voting scores in
\Cref{eq:best_of_n_score,eq:sc_score,eq:confidence_voting_score}.
Fix a voting method $\mathcal M$: its score $s_a^{\mathcal M}$ is either
$s_a^{\mathrm{SC}}$ or an instance of $s_a^{\mathrm{Conf\text{-}Vote}}$
with specified confidence scores $p_i$ and coefficients $h_i$.
Thus, Best-of-$N$ uses a confidence maximum, while voting sums the path
contributions prescribed in \Cref{sec:problem_formulation}.

\begin{figure*}[t]
\centering
\begin{minipage}[t]{0.235\textwidth}
\centering
\subfigure[Case I.]{{
\includegraphics[width=0.94\textwidth]{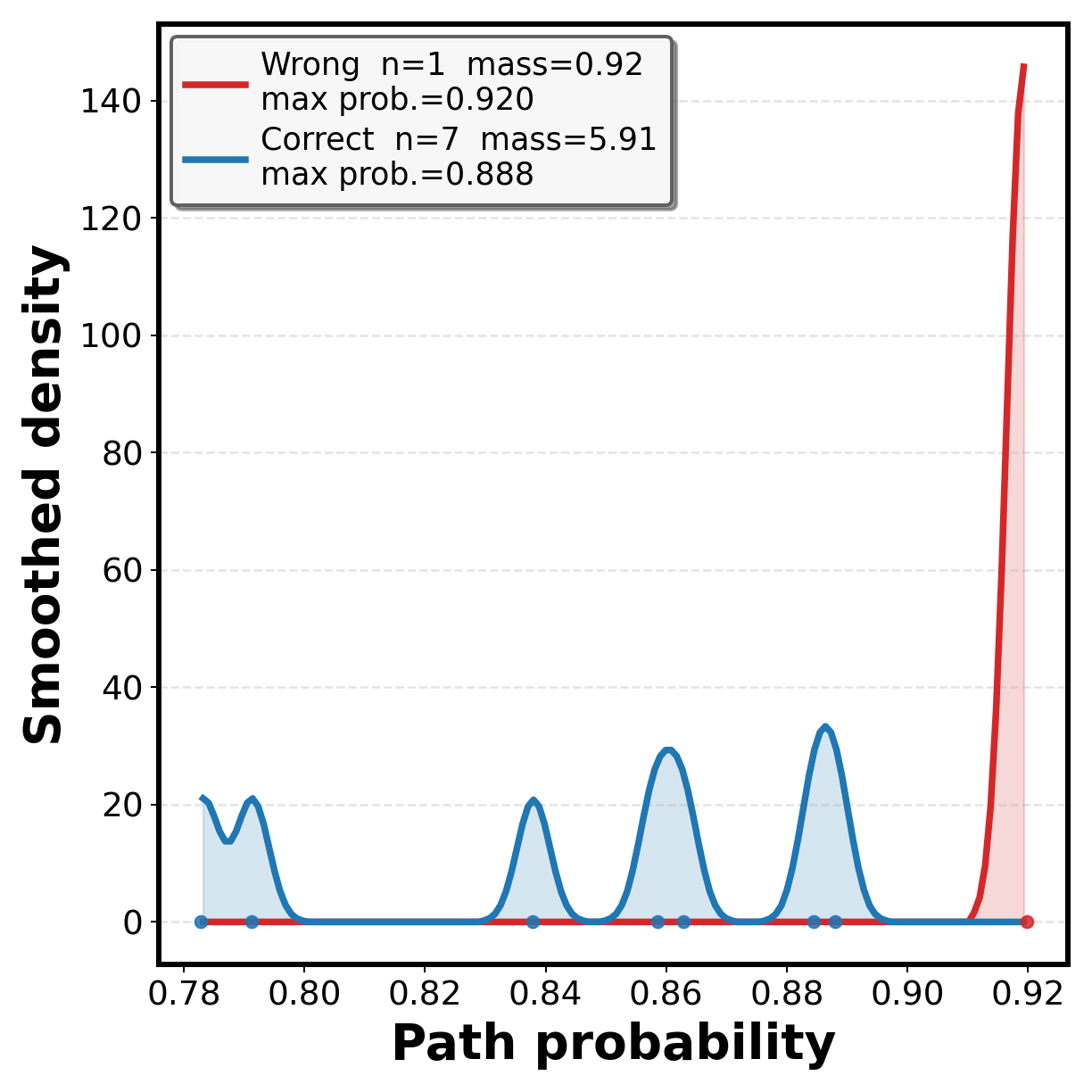}
}}
\end{minipage}
\hfill
\begin{minipage}[t]{0.235\textwidth}
\centering
\subfigure[Case II.]{{
\includegraphics[width=0.94\textwidth]{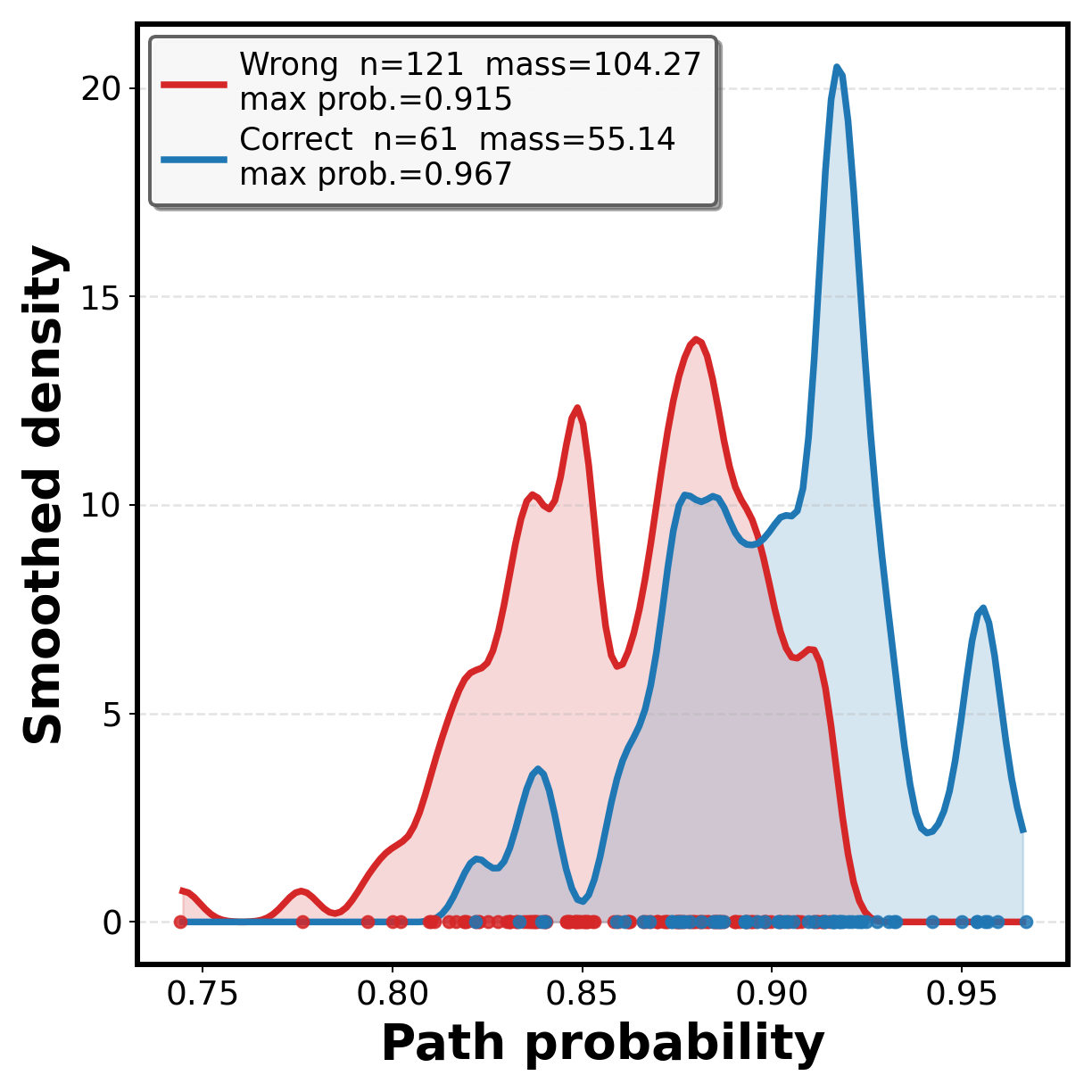}
}}
\end{minipage}
\hfill
\begin{minipage}[t]{0.235\textwidth}
\centering
\subfigure[MATH.]{{
\includegraphics[width=0.94\textwidth]{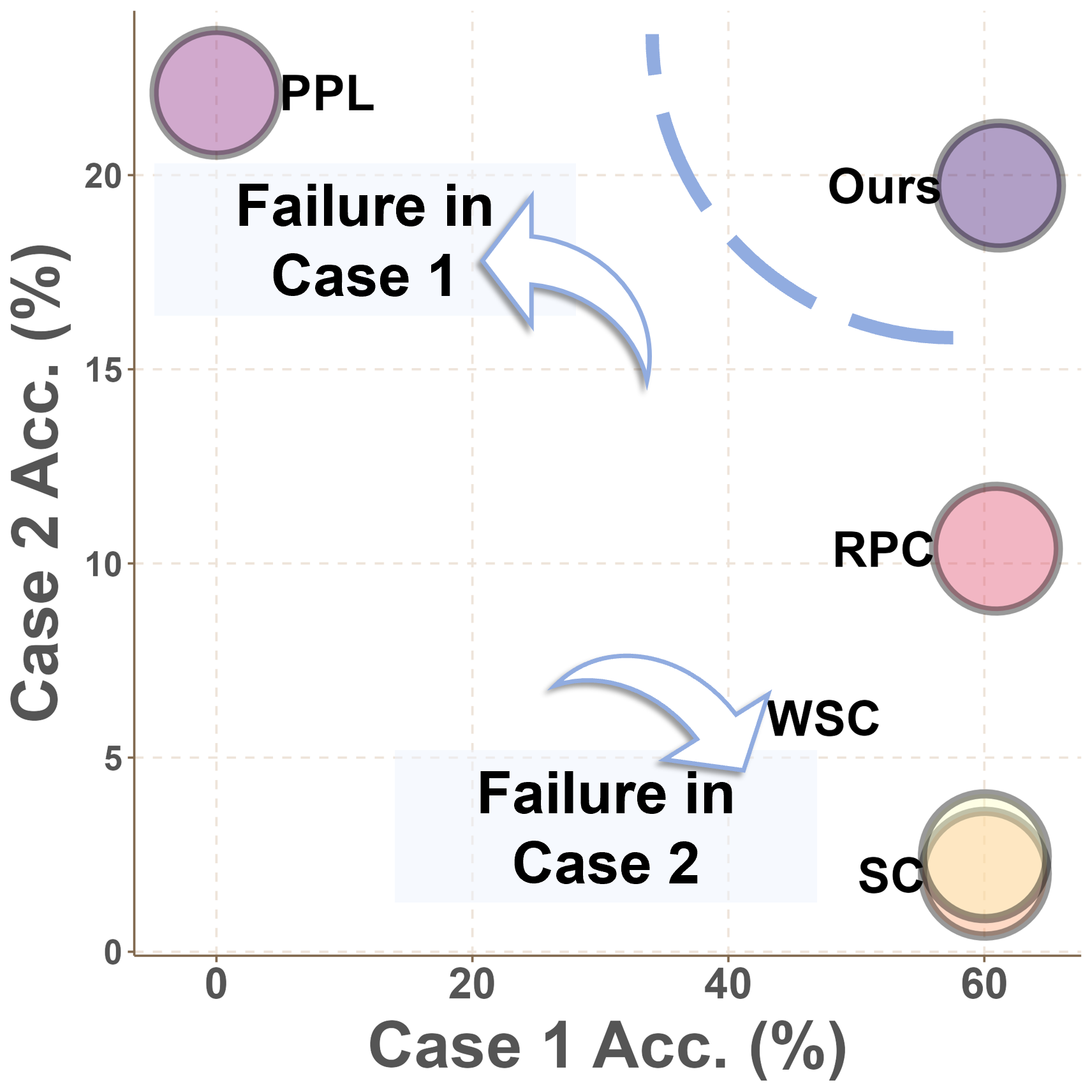}
}}
\end{minipage}
\hfill
\begin{minipage}[t]{0.235\textwidth}
\centering
\subfigure[MathOdyssey.]{{
\includegraphics[width=0.94\textwidth]{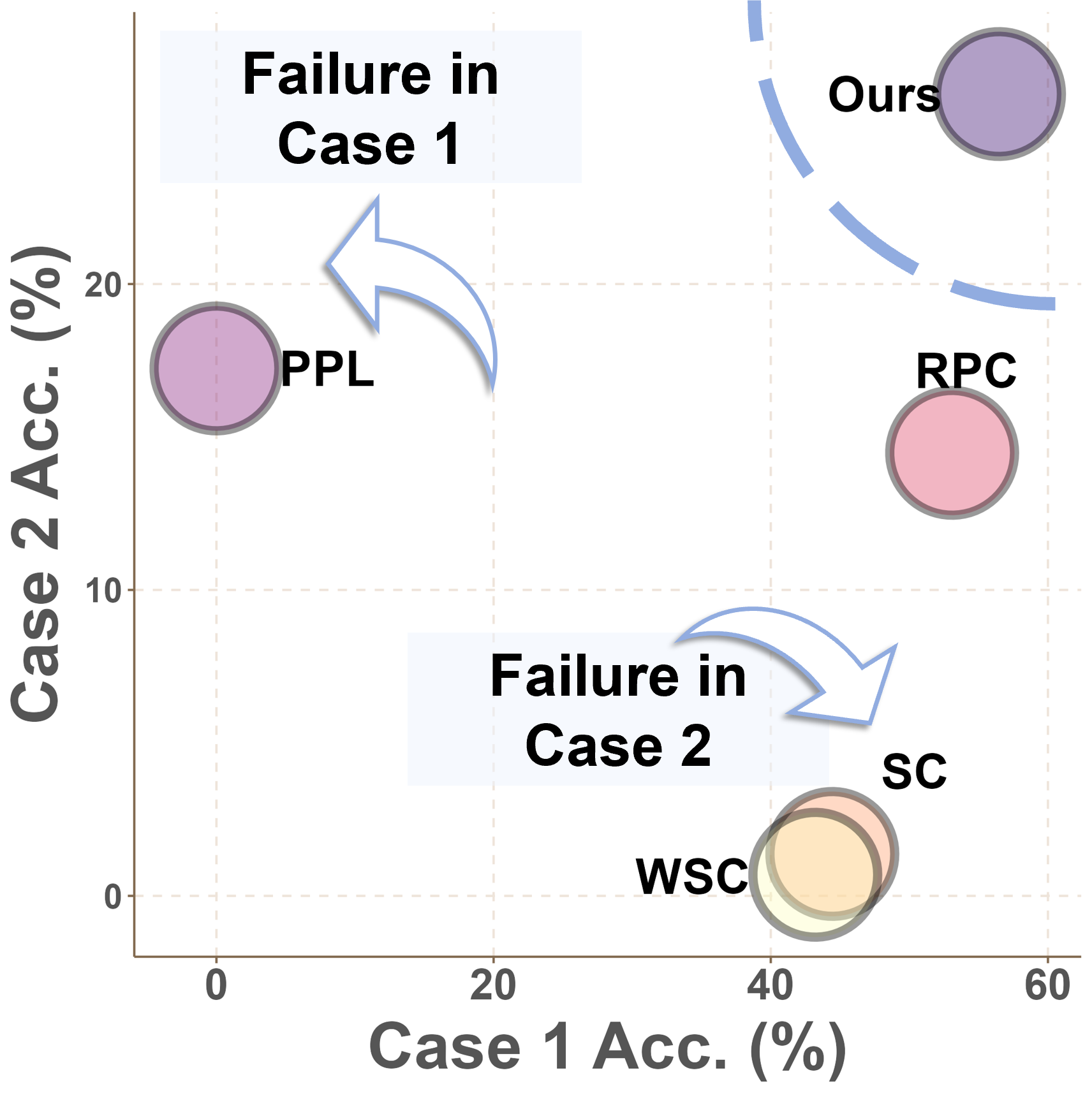}
}}
\end{minipage}
\caption{Complementary roles of voting support and pairwise confidence comparisons.
In Case I, an incorrect answer contains the highest-confidence path despite
weaker voting support. In Case II, an incorrect answer has stronger or tied
voting support, while pairwise confidence comparisons favor a correct answer.
The last two panels report the corresponding behavior on MATH and
MathOdyssey.}
\label{app:two_common_cases}
\end{figure*}

Let $c,w\in\mathcal A$ be two distinct observed answers, with $c$ correct
and $w$ incorrect. Each has at least one supporting path, so
$s_c^{\mathrm{SC}}$ and $s_w^{\mathrm{SC}}$ are positive.
In \textbf{Case I}, voting support favors $c$, whereas Best-of-$N$ favors
$w$ because an incorrect path has the higher confidence maximum:
\begin{equation}
    s_c^{\mathcal M}>s_w^{\mathcal M},
    \qquad
    s_w^{\mathrm{BoN}}>s_c^{\mathrm{BoN}}.
    \label{app:case_one}
\end{equation}

To describe \textbf{Case II}, define the empirical confidence preference
for $c$ over $w$:
\begin{equation}
    \widehat\rho_{cw}
    :=\frac{1}{s_c^{\mathrm{SC}}s_w^{\mathrm{SC}}}
      \sum_{i:a_i=c}\sum_{j:a_j=w}
      \left[\mathbf 1(p_i>p_j)+\frac12\mathbf 1(p_i=p_j)\right].
    \label{app:pairwise_confidence_preference}
\end{equation}
Here, $i$ and $j$ index observed paths supporting $c$ and $w$, respectively.
The denominator counts all such path pairs, so $\widehat\rho_{cw}$ is the
fraction favoring $c$, with confidence ties counted as one half.
Case II occurs when voting support favors or ties $w$, while these
pairwise comparisons favor $c$:
\begin{equation}
    s_w^{\mathcal M}\ge s_c^{\mathcal M},
    \qquad \widehat\rho_{cw}>\frac12.
    \label{app:case_two}
\end{equation}
The confidence-based comparison entry $Y_{cw}$ in
\Cref{eq:pairwise_flow} is the average signed comparison over the same
path pairs; hence $Y_{cw}=2\widehat\rho_{cw}-1$.
The SC-error analysis in \Cref{fig:sc_error_comparison} takes
$\mathcal M=\mathrm{SC}$, giving $s_w^{\mathrm{SC}}\ge s_c^{\mathrm{SC}}$
and $Y_{cw}>0$.

\paragraph{Conditioning on recoverable SC errors.}
We call an SC error \emph{recoverable} if its candidate pool contains at least
one correct answer, and denote the set of such pools by $\mathcal T_{\rm rec}$.
For each pool $t$, let $e_t\in[0,1]$ be its SC error weight, with fractional
credit used for ties.  We measure the prevalence of Case II among recoverable
SC errors by
\begin{equation}
    R_{\mathrm{II}}^{\mathrm{rec}}
    :=
    \frac{\sum_{t\in\mathcal T_{\rm rec}}e_t\,
          \mathbf 1\{\mathrm{Case\ II}_t\}}
         {\sum_{t\in\mathcal T_{\rm rec}}e_t}.
    \label{app:eq:recoverable_case_two_share}
\end{equation}
This conditioning excludes only pools with no correct candidate, for which
reranking over the observed answers is impossible.

\Cref{tab:recoverable_case_two_share} reports both the share of NoCorrect pools
among all SC errors and the Case-II share after excluding them. We use PPL
confidence, with $K=64$ on MATH and $K=128$ on MathOdyssey and AIME.

\begin{table}[H]
    \centering
    \caption{Breakdown of SC errors (\%). NoCorrect is measured over all SC
    errors; recoverable Case II is measured after excluding NoCorrect pools.
    Results use PPL confidence.}
    \label{tab:recoverable_case_two_share}
    \small
    \setlength{\tabcolsep}{4pt}
    \begin{tabular}{lccc}
        \toprule
        \textbf{Model} & \textbf{MATH} & \textbf{MathOdyssey} & \textbf{AIME} \\
        \midrule
        \multicolumn{4}{l}{\textit{NoCorrect / all SC errors}} \\
        InternLM2-Math-Plus-1.8B & $50.16$ & $46.94$ & $75.83$ \\
        InternLM2-Math-Plus-7B   & $48.58$ & $53.10$ & $63.70$ \\
        DeepSeek-Math-RL-7B      & $51.83$ & $59.42$ & $62.18$ \\
        \addlinespace[2pt]
        \multicolumn{4}{l}{\textit{Case II / recoverable SC errors}} \\
        InternLM2-Math-Plus-1.8B & $82.22$ & $83.81$ & $91.62$ \\
        InternLM2-Math-Plus-7B   & $79.30$ & $88.60$ & $88.62$ \\
        DeepSeek-Math-RL-7B      & $69.87$ & $81.42$ & $84.81$ \\
        \bottomrule
    \end{tabular}
\end{table}

These cases reveal complementary weaknesses: Best-of-$N$ is susceptible to
Case I because it relies on a single confidence maximum, whereas voting-based
methods are susceptible to Case II because they do not explicitly account for
answer-level pairwise confidence comparisons. Pairwise-Guided Logit Adjustment
($\mathsf{PGLA}$) combines the two
sources of evidence, using voting support to temper isolated confidence outliers
and comparative advantage to compensate for weaker voting support. This
combination yields a better empirical tradeoff between the two error types,
as shown in \Cref{fig:case_tradeoff,fig:case_tradeoff_full}.

\section{Online \texorpdfstring{$\mathsf{PGLA}$}{PGLA} Extension}
\label{app:online_pgla}

The full-budget formulation above applies $\mathsf{PGLA}$ after collecting all
sampled paths. To examine whether its comparative evidence can reduce
unnecessary inference-time sampling, we extend it to the \textbf{online
setting}. As new paths become available, $\mathsf{PGLA}$ updates its answer
prediction and stops sampling once the prediction becomes stable.

Let $t_j$ denote the number of observed paths at the $j$-th update, and let
$\mathcal D_j=\{(a_i,p_i)\}_{i=1}^{t_j}$ be this prefix of the evidence pool.
Its observed answers form $\mathcal A_j$, with supporting-path groups
$C_{u,j}=\{i\leq t_j:a_i=u\}$ for $u\in\mathcal A_j$.
From this prefix, we update the voting support $s_{u,j}^{\mathcal V}$ and
comparative advantage $s_{u,j}^{\mathrm{Pair}}$ using
\Cref{eq:pairwise_flow,eq:scpa_correction}, with all answer sums restricted
to $\mathcal A_j$. We set $s_{u,j}^{\mathrm{Pair}}=0$ when no other observed
answer has positive voting support, including the single-answer case.
The adjusted score is
\begin{equation}
    s_{u,j}^{\mathsf{PGLA}}
    =s_{u,j}^{\mathcal V}\exp(s_{u,j}^{\mathrm{Pair}}).
\end{equation}
For $s_{u,j}^{\mathcal V}>0$, this is equivalently
$\log s_{u,j}^{\mathsf{PGLA}}
=\log s_{u,j}^{\mathcal V}+s_{u,j}^{\mathrm{Pair}}$, as in
\Cref{eq:scpa_score}; the multiplicative form also applies to zero voting
support without adding a smoothing term.

Let $\hat a_j$ denote the unique highest-scoring answer in $\mathcal A_j$, with
$\hat a_j=\varnothing$ under a tie. Given a stability length $L$, sampling stops at
the first $j\geq L$ satisfying
\begin{equation}
    \hat a_{j-L+1}=\cdots=\hat a_j\neq\varnothing.
    \label{eq:online_PGLA_stopping}
\end{equation}
A tie resets the stability count, and sampling continues to $K_{\max}$ if the
condition is never met.
This label-free rule applies unchanged to a base aggregator and its
$\mathsf{PGLA}$-adjusted counterpart, enabling a matched online comparison.

\section{Proof for Proposition \ref{prop:variational_characterization}}
\label{app:variational_proof}

\begin{proof}
Consider an observed candidate pool with answer set $\mathcal A$ and
$U=|\mathcal A|\geq1$. The voting distribution $\bar{\mathbf s}^{\mathcal V}$
and comparative advantages $s_u^{\mathrm{Pair}}$ are determined by this pool
and are constants with respect to the optimization variable
$\bar{\mathbf s}$. They may be computed from the same paths; no statistical
independence between them is required. Under the proposition's positive-support
assumption,
\begin{equation}
    \bar{s}_u^{\mathcal V}
    =\frac{s_u^{\mathcal V}}{\sum_{v\in\mathcal A}s_v^{\mathcal V}}>0,
    \qquad
    \sum_{u\in\mathcal A}\bar{s}_u^{\mathcal V}=1,
    \label{app:eq:PGLA_positive_base}
\end{equation}
and each $s_u^{\mathrm{Pair}}\in[-1,1]$ is finite. For $U=1$, we take
$s_u^{\mathrm{Pair}}=0$ because there is no competing answer.

The objective in \Cref{eq:variational_scpa} is
\begin{align}
    \mathcal J(\bar{\mathbf s})
    &:=\sum_{u\in\mathcal A}\bar{s}_u s_u^{\mathrm{Pair}}
      -\sum_{u\in\mathcal A}\bar{s}_u
       \log\frac{\bar{s}_u}{\bar{s}_u^{\mathcal V}}\notag\\
    &=-\sum_{u\in\mathcal A}\bar{s}_u\log\bar{s}_u
      +\sum_{u\in\mathcal A}\bar{s}_u
       \bigl(s_u^{\mathrm{Pair}}+\log\bar{s}_u^{\mathcal V}\bigr),
    \label{app:eq:PGLA_variational_objective}
\end{align}
over the probability simplex
$\Delta_{\mathcal A}=\{\bar{\mathbf s}:\bar{s}_u\geq0,\quad
\sum_{u\in\mathcal A}\bar{s}_u=1\}$.
Terms with $\bar{s}_u=0$ are defined by continuity, so $0\log0=0$ and
$0\log(0/\bar{s}_u^{\mathcal V})=0$.

\textit{Existence and strict concavity.}
The function $x\mapsto x\log x$, with its continuous extension at zero,
is strictly convex on $[0,1]$. Thus the entropy term
$-\sum_{u\in\mathcal A}\bar{s}_u\log\bar{s}_u$ is strictly concave on the
simplex, while the second term in
\Cref{app:eq:PGLA_variational_objective} is linear in $\bar{\mathbf s}$.
Consequently, $\mathcal J$ is strictly concave for $U\geq2$.
It is also continuous on the compact simplex, so a maximizer exists and is
unique. For $U=1$, the simplex contains only the distribution assigning
probability one to its sole answer, and uniqueness is immediate.

\textit{Closed-form solution.}
Define the positive normalizing constant and a candidate distribution by
\begin{equation}
    Z:=\sum_{v\in\mathcal A}\bar{s}_v^{\mathcal V}
       \exp(s_v^{\mathrm{Pair}}),
    \qquad
    \bar{s}_u^\star:=\frac{\bar{s}_u^{\mathcal V}
       \exp(s_u^{\mathrm{Pair}})}{Z}.
    \label{app:eq:PGLA_variational_candidate}
\end{equation}
Since $\bar{s}_u^{\mathcal V}>0$ and $s_u^{\mathrm{Pair}}$ is finite,
$0<Z<\infty$, $\bar{s}_u^\star>0$, and
$\sum_{u\in\mathcal A}\bar{s}_u^\star=1$. Hence $\bar{\mathbf s}^\star$
is feasible. Moreover,
$\log\bar{s}_u^\star=\log\bar{s}_u^{\mathcal V}
+s_u^{\mathrm{Pair}}-\log Z$.
For every $\bar{\mathbf s}\in\Delta_{\mathcal A}$, including its boundary,
substitution gives
\begin{align}
    \mathcal J(\bar{\mathbf s})
    &=\sum_{u\in\mathcal A}\bar{s}_u
      \bigl(s_u^{\mathrm{Pair}}+\log\bar{s}_u^{\mathcal V}
      -\log\bar{s}_u\bigr)\notag\\
    &=\sum_{u\in\mathcal A}\bar{s}_u
      \bigl(\log Z+\log\bar{s}_u^\star-\log\bar{s}_u\bigr)\notag\\
    &=\log Z-D_{\mathrm{KL}}\!\left(
        \bar{\mathbf s}\,\|\,\bar{\mathbf s}^\star\right).
    \label{app:eq:PGLA_variational_gap}
\end{align}
The KL divergence is nonnegative and vanishes if and only if
$\bar{\mathbf s}=\bar{\mathbf s}^\star$. Therefore
$\mathcal J(\bar{\mathbf s})\leq\log Z$, with equality uniquely at
$\bar{\mathbf s}^\star$. This proves that
$\bar{\mathbf s}^{\mathsf{PGLA}}=\bar{\mathbf s}^\star$ and yields exactly
\Cref{eq:scpa_distribution}.
\end{proof}

\paragraph{Connection to the logit adjustment.}
Substituting \Cref{app:eq:PGLA_positive_base} into the optimizer gives
\begin{align}
    \bar{s}_u^{\mathsf{PGLA}}
    &=\frac{s_u^{\mathcal V}\exp(s_u^{\mathrm{Pair}})}
           {\sum_{v\in\mathcal A}s_v^{\mathcal V}
            \exp(s_v^{\mathrm{Pair}})}\notag\\
    &=\operatorname{softmax}\!\left(
       \bigl(\log s_v^{\mathcal V}+s_v^{\mathrm{Pair}}\bigr)_{v\in\mathcal A}
      \right)_u.
    \label{app:eq:PGLA_variational_logit}
\end{align}
Thus the adjusted distribution is obtained by normalizing the answer-level
scores $s_u^{\mathsf{PGLA}}=s_u^{\mathcal V}\exp(s_u^{\mathrm{Pair}})$.
For positive voting support, taking logarithms gives
$\log s_u^{\mathsf{PGLA}}=\log s_u^{\mathcal V}+s_u^{\mathrm{Pair}}$,
exactly as in \Cref{eq:scpa_score}.
If all comparative advantages are equal, their common exponential factor
cancels and $\bar{\mathbf s}^{\mathsf{PGLA}}=\bar{\mathbf s}^{\mathcal V}$.

\section{Additional Experimental Setup}
\label{app:additional_setup}

\subsection{Comparison Methods}
\label{app:comparison_methods}

\textbf{Competitors.} We compare 11 internal-confidence and voting baselines on the same sampled paths, together with three external-PRM baselines. We also include Online SC for the sampling-efficiency comparison; \textbf{+ Ours} denotes the corresponding voting rule enhanced by $\mathsf{PGLA}$.

\textbf{PPL Best-of-$N$.} This baseline selects the path with the highest mean token log-probability, equivalently the lowest perplexity~\citep{chen1998evaluation}. Its answer-level score is the maximum PPL confidence among paths supporting that answer.

\textbf{DeepConf Best-of-$N$.} DeepConf summarizes token confidence over local windows to identify low-confidence portions of a path~\citep{fu2025deepconf}. We use the average of the bottom-$10\%$ window confidences as the path score and select the highest-scoring path.

\textbf{Self-Certainty Best-of-$N$.} Self-Certainty measures the divergence of the model's next-token distribution from a uniform vocabulary distribution~\citep{kang2025selfcertainty}. This baseline selects the path with the highest average Self-Certainty score.

\textbf{SC.} Self-Consistency gives each sampled path one vote for its final answer~\citep{wang2022self}. It selects the answer with the largest voting support, $s_u^{\mathrm{SC}}=|C_u|$.

\textbf{PPL-Weighted SC.} This variant uses the inverse-perplexity confidence $p_i^{\mathrm{PPL}}$ in \Cref{eq:ppl_confidence} and sets $h_i=1$ in \Cref{eq:confidence_voting_score}. It is denoted by Weighted SC in the PPL-based experiments.

\textbf{DeepConf-Weighted SC.} This variant sums the DeepConf path scores over each answer group~\citep{fu2025deepconf}. It uses all eligible paths, with no additional path-level confidence filter.

\textbf{Self-Certainty-Weighted SC.} This variant sums the Self-Certainty path scores over each answer group~\citep{kang2025selfcertainty}. It retains the numerical confidence values as voting weights.

\textbf{Self-Certainty Borda.} This rule ranks paths by decreasing Self-Certainty and assigns rank $r$ the voting weight $(K-r+1)^{\gamma}$~\citep{kang2025selfcertainty}. We use $\gamma=1.2$ and sum the rank-based weights for each answer.

\textbf{DeepConf-Filter ($90\%$).} This offline rule retains paths at or above the tenth percentile of eligible path-confidence scores and sums their DeepConf weights for each answer~\citep{fu2025deepconf}. The top-$90\%$ path filter is separate from the bottom-$10\%$ window summary used to score each path; threshold ties are retained.

\textbf{Adaptive SC.} Adaptive Self-Consistency stops sampling according to the accumulated evidence for the leading answer~\citep{aggarwal2023adaptive}. We retain its original stopping criterion as a dynamic-budget baseline.

\textbf{RPC.} RPC combines Perplexity Consistency, which aggregates probability-derived support from distinct path--confidence pairs, with Reasoning Pruning, which filters low-confidence paths~\citep{zhou2025rpc}. It provides a confidence-aware voting baseline without an external scorer.

\textbf{PRM Best-of-$N$.} This baseline uses Qwen2.5-Math-PRM-7B to score the reasoning steps of each sampled path~\citep{zhang2025prm}. It multiplies the step scores into a path score and selects the highest-scoring path.

\textbf{PRM-Weighted SC.} This variant sums the same product of step-level PRM scores over each answer group. It combines answer agreement with externally estimated path quality~\citep{lightman2023verify,zhang2025prm}.

\textbf{PRM-RPC.} This variant substitutes the product of step-level PRM scores for the internal confidence weights in RPC. It retains RPC's path deduplication and pruning before aggregating the remaining voting support~\citep{zhou2025rpc,zhang2025prm}.

\textbf{Online SC.} This baseline updates the majority-vote prediction after every eight paths and stops after the unique winning answer remains unchanged for three consecutive updates. It uses the same stability rule as Online SC+$\mathsf{PGLA}$, as detailed in \Cref{app:online_sampling_efficiency}.

\subsection{LLM Backbones}

\textbf{LLM Backbones.} We use seven generators spanning the Qwen3, DeepSeekMath, and InternLM-Math families. Qwen3 models use reasoning mode unless a result is explicitly marked as non-thinking; the external PRM serves only as a scorer.

\textbf{Qwen3-1.7B.} This is the smallest Qwen3 generator in our experiments~\citep{yang2025qwen3}. It provides a small-model setting for evaluating aggregation across mathematical reasoning and code generation.

\textbf{Qwen3-4B.} This Qwen3 model increases the parameter scale to 4B while retaining the family's thinking and non-thinking modes~\citep{yang2025qwen3}. We include it in the confidence-transfer comparisons and complementary non-thinking evaluations.

\textbf{Qwen3-8B.} This 8B Qwen3 model extends the reasoning-mode comparisons to a larger backbone~\citep{yang2025qwen3}. Together with the 1.7B and 4B models, it supports comparisons across model scales within the same family.

\textbf{Qwen3-32B.} This is the largest Qwen3 generator evaluated in our study~\citep{yang2025qwen3}. We report its reasoning-mode results on AIME 2024, AIME 2025, and HMMT 2025 in \Cref{app:qwen32_extension}.

\textbf{DeepSeek-Math-RL-7B.} This math-specialized 7B model is the reinforcement-learning variant of DeepSeekMath~\citep{shao2024deepseekmath}. It tests aggregation on a generator trained specifically for mathematical reasoning.

\textbf{InternLM2-Math-Plus-1.8B.} This is the smallest math-specialized generator in our evaluation~\citep{ying2024internlmmath}. It provides a lower-capacity counterpart for the mathematical-reasoning and online sampling experiments.

\textbf{InternLM2-Math-Plus-7B.} This 7B model belongs to the same math-specialized family as the 1.8B variant~\citep{ying2024internlmmath}. It is included in the main mathematical-reasoning comparisons and the detailed aggregation analyses.

\textbf{Qwen2.5-Math-PRM-7B.} This process reward model assigns scores to individual reasoning steps~\citep{zhang2025prm}. We use it as an external scorer for the three PRM baselines and their applicable $\mathsf{PGLA}$ variants, separately from the seven generators above.

\subsection{Datasets}

\textbf{Datasets.} The ten benchmarks in the main experimental setup cover competition mathematics, graduate-level science, and code generation. HumanEval+ and MBPP+ provide additional evaluations with strengthened tests for the corresponding code benchmarks.

\textbf{AIME 2024.} This benchmark contains problems from the 2024 American Invitational Mathematics Examination~\citep{aimo2024aime}. It evaluates short-answer mathematical reasoning and is reported separately from the historical AIME collection.

\textbf{AIME 2025.} This benchmark contains the 2025 AIME problems released by OpenCompass~\citep{opencompass2025aime}. It provides a separate competition-year evaluation under the same short-answer format.

\textbf{HMMT 2025.} We use the February 2025 Harvard--MIT Mathematics Tournament problems provided by MathArena~\citep{balunovic2025matharena}. The benchmark covers challenging competition mathematics, including algebra, geometry, combinatorics, and number theory.

\textbf{AIME.} The historical AIME collection contains 993 problems from the American Invitational Mathematics Examination from 1983 to 2024~\citep{zamil2024aime}. This larger short-answer benchmark is used in the math-specialized-model experiments.

\textbf{MATH.} MATH contains challenging competition mathematics problems spanning algebra, geometry, number theory, counting, and combinatorics~\citep{hendrycks2021math}. We use its 5,000-problem test split.

\textbf{MathOdyssey.} MathOdyssey contains 387 problems covering advanced high-school, university-level, and olympiad-level mathematics~\citep{fang2024mathodyssey}. It extends the evaluation beyond a single competition or curriculum level.

\textbf{GPQA-Diamond (GPQA-D).} GPQA comprises expert-written multiple-choice questions in biology, physics, and chemistry~\citep{rein2024gpqa}. We use its 198-question Diamond subset to evaluate graduate-level scientific reasoning.

\textbf{HumanEval.} HumanEval contains 164 Python programming tasks specified by function signatures and natural-language descriptions~\citep{chen2021evaluatingcode}. We evaluate the selected programs using the original-test outcomes in our evaluation pipeline.

\textbf{MBPP.} Mostly Basic Programming Problems consists of short Python programming tasks with natural-language descriptions and tests~\citep{austin2021program}. Our code-transfer evaluations use the cached 378-problem split.

\textbf{LiveCodeBench V6.} LiveCodeBench collects programming-contest problems from LeetCode, AtCoder, and Codeforces~\citep{jain2024livecodebench}. We use the V6 code-generation benchmark to evaluate the correctness of selected programs by execution.

\textbf{HumanEval+ and MBPP+.} EvalPlus augments HumanEval and MBPP with additional tests that detect errors missed by the original suites~\citep{liu2023evalplus}. We retain the sampled programs and problem sets and report results under the strengthened tests separately from the original-test results.


\subsection{Implementation Details}

\textbf{Sampling and consistency.} For each query, all methods use matched
$K$-path subsets of the same sampled pool, generated with common decoding
parameters. The consistency
function treats two paths as consistent in mathematical and knowledge
reasoning when their normalized final answers match. For code generation, we
follow RPC and define consistency through execution behavior~\citep{zhou2025rpc}.
For the math-specialized backbones, we set
$K=64$ on MATH and $K=128$ on MathOdyssey and AIME. For Qwen3 comparisons
in reasoning and non-thinking modes, we
set $K=32$ on AIME 2024, AIME 2025, HMMT 2025, GPQA-D, HumanEval, MBPP, and
LiveCodeBench V6. We use a temperature of $1.0$ and top-$p$ of $0.95$. The maximum
generation length is 2,048 tokens for MATH, MathOdyssey, and AIME, and 16,384
tokens for the remaining datasets. Within each experimental setting, all methods are evaluated on the same problem set and sampled-path subsets. The same repetitions are used for each baseline and its PGLA-enhanced counterpart.We report the mean and standard deviation. All experiments are conducted on Linux servers equipped
with NVIDIA A100 and RTX 3090 GPUs, except for the CPU-only aggregation
microbenchmark in \Cref{app:computational_overhead}.

\begin{figure}[H]
    \centering
    \includegraphics[width=0.6\linewidth]{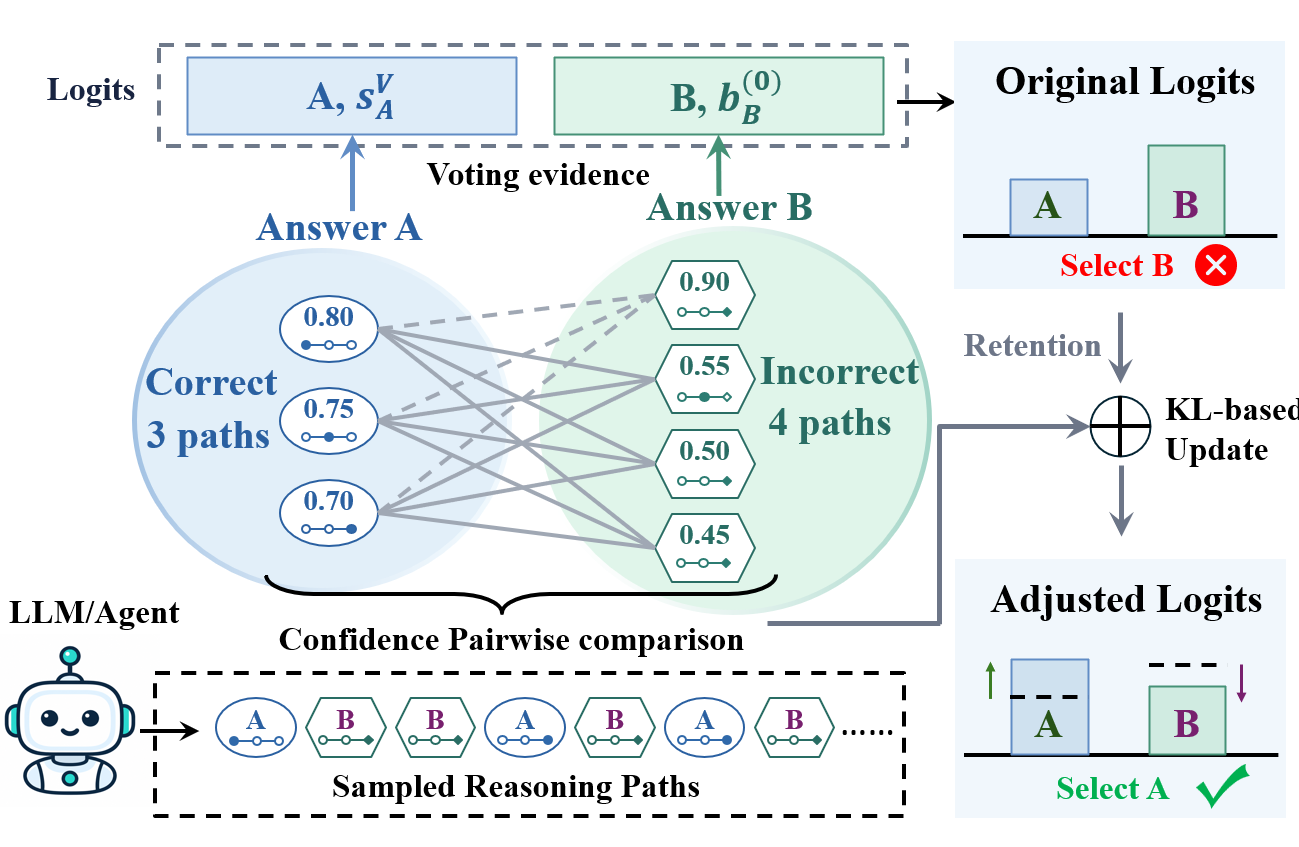}
    \caption{Illustration of $\mathsf{PGLA}$.}
    \label{fig:pga_pipeline}
\end{figure}

\textbf{Aggregation.} For a voting rule $\mathcal V$, $s_u^{\mathcal V}$
denotes its answer-level voting support and $s_u^{\mathrm{Pair}}$ denotes
the voting-weighted comparative advantage in \Cref{eq:scpa_correction}.
The default adjustment is
$\log s_u^{\mathsf{PGLA}}=\log s_u^{\mathcal V}+s_u^{\mathrm{Pair}}$
from \Cref{eq:scpa_score}.  A illustrative example of the adjustment is given in \Cref{fig:pga_pipeline}.  

\section{Additional Experimental Results}
\label{app:additional_results}

\subsection{Enhanced Code Tests in Reasoning Mode}
\label{app:enhanced_code_transfer}

\Cref{tab:confidence_transfer_code_plus} reports the strengthened
HumanEval+ and MBPP+ tests for the same sampled pools used in
\Cref{tab:reasoning}. A prediction must pass both the
original and enhanced tests. $\mathsf{PGLA}$ generally improves accuracy
under these strengthened tests, suggesting that its benefits persist
under stricter verification.

\begin{table*}[p]
    \centering
    \caption{Enhanced code-test accuracy (\%) in reasoning mode ($K=32$). Here \textbf{+ Ours} denotes a
 $\mathsf{PGLA}$-enhanced variant.}
    \label{tab:confidence_transfer_code_plus}
    \footnotesize
    \renewcommand{\arraystretch}{1.08}
    \setlength{\tabcolsep}{4pt}
    \begin{tabular*}{\textwidth}{@{\extracolsep{\fill}}lcccc@{}}
        \toprule
        & \multicolumn{2}{c}{\textbf{Qwen3-1.7B Reasoning}}
        & \multicolumn{2}{c}{\textbf{Qwen3-4B Reasoning}} \\
        \cmidrule(lr){2-3}\cmidrule(lr){4-5}
        \textbf{Methods} & \textbf{HumanEval+} & \textbf{MBPP+}
        & \textbf{HumanEval+} & \textbf{MBPP+} \\
        \midrule
        PPL Best-of-$N$ & $73.48_{\pm 1.80}$ & $64.55_{\pm 0.99}$ & $86.99_{\pm 0.55}$ & $77.59_{\pm 0.74}$ \\
        DeepConf Best-of-$N$ & $77.44_{\pm 2.36}$ & $66.40_{\pm 0.76}$ & $90.04_{\pm 0.99}$ & $78.89_{\pm 0.76}$ \\
        Self-Certainty Best-of-$N$ & $78.05_{\pm 2.67}$ & $67.46_{\pm 0.97}$ & $90.24_{\pm 1.06}$ & $78.99_{\pm 0.62}$ \\
        \addlinespace[1.5pt]
        SC & $77.29_{\pm 1.14}$ & $71.16_{\pm 0.83}$ & $88.01_{\pm 1.16}$ & $80.58_{\pm 0.62}$ \\
        \rowcolor{gray!10} \textbf{SC + Ours (DeepConf)} & $81.10_{\pm 2.46}$ & $71.43_{\pm 0.53}$ & $90.04_{\pm 1.14}$ & $80.79_{\pm 0.46}$ \\
        \rowcolor{gray!10} \textbf{SC + Ours (Self-Certainty)} & $\mathbf{81.25}_{\pm 1.91}$ & $\mathbf{71.96}_{\pm 0.37}$ & $\mathbf{90.65}_{\pm 1.36}$ & $\mathbf{80.85}_{\pm 0.41}$ \\
        \addlinespace[1.5pt]
        DeepConf-Weighted SC & $78.51_{\pm 1.64}$ & $70.37_{\pm 0.85}$ & $89.02_{\pm 0.85}$ & $\mathbf{80.77}_{\pm 0.46}$ \\
        \rowcolor{gray!10} \textbf{DeepConf-Weighted SC + Ours} & $\mathbf{81.10}_{\pm 2.38}$ & $\mathbf{71.69}_{\pm 0.46}$ & $\mathbf{90.24}_{\pm 0.92}$ & $\mathbf{80.77}_{\pm 0.46}$ \\
        \addlinespace[1.5pt]
        Self-Certainty-Weighted SC & $77.59_{\pm 1.54}$ & $70.90_{\pm 0.82}$ & $88.82_{\pm 0.78}$ & $80.85_{\pm 0.41}$ \\
        \rowcolor{gray!10} \textbf{Self-Certainty-Weighted SC + Ours} & $\mathbf{81.25}_{\pm 1.95}$ & $\mathbf{71.96}_{\pm 0.45}$ & $\mathbf{90.65}_{\pm 1.22}$ & $\mathbf{80.87}_{\pm 0.41}$ \\
        \addlinespace[1.5pt]
        Self-Certainty Borda & $80.18_{\pm 1.78}$ & $71.43_{\pm 0.66}$ & $89.02_{\pm 0.99}$ & $\mathbf{80.32}_{\pm 0.50}$ \\
        \rowcolor{gray!10} \textbf{Self-Certainty Borda + Ours} & $\mathbf{81.40}_{\pm 1.98}$ & $\mathbf{72.22}_{\pm 0.54}$ & $\mathbf{90.65}_{\pm 1.09}$ & $\mathbf{80.32}_{\pm 0.50}$ \\
        \addlinespace[1.5pt]
        DeepConf-Filter (90\%) & $80.03_{\pm 1.89}$ & $70.63_{\pm 0.73}$ & $89.02_{\pm 0.87}$ & $80.53_{\pm 0.46}$ \\
        \rowcolor{gray!10} \textbf{DeepConf-Filter (90\%) + Ours} & $\mathbf{81.10}_{\pm 2.49}$ & $\mathbf{71.69}_{\pm 0.44}$ & $\mathbf{90.24}_{\pm 0.95}$ & $\mathbf{80.58}_{\pm 0.45}$ \\
        \bottomrule
    \end{tabular*}
\end{table*}

\subsection{Comparison of Non-Thinking and Reasoning Modes}
\label{app:full_confidence_transfer}

\Cref{tab:confidence_transfer_full_17b,tab:confidence_transfer_full_4b}
compare non-thinking and reasoning modes of Qwen3-1.7B and Qwen3-4B on
AIME 2024, AIME 2025, and HMMT 2025. Reasoning-mode results are reproduced
from \Cref{tab:reasoning} for the methods shared by both modes. All columns
use $K=32$. Reasoning mode consistently achieves higher accuracy, and
$\mathsf{PGLA}$ generally improves the voting baselines in both modes.

\begin{table*}[p]
    \centering
    \caption{Comparison of non-thinking and reasoning modes for Qwen3-1.7B on three mathematical reasoning
benchmarks ($K=32$).}
    \label{tab:confidence_transfer_full_17b}
    \footnotesize
    \renewcommand{\arraystretch}{1.08}
    \setlength{\tabcolsep}{3.5pt}
    \resizebox{\textwidth}{!}{%
    \begin{tabular}{lcccccc}
        \toprule
        \multicolumn{7}{c}{\textbf{Qwen3-1.7B}} \\
        \midrule
        \multirow{2}{*}{\textbf{Methods}} & \multicolumn{3}{c}{\textbf{Non-thinking mode}} & \multicolumn{3}{c}{\textbf{Reasoning mode}} \\
        \cmidrule(lr){2-4}\cmidrule(lr){5-7}
        & \textbf{AIME 2024} & \textbf{AIME 2025} & \textbf{HMMT 2025} & \textbf{AIME 2024} & \textbf{AIME 2025} & \textbf{HMMT 2025} \\
        \midrule
        PPL Best-of-$N$ & $12.00_{\pm 2.21}$ & $11.33_{\pm 3.40}$ & $5.00_{\pm 3.42}$ & $36.00_{\pm 4.42}$ & $34.00_{\pm 2.91}$ & $23.00_{\pm 3.48}$ \\
        DeepConf Best-of-$N$ & $12.33_{\pm 2.13}$ & $14.33_{\pm 2.60}$ & $8.33_{\pm 3.73}$ & $62.00_{\pm 4.52}$ & $42.67_{\pm 2.49}$ & $28.33_{\pm 3.73}$ \\
        Self-Certainty Best-of-$N$ & $12.67_{\pm 2.49}$ & $15.00_{\pm 3.42}$ & $9.67_{\pm 2.33}$ & $60.67_{\pm 4.16}$ & $46.67_{\pm 3.33}$ & $30.67_{\pm 4.42}$ \\
        \addlinespace[1.5pt]
        SC & $15.17_{\pm 1.55}$ & $14.83_{\pm 1.57}$ & $6.94_{\pm 0.57}$ & $69.50_{\pm 1.98}$ & $47.33_{\pm 3.67}$ & $27.00_{\pm 1.00}$ \\
        \rowcolor{gray!10} \textbf{SC + Ours (DeepConf)} & $16.00_{\pm 2.49}$ & $\mathbf{16.67}_{\pm 0.00}$ & $\mathbf{7.67}_{\pm 1.53}$ & $\mathbf{75.00}_{\pm 1.67}$ & $\mathbf{51.33}_{\pm 2.67}$ & $30.00_{\pm 2.98}$ \\
        \rowcolor{gray!10} \textbf{SC + Ours (Self-Certainty)} & $\mathbf{16.50}_{\pm 2.17}$ & $\mathbf{16.67}_{\pm 0.00}$ & $\mathbf{7.67}_{\pm 1.53}$ & $74.00_{\pm 1.33}$ & $50.67_{\pm 3.27}$ & $\mathbf{30.33}_{\pm 2.77}$ \\
        \addlinespace[1.5pt]
        DeepConf-Weighted SC & $\mathbf{15.67}_{\pm 2.13}$ & $15.67_{\pm 1.53}$ & $7.33_{\pm 1.33}$ & $70.00_{\pm 1.49}$ & $48.33_{\pm 4.28}$ & $27.00_{\pm 1.00}$ \\
        \rowcolor{gray!10} \textbf{DeepConf-Weighted SC + Ours} & $\mathbf{15.67}_{\pm 2.13}$ & $\mathbf{16.67}_{\pm 0.00}$ & $\mathbf{7.67}_{\pm 1.53}$ & $\mathbf{75.00}_{\pm 1.67}$ & $\mathbf{51.00}_{\pm 3.00}$ & $\mathbf{31.33}_{\pm 3.40}$ \\
        \addlinespace[1.5pt]
        Self-Certainty-Weighted SC & $15.67_{\pm 2.13}$ & $15.00_{\pm 1.67}$ & $7.33_{\pm 1.33}$ & $70.33_{\pm 1.80}$ & $48.67_{\pm 4.27}$ & $27.00_{\pm 1.00}$ \\
        \rowcolor{gray!10} \textbf{Self-Certainty-Weighted SC + Ours} & $\mathbf{16.00}_{\pm 2.49}$ & $\mathbf{16.67}_{\pm 0.00}$ & $\mathbf{7.67}_{\pm 1.53}$ & $\mathbf{74.00}_{\pm 1.33}$ & $\mathbf{50.67}_{\pm 3.27}$ & $\mathbf{31.00}_{\pm 3.35}$ \\
        \bottomrule
    \end{tabular}}
\end{table*}

\begin{table*}[p]
    \centering
    \caption{Comparison of non-thinking and reasoning modes for Qwen3-4B on three mathematical reasoning
benchmarks ($K=32$).}
    \label{tab:confidence_transfer_full_4b}
    \footnotesize
    \renewcommand{\arraystretch}{1.08}
    \setlength{\tabcolsep}{3.5pt}
    \resizebox{\textwidth}{!}{%
    \begin{tabular}{lcccccc}
        \toprule
        \multicolumn{7}{c}{\textbf{Qwen3-4B}} \\
        \midrule
        \multirow{2}{*}{\textbf{Methods}} & \multicolumn{3}{c}{\textbf{Non-thinking mode}} & \multicolumn{3}{c}{\textbf{Reasoning mode}} \\
        \cmidrule(lr){2-4}\cmidrule(lr){5-7}
        & \textbf{AIME 2024} & \textbf{AIME 2025} & \textbf{HMMT 2025} & \textbf{AIME 2024} & \textbf{AIME 2025} & \textbf{HMMT 2025} \\
        \midrule
        PPL Best-of-$N$ & $30.00_{\pm 3.94}$ & $24.14_{\pm 2.82}$ & $13.00_{\pm 2.77}$ & $57.00_{\pm 2.33}$ & $64.33_{\pm 3.96}$ & $40.33_{\pm 4.33}$ \\
        DeepConf Best-of-$N$ & $31.67_{\pm 4.77}$ & $25.29_{\pm 1.63}$ & $15.33_{\pm 3.06}$ & $77.00_{\pm 2.33}$ & $76.00_{\pm 3.89}$ & $50.00_{\pm 4.94}$ \\
        Self-Certainty Best-of-$N$ & $32.33_{\pm 3.96}$ & $26.44_{\pm 3.25}$ & $14.33_{\pm 2.60}$ & $81.67_{\pm 2.24}$ & $71.67_{\pm 3.07}$ & $51.67_{\pm 4.01}$ \\
        \addlinespace[1.5pt]
        SC & $39.56_{\pm 4.24}$ & $26.25_{\pm 1.51}$ & $15.67_{\pm 1.53}$ & $80.00_{\pm 0.00}$ & $76.33_{\pm 0.67}$ & $51.50_{\pm 1.57}$ \\
        \rowcolor{gray!10} \textbf{SC + Ours (DeepConf)} & $\mathbf{41.83}_{\pm 3.69}$ & $\mathbf{28.74}_{\pm 1.63}$ & $\mathbf{16.67}_{\pm 0.00}$ & $\mathbf{83.00}_{\pm 1.00}$ & $76.33_{\pm 1.80}$ & $54.00_{\pm 2.91}$ \\
        \rowcolor{gray!10} \textbf{SC + Ours (Self-Certainty)} & $41.33_{\pm 3.40}$ & $\mathbf{28.74}_{\pm 1.63}$ & $\mathbf{16.67}_{\pm 0.00}$ & $82.00_{\pm 1.63}$ & $\mathbf{78.67}_{\pm 1.63}$ & $\mathbf{55.67}_{\pm 1.53}$ \\
        \addlinespace[1.5pt]
        DeepConf-Weighted SC & $41.00_{\pm 4.23}$ & $26.44_{\pm 1.63}$ & $15.67_{\pm 1.53}$ & $80.00_{\pm 0.00}$ & $\mathbf{76.33}_{\pm 1.00}$ & $51.33_{\pm 1.63}$ \\
        \rowcolor{gray!10} \textbf{DeepConf-Weighted SC + Ours} & $\mathbf{42.33}_{\pm 3.96}$ & $\mathbf{27.59}_{\pm 2.82}$ & $\mathbf{16.67}_{\pm 0.00}$ & $\mathbf{83.33}_{\pm 1.49}$ & $\mathbf{76.33}_{\pm 2.33}$ & $\mathbf{54.00}_{\pm 2.91}$ \\
        \addlinespace[1.5pt]
        Self-Certainty-Weighted SC & $40.33_{\pm 4.07}$ & $26.44_{\pm 1.63}$ & $15.67_{\pm 1.53}$ & $80.00_{\pm 0.00}$ & $76.33_{\pm 1.00}$ & $52.67_{\pm 2.00}$ \\
        \rowcolor{gray!10} \textbf{Self-Certainty-Weighted SC + Ours} & $\mathbf{41.67}_{\pm 3.07}$ & $\mathbf{27.59}_{\pm 2.82}$ & $\mathbf{16.67}_{\pm 0.00}$ & $\mathbf{82.00}_{\pm 1.63}$ & $\mathbf{79.33}_{\pm 1.33}$ & $\mathbf{55.33}_{\pm 1.63}$ \\
        \bottomrule
    \end{tabular}}
\end{table*}

\clearpage

\subsection{Performance with the External Reward Setting}

\Cref{tab:aggregation_performance} compares the methods using
product-aggregated rewards from an external process reward model. We observe
that our method improves PRM-Weighted SC and PRM-RPC in most settings. To be specific, at the full budget, \(\mathsf{PGLA}\) raises
their average accuracy from 33.98 to 35.51 and from 34.30 to 35.55,
respectively. This further demonstrates that our method can effectively use
externally provided reward signals.

\begin{table*}[t]
    \centering
    \caption{Accuracy comparison under the sampling budgets of $\frac{K}{2}$ and $K$, using
product-aggregated rewards from an external process reward model Qwen2.5-Math-PRM-7B ~\citep{lightman2023verify,zhang2025prm}. \textbf{+ Ours} denotes the
$\mathsf{PGLA}$-enhanced variant of each voting baseline.}
    \label{tab:aggregation_performance}
    \fontsize{7.3}{6.9}\selectfont
    \renewcommand{\arraystretch}{0.99}
    \setlength{\tabcolsep}{2pt}
    \begin{tabular}{@{\hspace{2pt}}p{\dimexpr\textwidth-324pt\relax}*{6}{>{\centering\arraybackslash}p{40.5pt}}*{2}{>{\centering\arraybackslash}p{22.5pt}}@{\hspace{2pt}}}
        \toprule
        \multirow{2}{*}{\textbf{Method}} & \multicolumn{2}{c}{\textbf{MATH}} & \multicolumn{2}{c}{\textbf{MathOdyssey}} & \multicolumn{2}{c}{\textbf{AIME}} & \multicolumn{2}{c}{\textbf{Average}} \\
        \cmidrule(lr){2-3} \cmidrule(lr){4-5} \cmidrule(lr){6-7} \cmidrule(lr){8-9}
        & $K/2$ $\uparrow$ & $K$ $\uparrow$ & $K/2$ $\uparrow$ & $K$ $\uparrow$ & $K/2$ $\uparrow$ & $K$ $\uparrow$ & $K/2$ $\uparrow$ & $K$ $\uparrow$ \\
        \addlinespace[2pt]
        \rowcolor{gray!14} \multicolumn{9}{l}{\strut\hspace{1pt}\textbf{InternLM2-Math-Plus-1.8B}} \\
        \addlinespace[1pt]
        PRM Best-of-$N$ & $46.50_{\pm 0.33}$ & $48.48_{\pm 0.21}$ & $24.75_{\pm 1.44}$ & $25.47_{\pm 1.24}$ & $4.43_{\pm 0.24}$ & $4.93_{\pm 0.34}$ & $25.23$ & $26.30$ \\
        \addlinespace[1pt]
        PRM-Weighted SC & $45.64_{\pm 0.26}$ & $47.39_{\pm 0.23}$ & $23.24_{\pm 1.04}$ & $24.91_{\pm 1.07}$ & $4.63_{\pm 0.25}$ & $5.02_{\pm 0.23}$ & $24.50$ & $25.77$ \\
        \rowcolor{gray!5} \hspace{7pt}\textbf{+ Ours} & $\mathbf{46.88}_{\pm 0.29}$ & $\mathbf{49.09}_{\pm 0.20}$ & $\mathbf{26.99}_{\pm 1.16}$ & $\mathbf{28.95}_{\pm 1.26}$ & $\mathbf{4.73}_{\pm 0.33}$ & $\mathbf{5.17}_{\pm 0.27}$ & $\mathbf{26.20}$ & $\mathbf{27.73}$ \\
        \addlinespace[1pt]
        PRM-RPC & $45.77_{\pm 0.25}$ & $47.61_{\pm 0.23}$ & $23.60_{\pm 1.19}$ & $25.32_{\pm 1.04}$ & $4.66_{\pm 0.24}$ & $5.07_{\pm 0.21}$ & $24.68$ & $26.00$ \\
        \rowcolor{gray!5} \hspace{7pt}\textbf{+ Ours} & $\mathbf{46.88}_{\pm 0.28}$ & $\mathbf{49.18}_{\pm 0.17}$ & $\mathbf{27.02}_{\pm 1.28}$ & $\mathbf{29.05}_{\pm 1.30}$ & $\mathbf{4.73}_{\pm 0.32}$ & $\mathbf{5.19}_{\pm 0.26}$ & $\mathbf{26.21}$ & $\mathbf{27.80}$ \\
        \addlinespace[2pt]
        \rowcolor{gray!14} \multicolumn{9}{l}{\strut\hspace{1pt}\textbf{InternLM2-Math-Plus-7B}} \\
        \addlinespace[1pt]
        PRM Best-of-$N$ & $58.22_{\pm 0.34}$ & $59.34_{\pm 0.32}$ & $32.85_{\pm 1.78}$ & $33.07_{\pm 0.99}$ & $15.10_{\pm 0.50}$ & $15.96_{\pm 0.36}$ & $35.39$ & $36.12$ \\
        \addlinespace[1pt]
        PRM-Weighted SC & $57.56_{\pm 0.31}$ & $58.67_{\pm 0.10}$ & $33.57_{\pm 1.13}$ & $34.34_{\pm 0.58}$ & $15.49_{\pm 0.36}$ & $16.23_{\pm 0.48}$ & $35.54$ & $36.42$ \\
        \rowcolor{gray!5} \hspace{7pt}\textbf{+ Ours} & $\mathbf{58.86}_{\pm 0.29}$ & $\mathbf{60.40}_{\pm 0.24}$ & $\mathbf{37.63}_{\pm 1.18}$ & $\mathbf{38.79}_{\pm 0.62}$ & $\mathbf{15.55}_{\pm 0.42}$ & $\mathbf{16.57}_{\pm 0.35}$ & $\mathbf{37.35}$ & $\mathbf{38.59}$ \\
        \addlinespace[1pt]
        PRM-RPC & $57.74_{\pm 0.28}$ & $58.90_{\pm 0.13}$ & $34.83_{\pm 1.18}$ & $35.48_{\pm 0.40}$ & $\mathbf{15.57}_{\pm 0.42}$ & $16.31_{\pm 0.44}$ & $36.05$ & $36.90$ \\
        \rowcolor{gray!5} \hspace{7pt}\textbf{+ Ours} & $\mathbf{58.88}_{\pm 0.24}$ & $\mathbf{60.41}_{\pm 0.25}$ & $\mathbf{37.63}_{\pm 1.24}$ & $\mathbf{38.87}_{\pm 0.60}$ & $15.53_{\pm 0.37}$ & $\mathbf{16.50}_{\pm 0.28}$ & $\mathbf{37.35}$ & $\mathbf{38.59}$ \\
        \addlinespace[2pt]
        \rowcolor{gray!14} \multicolumn{9}{l}{\strut\hspace{1pt}\textbf{Deepseek-Math-RL-7B}} \\
        \addlinespace[1pt]
        PRM Best-of-$N$ & $61.27_{\pm 0.24}$ & $62.79_{\pm 0.16}$ & $34.34_{\pm 1.27}$ & $34.31_{\pm 0.66}$ & $15.93_{\pm 0.23}$ & $17.14_{\pm 0.40}$ & $37.18$ & $38.08$ \\
        \addlinespace[1pt]
        PRM-Weighted SC & $60.76_{\pm 0.25}$ & $62.04_{\pm 0.20}$ & $39.69_{\pm 0.59}$ & $40.28_{\pm 0.53}$ & $15.77_{\pm 0.27}$ & $16.93_{\pm 0.43}$ & $38.74$ & $39.75$ \\
        \rowcolor{gray!5} \hspace{7pt}\textbf{+ Ours} & $\mathbf{61.17}_{\pm 0.24}$ & $\mathbf{62.41}_{\pm 0.20}$ & $\mathbf{40.18}_{\pm 0.64}$ & $\mathbf{40.69}_{\pm 0.53}$ & $\mathbf{16.13}_{\pm 0.48}$ & $\mathbf{17.49}_{\pm 0.57}$ & $\mathbf{39.16}$ & $\mathbf{40.20}$ \\
        \addlinespace[1pt]
        PRM-RPC & $60.91_{\pm 0.27}$ & $62.26_{\pm 0.22}$ & $\mathbf{40.05}_{\pm 0.83}$ & $40.67_{\pm 0.51}$ & $15.85_{\pm 0.32}$ & $17.05_{\pm 0.36}$ & $38.94$ & $39.99$ \\
        \rowcolor{gray!5} \hspace{7pt}\textbf{+ Ours} & $\mathbf{61.19}_{\pm 0.23}$ & $\mathbf{62.55}_{\pm 0.19}$ & $39.85_{\pm 0.75}$ & $\mathbf{40.69}_{\pm 0.42}$ & $\mathbf{16.14}_{\pm 0.51}$ & $\mathbf{17.52}_{\pm 0.66}$ & $\mathbf{39.06}$ & $\mathbf{40.26}$ \\
        \addlinespace[1pt]
        \bottomrule
    \end{tabular}
\end{table*}

\clearpage
\subsection{Tradeoff between Case I and Case II}

\begin{figure}[!ht]
    \centering
    \begin{minipage}[t]{0.32\textwidth}
        \centering
        \subfigure[AIME24, 1.7B]{
            \includegraphics[width=0.96\textwidth]{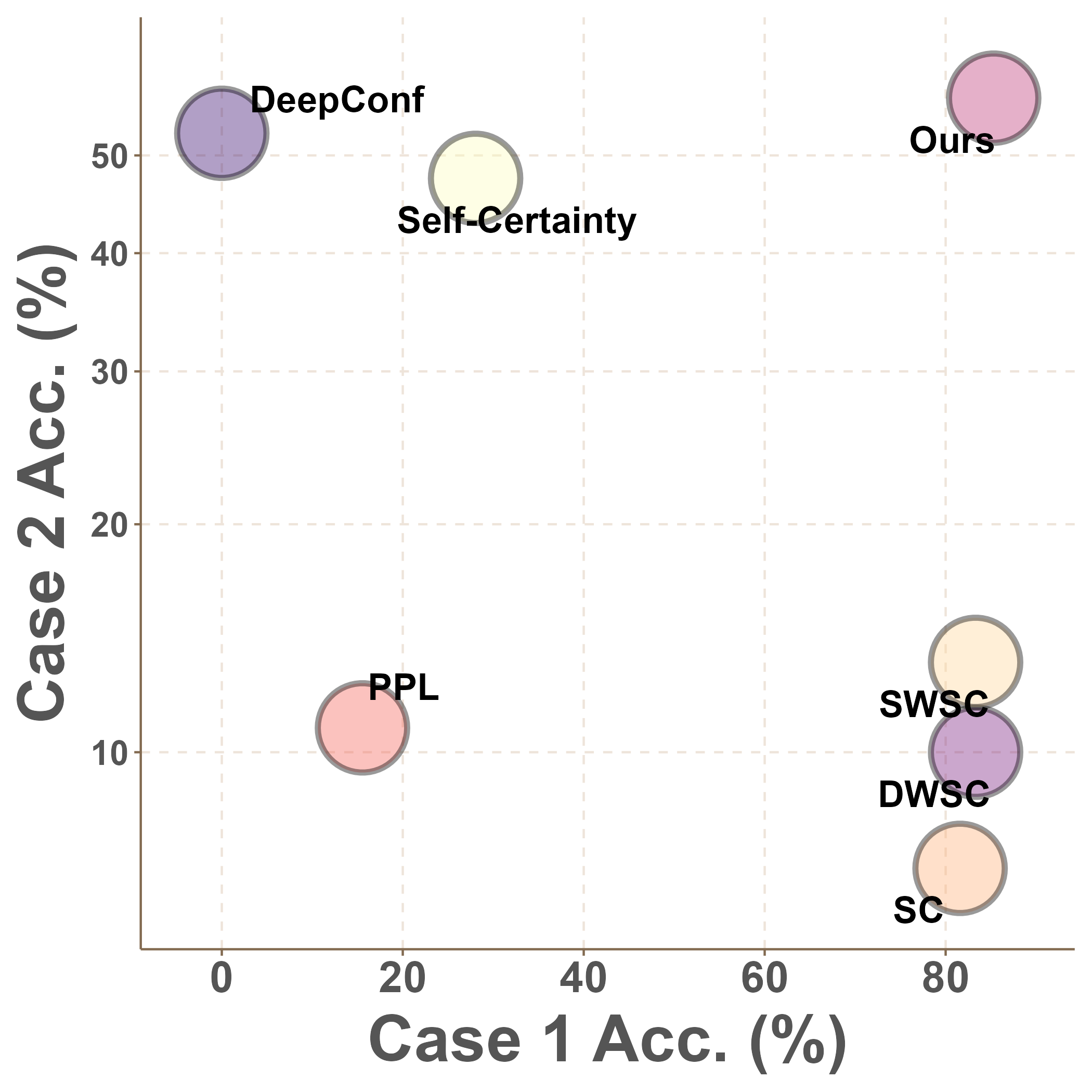}}
    \end{minipage}
    \hfill
    \begin{minipage}[t]{0.32\textwidth}
        \centering
        \subfigure[AIME24, 4B]{
            \includegraphics[width=0.96\textwidth]{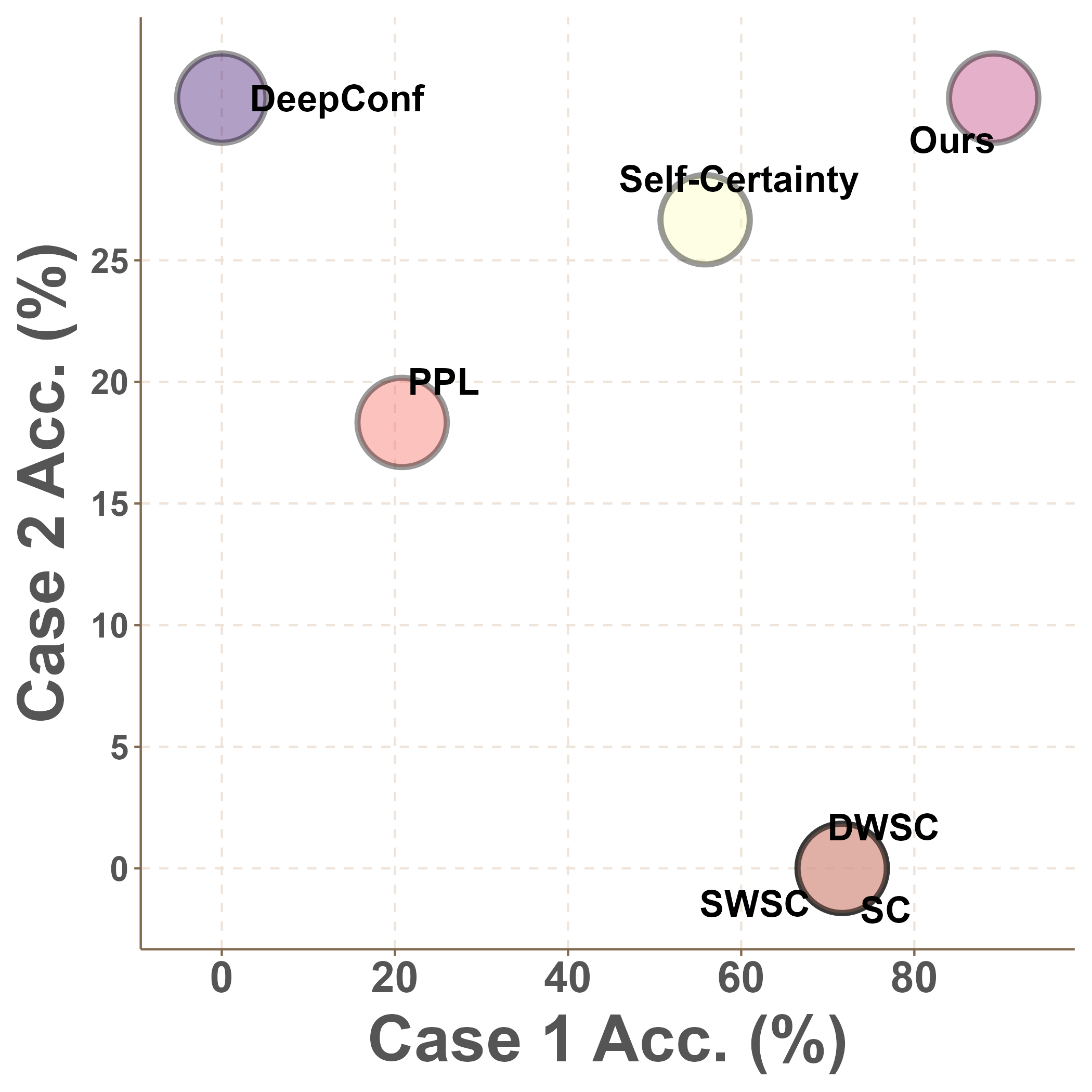}}
    \end{minipage}
    \hfill
    \begin{minipage}[t]{0.32\textwidth}
        \centering
        \subfigure[AIME24, 8B]{
            \includegraphics[width=0.96\textwidth]{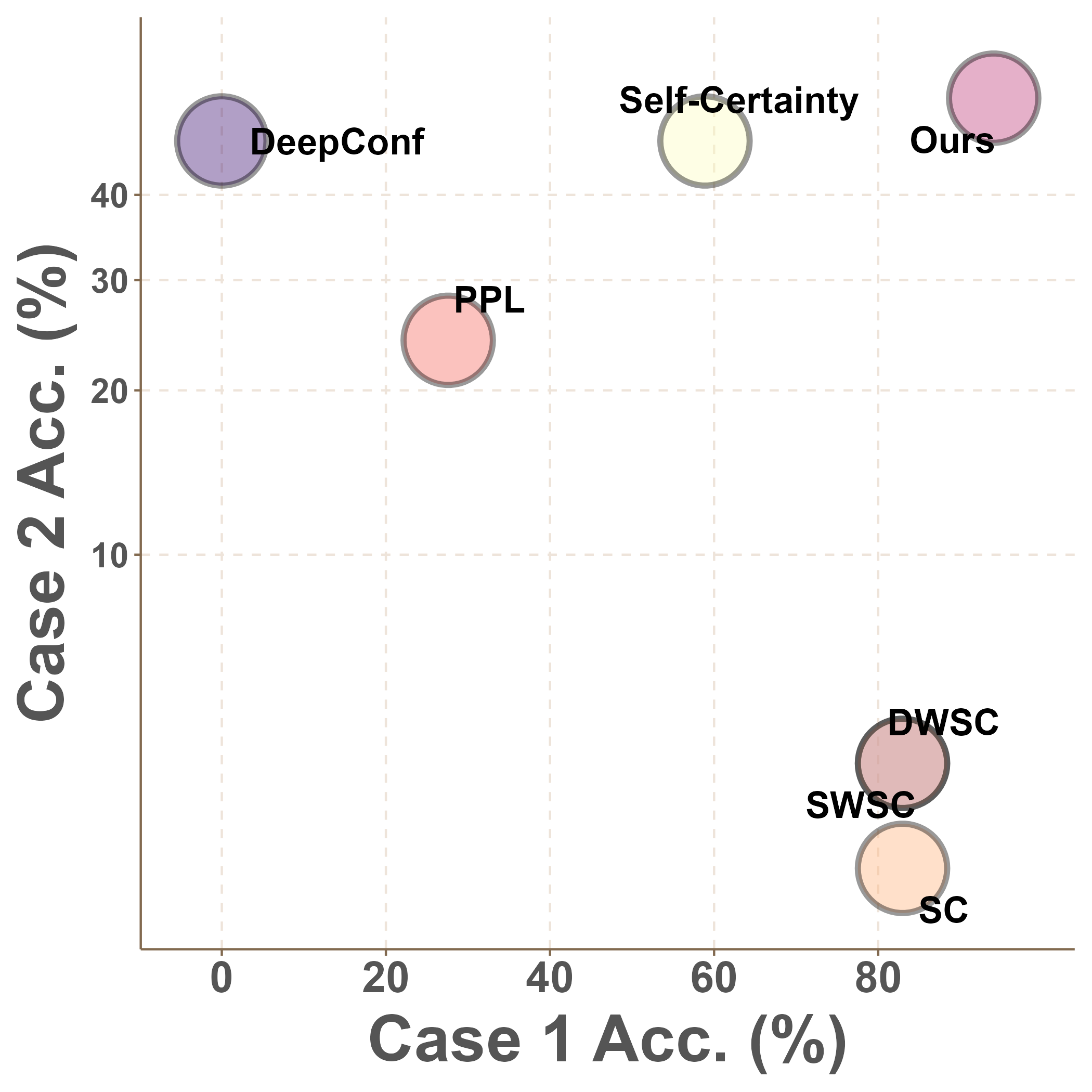}}
    \end{minipage}

    \medskip

    \begin{minipage}[t]{0.32\textwidth}
        \centering
        \subfigure[AIME25, 1.7B]{
            \includegraphics[width=0.96\textwidth]{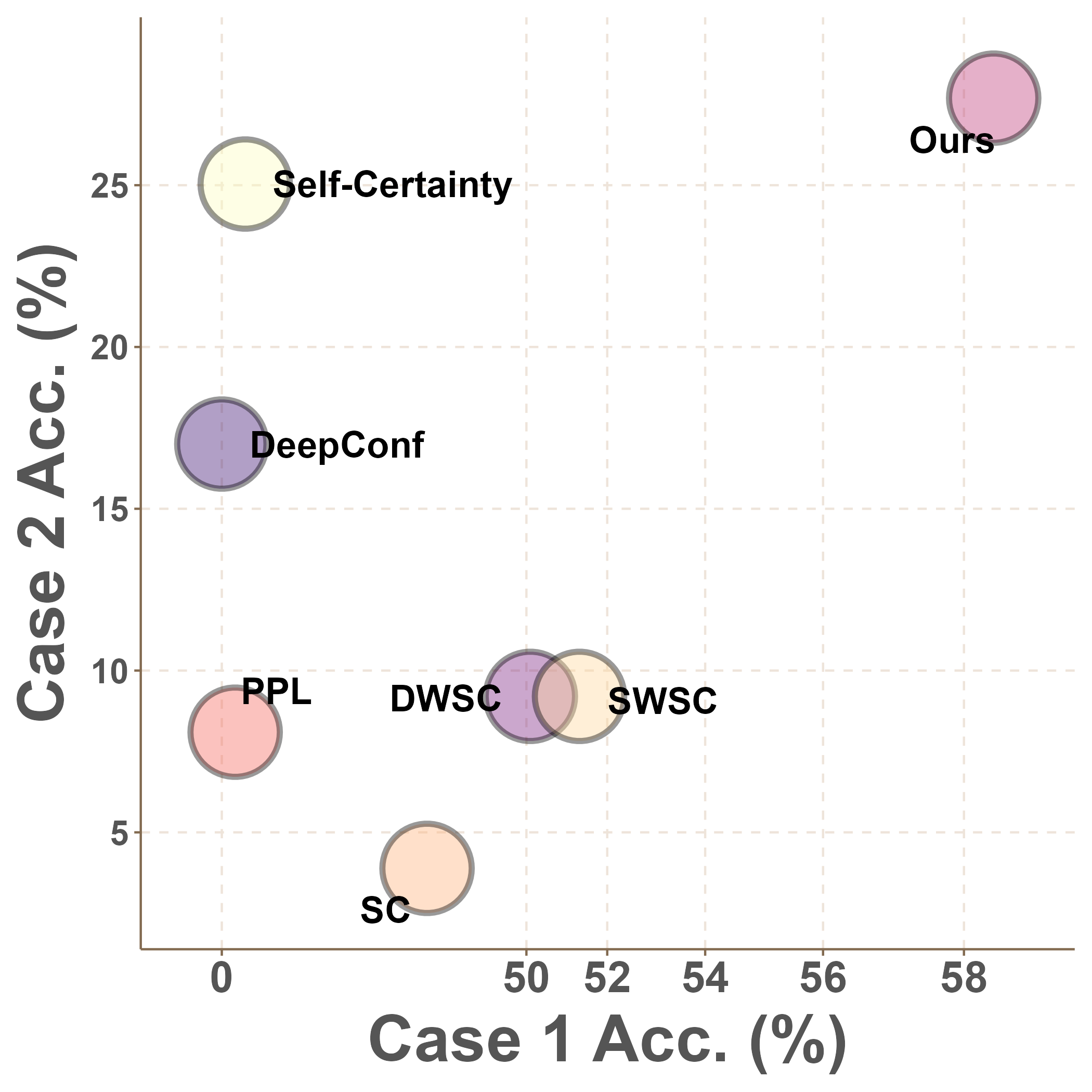}}
    \end{minipage}
    \hfill
    \begin{minipage}[t]{0.32\textwidth}
        \centering
        \subfigure[AIME25, 4B]{
            \includegraphics[width=0.96\textwidth]{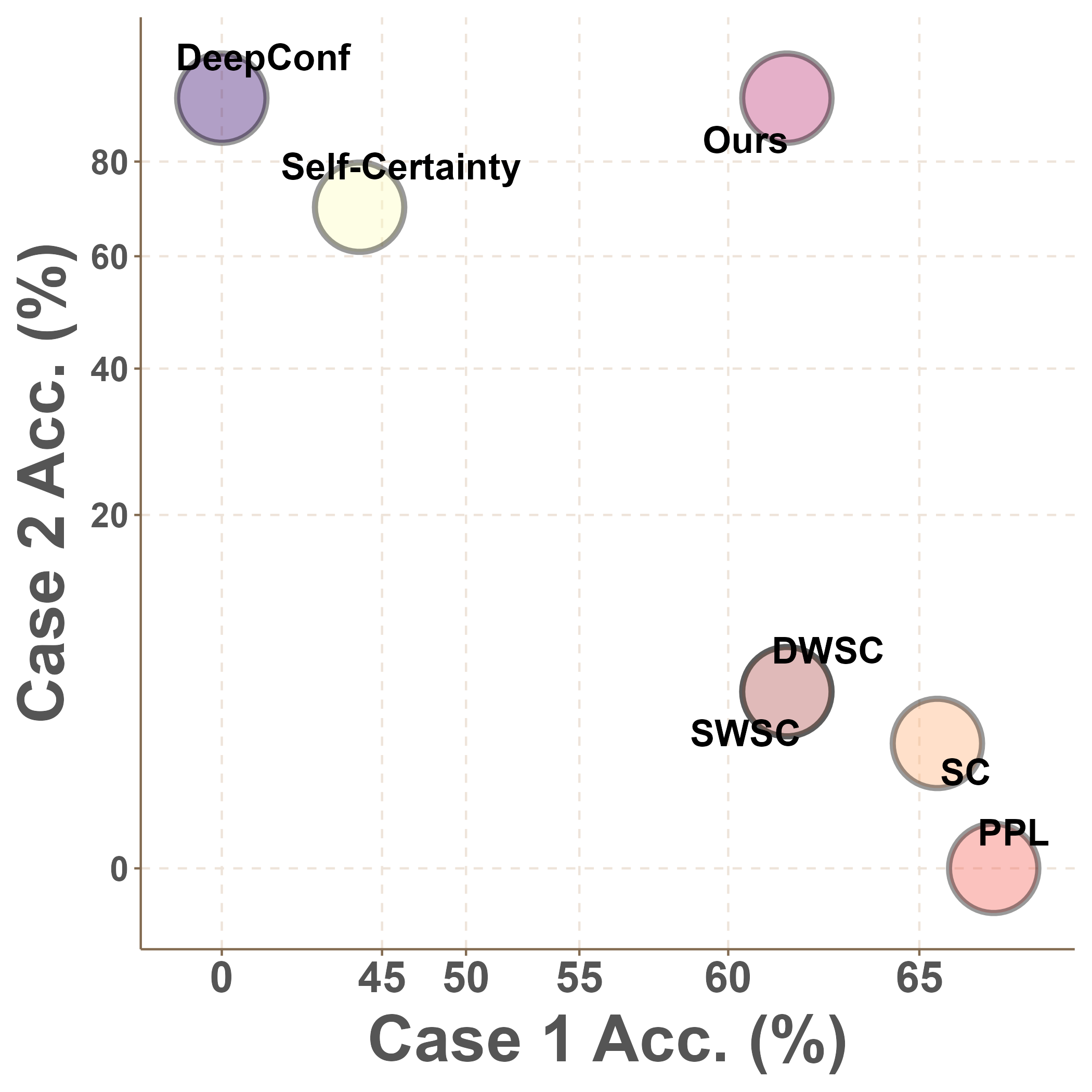}}
    \end{minipage}
    \hfill
    \begin{minipage}[t]{0.32\textwidth}
        \centering
        \subfigure[AIME25, 8B]{
            \includegraphics[width=0.96\textwidth]{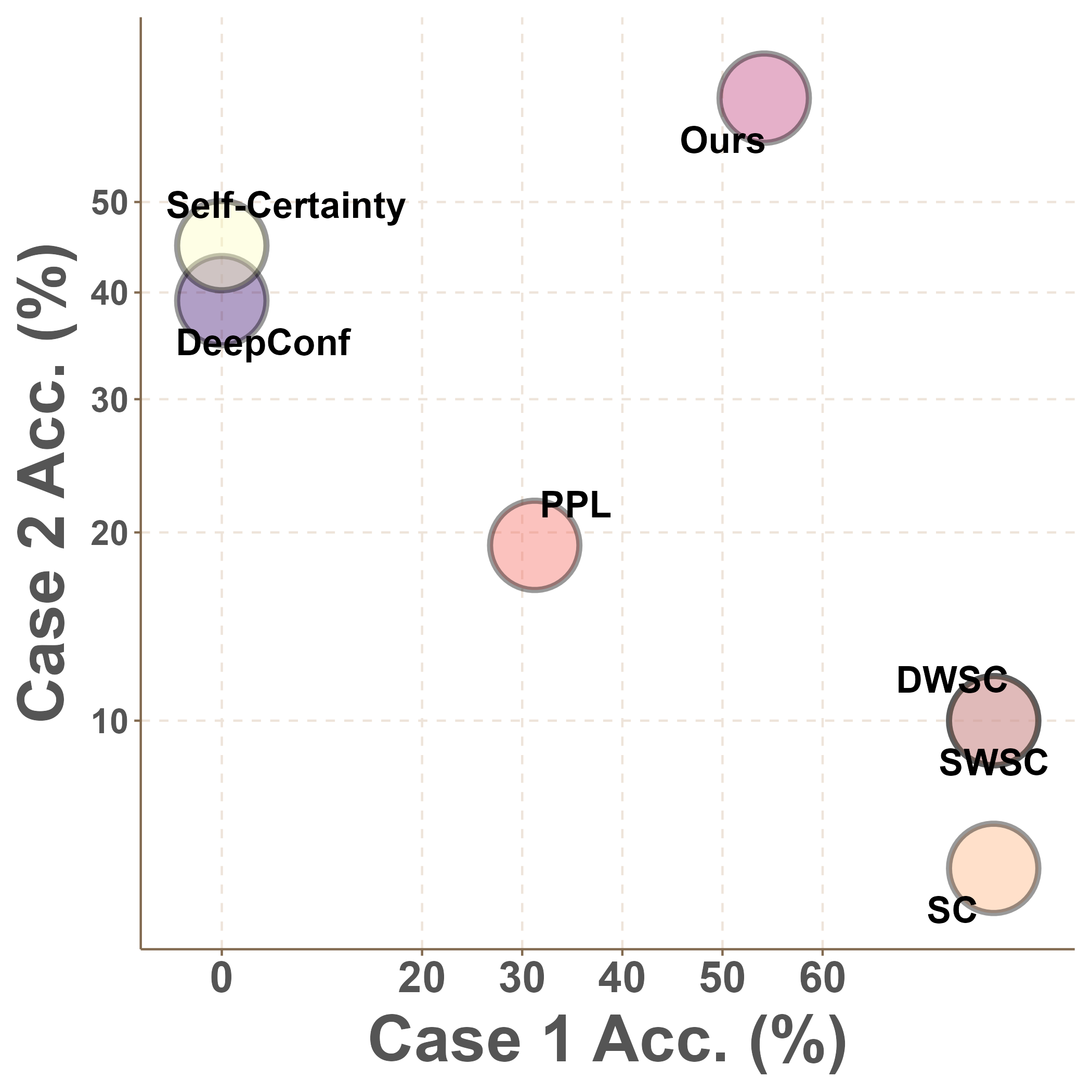}}
    \end{minipage}

    \medskip

    \begin{minipage}[t]{0.32\textwidth}
        \centering
        \subfigure[HMMT25, 1.7B]{
            \includegraphics[width=0.96\textwidth]{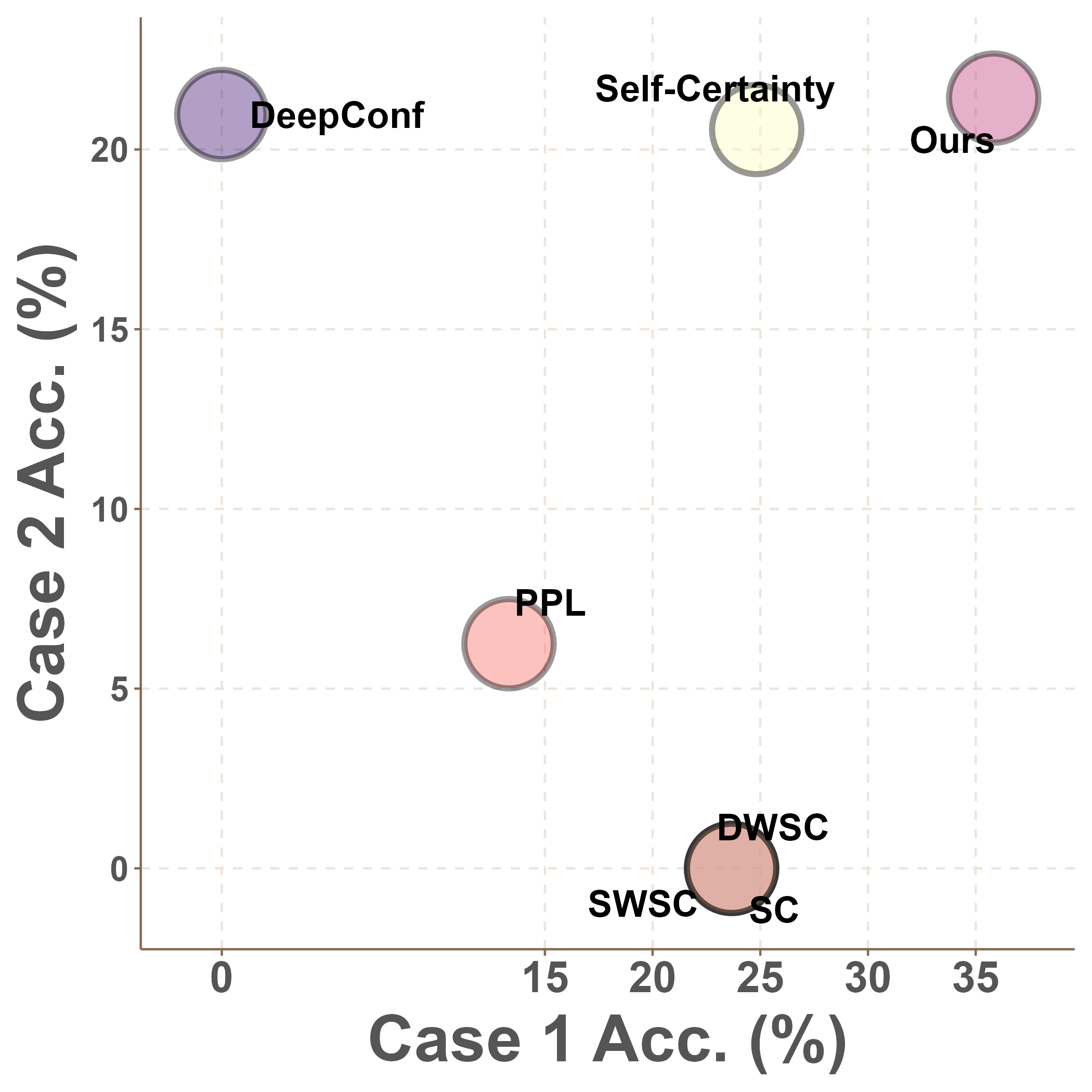}}
    \end{minipage}
    \hfill
    \begin{minipage}[t]{0.32\textwidth}
        \centering
        \subfigure[HMMT25, 4B]{
            \includegraphics[width=0.96\textwidth]{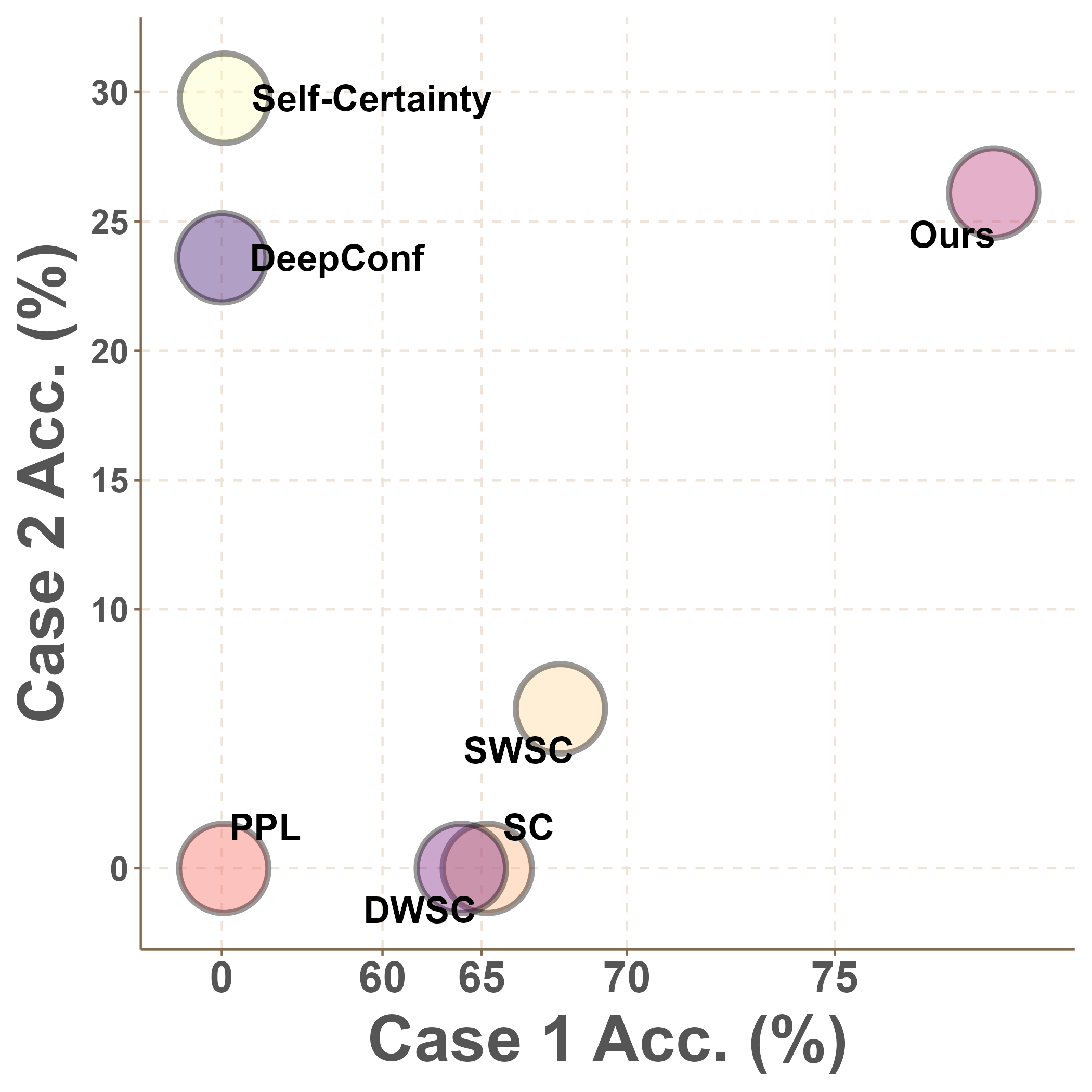}}
    \end{minipage}
    \hfill
    \begin{minipage}[t]{0.32\textwidth}
        \centering
        \subfigure[HMMT25, 8B]{
            \includegraphics[width=0.96\textwidth]{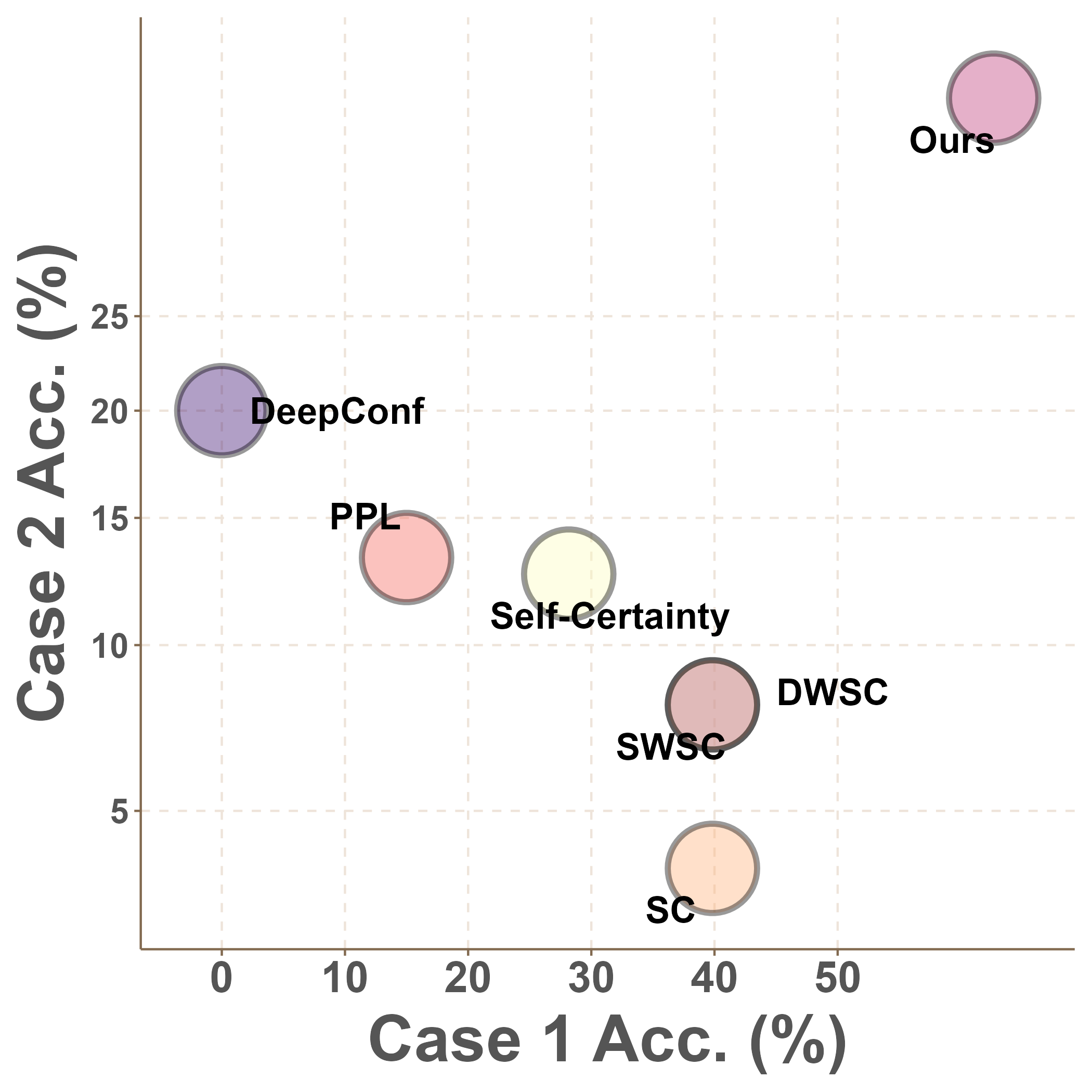}}
    \end{minipage}
    \caption{Tradeoff between Case I and Case II on AIME24, AIME25, and HMMT25 using Qwen3-1.7B, Qwen3-4B, and Qwen3-8B. Ours denotes $\mathsf{PGLA}$. DeepConf, PPL, and Self-Certainty denote DeepConf Best-of-$N$, PPL Best-of-$N$, and Self-Certainty Best-of-$N$, respectively. SWSC and DWSC denote Self-Certainty-Weighted SC and DeepConf-Weighted SC, respectively.}
    \label{fig:case_tradeoff_full}
\end{figure}

\Cref{fig:case_tradeoff_full} illustrates complementary failure modes: Best-of-$N$ often fails in Case I, while voting-based methods tend to fail in Case II. By combining voting support with pairwise confidence comparisons, $\mathsf{PGLA}$ generally achieves a better balance between the two cases.

\clearpage

\subsection{Effect of Ranking Quality by Dataset}
\label{app:controlled_rank_details}

We study how the ranking quality of path scores affects $\mathsf{PGLA}$.
Following \Cref{sec:problem_formulation}, let $a_i$ and $p_i$ be the answer
and confidence of path $i$. Here $p_i$ is the product-aggregated reward from
Qwen2.5-Math-PRM-7B, floored at $10^{-8}$. We keep the sampled paths,
answers, and PRM-Weighted SC support
$s_a^{\mathcal V}=\sum_{i=1}^{K}p_i\mathbf{1}(a_i=a)$ fixed, where
$\mathcal V$ denotes PRM-Weighted SC. Only the scores used for pairwise
comparisons are perturbed; the opponent weights and unit-strength
$\mathsf{PGLA}$ update remain unchanged.

\textbf{Controlled perturbation.}
The retention parameter $\lambda\in\{0,0.2,0.4,0.6,0.8,1\}$ controls
how many path-score assignments are preserved. For each query, define
\begin{equation}
    m_\lambda=\operatorname{round}((1-\lambda)K),
    \qquad
    \widetilde p_i^{(\lambda)}=p_{\sigma_\lambda(i)},
    \label{eq:controlled_rank_perturbation}
\end{equation}
where $\operatorname{round}$ denotes nearest-integer rounding and
$\sigma_\lambda$ is a random cyclic permutation on $m_\lambda$ uniformly
selected distinct path indices, extended by the identity on the remaining
indices. If $m_\lambda<2$, $\sigma_\lambda$ is the identity everywhere.
Thus, $\lambda=1$ preserves the original ranking, while $\lambda=0$
permutes all score assignments within the query. This procedure preserves
the score multiset and uses no correctness labels. Because of rounding and
score ties, $\lambda$ is a nominal retention fraction, rather than the
exact fraction of unchanged score values.

\textbf{Ranking quality.}
We measure ranking quality by the empirical area under the ROC curve (AUC)
for distinguishing correct from incorrect paths. Let $q$ index evaluated
queries, and let $\widetilde p_{q,i}^{(\lambda)}$ be the perturbed confidence
of path $i$ for query $q$. Define $\mathcal I_q^+$ and $\mathcal I_q^-$ as
the index sets of paths whose answers are correct and incorrect,
respectively. For a query containing both kinds of paths, its AUC is
\begin{equation}
    \widehat{\rho}_q(\lambda)=
    \frac{1}{|\mathcal I_q^+|\,|\mathcal I_q^-|}
    \sum_{i\in\mathcal I_q^+}\sum_{j\in\mathcal I_q^-}
    \left[
        \mathbf{1}\!\left(\widetilde p_{q,i}^{(\lambda)}>
                            \widetilde p_{q,j}^{(\lambda)}\right)
        +\frac{1}{2}\mathbf{1}\!\left(\widetilde p_{q,i}^{(\lambda)}=
                                      \widetilde p_{q,j}^{(\lambda)}\right)
    \right].
    \label{eq:controlled_rank_query_auc}
\end{equation}
This uses the same tie convention as \Cref{eq:pairwise_flow}: a correct
path outranking an incorrect path contributes one, and a tie contributes
one half. Let $\mathcal Q_{\mathrm{mix}}$ contain the queries for which
both $\mathcal I_q^+$ and $\mathcal I_q^-$ are nonempty. The dataset-level
AUC is their equally weighted average,
\begin{equation}
    \widehat{\rho}(\lambda)=
    \frac{1}{|\mathcal Q_{\mathrm{mix}}|}
    \sum_{q\in\mathcal Q_{\mathrm{mix}}}\widehat{\rho}_q(\lambda).
    \label{eq:controlled_rank_auc}
\end{equation}
Queries containing only correct or only incorrect paths are excluded from
this AUC but remain in the accuracy evaluation. The tables abbreviate
$\widehat{\rho}(\lambda)$ as $\widehat{\rho}$; the model-level curves in
\Cref{fig:rank_comparison} further average the dataset metrics equally.

The gains from $\mathsf{PGLA}$ generally grow as ranking quality improves
across datasets. In \Cref{fig:rank_comparison}, increasing
$\widehat{\rho}$ from $0.496$ to $0.729$ on InternLM2-Math-Plus-7B raises
the absolute accuracy gain from $-0.54\%$ to $+2.17\%$. A similar trend appears on
DeepSeek-Math-RL-7B in
\Cref{tab:controlled_rank_deepseek7b_math,tab:controlled_rank_deepseek7b_mathodyssey,tab:controlled_rank_deepseek7b_aime},
supporting the benefit of more reliable pairwise confidence comparisons.

\begin{table}[!htbp]
\centering
\begin{minipage}[t]{0.49\linewidth}
\centering
\caption{Controlled ranking on MATH with InternLM2-Math-Plus-7B ($K=64$).}
\label{tab:controlled_rank_internlm7b_math}
\footnotesize Base Acc.: $58.67_{\pm0.10}$\par\smallskip
\setlength{\tabcolsep}{3pt}
\renewcommand{\arraystretch}{1.12}
\begin{tabular*}{\linewidth}{@{\extracolsep{\fill}}cccc@{}}
\toprule
$\lambda$ & $\widehat{\rho}$ & $\mathsf{PGLA}$ Acc. & $\Delta$Acc \\
\midrule
$0.0$ & $0.496_{\pm0.003}$ & $58.23_{\pm0.09}$ & $-0.45_{\pm0.12}$ \\
$0.2$ & $0.557_{\pm0.002}$ & $58.69_{\pm0.16}$ & $+0.01_{\pm0.16}$ \\
$0.4$ & $0.617_{\pm0.002}$ & $59.23_{\pm0.20}$ & $+0.56_{\pm0.19}$ \\
$0.6$ & $0.671_{\pm0.002}$ & $59.61_{\pm0.07}$ & $+0.94_{\pm0.15}$ \\
$0.8$ & $0.732_{\pm0.002}$ & $60.05_{\pm0.20}$ & $+1.37_{\pm0.17}$ \\
$1.0$ & $0.791_{\pm0.002}$ & $\mathbf{60.40}_{\pm0.24}$ & $\mathbf{+1.73}_{\pm0.18}$ \\
\bottomrule
\end{tabular*}
\end{minipage}
\hfill
\begin{minipage}[t]{0.49\linewidth}
\centering
\caption{Controlled ranking on MATH with DeepSeek-Math-RL-7B ($K=64$).}
\label{tab:controlled_rank_deepseek7b_math}
\footnotesize Base Acc.: $62.04_{\pm0.20}$\par\smallskip
\setlength{\tabcolsep}{3pt}
\renewcommand{\arraystretch}{1.12}
\begin{tabular*}{\linewidth}{@{\extracolsep{\fill}}cccc@{}}
\toprule
$\lambda$ & $\widehat{\rho}$ & $\mathsf{PGLA}$ Acc. & $\Delta$Acc \\
\midrule
$0.0$ & $0.495_{\pm0.003}$ & $61.53_{\pm0.18}$ & $-0.51_{\pm0.23}$ \\
$0.2$ & $0.561_{\pm0.002}$ & $61.83_{\pm0.24}$ & $-0.22_{\pm0.13}$ \\
$0.4$ & $0.627_{\pm0.003}$ & $62.03_{\pm0.23}$ & $-0.01_{\pm0.07}$ \\
$0.6$ & $0.688_{\pm0.003}$ & $62.15_{\pm0.16}$ & $+0.11_{\pm0.20}$ \\
$0.8$ & $0.754_{\pm0.001}$ & $62.32_{\pm0.20}$ & $+0.28_{\pm0.23}$ \\
$1.0$ & $0.821_{\pm0.001}$ & $\mathbf{62.41}_{\pm0.20}$ & $\mathbf{+0.37}_{\pm0.19}$ \\
\bottomrule
\end{tabular*}
\end{minipage}
\par\vspace{1.5em}
\begin{minipage}[t]{0.49\linewidth}
\centering
\caption{Controlled ranking on MathOdyssey with InternLM2-Math-Plus-7B ($K=128$).}
\label{tab:controlled_rank_internlm7b_mathodyssey}
\footnotesize Base Acc.: $34.34_{\pm0.58}$\par\smallskip
\setlength{\tabcolsep}{3pt}
\renewcommand{\arraystretch}{1.12}
\begin{tabular*}{\linewidth}{@{\extracolsep{\fill}}cccc@{}}
\toprule
$\lambda$ & $\widehat{\rho}$ & $\mathsf{PGLA}$ Acc. & $\Delta$Acc \\
\midrule
$0.0$ & $0.499_{\pm0.008}$ & $33.60_{\pm0.85}$ & $-0.75_{\pm0.89}$ \\
$0.2$ & $0.538_{\pm0.008}$ & $35.35_{\pm0.55}$ & $+1.00_{\pm0.72}$ \\
$0.4$ & $0.577_{\pm0.010}$ & $36.25_{\pm0.46}$ & $+1.90_{\pm0.81}$ \\
$0.6$ & $0.615_{\pm0.006}$ & $37.56_{\pm0.60}$ & $+3.21_{\pm0.45}$ \\
$0.8$ & $0.652_{\pm0.009}$ & $38.17_{\pm0.49}$ & $+3.83_{\pm0.82}$ \\
$1.0$ & $0.692_{\pm0.008}$ & $\mathbf{38.79}_{\pm0.62}$ & $\mathbf{+4.45}_{\pm0.78}$ \\
\bottomrule
\end{tabular*}
\end{minipage}
\hfill
\begin{minipage}[t]{0.49\linewidth}
\centering
\caption{Controlled ranking on MathOdyssey with DeepSeek-Math-RL-7B ($K=128$).}
\label{tab:controlled_rank_deepseek7b_mathodyssey}
\footnotesize Base Acc.: $40.28_{\pm0.53}$\par\smallskip
\setlength{\tabcolsep}{3pt}
\renewcommand{\arraystretch}{1.12}
\begin{tabular*}{\linewidth}{@{\extracolsep{\fill}}cccc@{}}
\toprule
$\lambda$ & $\widehat{\rho}$ & $\mathsf{PGLA}$ Acc. & $\Delta$Acc \\
\midrule
$0.0$ & $0.497_{\pm0.009}$ & $40.00_{\pm0.53}$ & $-0.28_{\pm0.52}$ \\
$0.2$ & $0.539_{\pm0.010}$ & $40.41_{\pm0.47}$ & $+0.13_{\pm0.35}$ \\
$0.4$ & $0.564_{\pm0.008}$ & $40.33_{\pm0.33}$ & $+0.05_{\pm0.54}$ \\
$0.6$ & $0.603_{\pm0.007}$ & $40.44_{\pm0.75}$ & $+0.15_{\pm0.88}$ \\
$0.8$ & $0.634_{\pm0.006}$ & $40.57_{\pm0.69}$ & $+0.28_{\pm0.62}$ \\
$1.0$ & $0.670_{\pm0.005}$ & $\mathbf{40.69}_{\pm0.53}$ & $\mathbf{+0.41}_{\pm0.52}$ \\
\bottomrule
\end{tabular*}
\end{minipage}
\par\vspace{1.5em}
\begin{minipage}[t]{0.49\linewidth}
\centering
\caption{Controlled ranking on AIME with InternLM2-Math-Plus-7B ($K=128$).}
\label{tab:controlled_rank_internlm7b_aime}
\footnotesize Base Acc.: $16.23_{\pm0.48}$\par\smallskip
\setlength{\tabcolsep}{3pt}
\renewcommand{\arraystretch}{1.12}
\begin{tabular*}{\linewidth}{@{\extracolsep{\fill}}cccc@{}}
\toprule
$\lambda$ & $\widehat{\rho}$ & $\mathsf{PGLA}$ Acc. & $\Delta$Acc \\
\midrule
$0.0$ & $0.493_{\pm0.011}$ & $15.81_{\pm0.32}$ & $-0.42_{\pm0.32}$ \\
$0.2$ & $0.542_{\pm0.007}$ & $16.08_{\pm0.42}$ & $-0.15_{\pm0.38}$ \\
$0.4$ & $0.579_{\pm0.011}$ & $16.14_{\pm0.53}$ & $-0.09_{\pm0.44}$ \\
$0.6$ & $0.623_{\pm0.010}$ & $16.36_{\pm0.55}$ & $+0.13_{\pm0.51}$ \\
$0.8$ & $0.659_{\pm0.010}$ & $16.50_{\pm0.59}$ & $+0.27_{\pm0.38}$ \\
$1.0$ & $0.703_{\pm0.008}$ & $\mathbf{16.57}_{\pm0.35}$ & $\mathbf{+0.34}_{\pm0.35}$ \\
\bottomrule
\end{tabular*}
\end{minipage}
\hfill
\begin{minipage}[t]{0.49\linewidth}
\centering
\caption{Controlled ranking on AIME with DeepSeek-Math-RL-7B ($K=128$).}
\label{tab:controlled_rank_deepseek7b_aime}
\footnotesize Base Acc.: $16.93_{\pm0.43}$\par\smallskip
\setlength{\tabcolsep}{3pt}
\renewcommand{\arraystretch}{1.12}
\begin{tabular*}{\linewidth}{@{\extracolsep{\fill}}cccc@{}}
\toprule
$\lambda$ & $\widehat{\rho}$ & $\mathsf{PGLA}$ Acc. & $\Delta$Acc \\
\midrule
$0.0$ & $0.497_{\pm0.008}$ & $16.53_{\pm0.42}$ & $-0.41_{\pm0.37}$ \\
$0.2$ & $0.547_{\pm0.013}$ & $16.68_{\pm0.34}$ & $-0.26_{\pm0.41}$ \\
$0.4$ & $0.590_{\pm0.008}$ & $16.78_{\pm0.57}$ & $-0.15_{\pm0.38}$ \\
$0.6$ & $0.631_{\pm0.007}$ & $16.87_{\pm0.60}$ & $-0.06_{\pm0.52}$ \\
$0.8$ & $0.675_{\pm0.010}$ & $17.18_{\pm0.61}$ & $+0.25_{\pm0.39}$ \\
$1.0$ & $0.720_{\pm0.007}$ & $\mathbf{17.49}_{\pm0.57}$ & $\mathbf{+0.56}_{\pm0.42}$ \\
\bottomrule
\end{tabular*}
\end{minipage}
\end{table}

\subsection{Temperature Sensitivity}

To further examine robustness under more diverse sampled reasoning paths, we conduct additional experiments on MathOdyssey and AIME using \textbf{InternLM2-Math-Plus-1.8B}, \textbf{InternLM2-Math-Plus-7B}, and \textbf{DeepSeek-Math-RL-7B}. \Cref{tab:temperature_sensitivity} compares the methods under higher decoding temperatures with $T=1.1$ and $T=1.3$. Our method remains competitive and generally outperforms the comparison methods, indicating that the proposed aggregation mechanism is still effective when the sampled proposal pool becomes more diverse.

\begin{table*}[p]
\centering
\scriptsize
\caption{Temperature sensitivity under the same internal-confidence protocol as \Cref{tab:main_results}. We vary only the decoding temperature; \textbf{+ Ours} denotes the $\mathsf{PGLA}$-enhanced variant of each voting baseline.}
\label{tab:temperature_sensitivity}
\renewcommand{\arraystretch}{1.38}
\setlength{\tabcolsep}{3.2pt}
\resizebox{\textwidth}{!}{%
\begin{tabular}{@{}lcccccccc@{}}
\toprule
\multirow{3}{*}{\textbf{Method}} & \multicolumn{4}{c}{\textbf{MathOdyssey}} & \multicolumn{4}{c}{\textbf{AIME}} \\
\cmidrule(lr){2-5}\cmidrule(lr){6-9}
 & \multicolumn{2}{c}{$T=1.1$} & \multicolumn{2}{c}{$T=1.3$} & \multicolumn{2}{c}{$T=1.1$} & \multicolumn{2}{c}{$T=1.3$} \\
\cmidrule(lr){2-3}\cmidrule(lr){4-5}\cmidrule(lr){6-7}\cmidrule(lr){8-9}
 & Acc. $\uparrow$ & ECE $\downarrow$ & Acc. $\uparrow$ & ECE $\downarrow$ & Acc. $\uparrow$ & ECE $\downarrow$ & Acc. $\uparrow$ & ECE $\downarrow$ \\
\midrule
\rowcolor{gray!20} \multicolumn{9}{c}{\textbf{InternLM2-Math-Plus-1.8B}} \\
\midrule
PPL & $ 16.43_{\pm 0.91} $ & $ 75.99_{\pm 0.89} $ & $ 16.38_{\pm 1.42} $ & $ 71.72_{\pm 1.43} $ & $ 1.97_{\pm 0.29} $ & $ 90.29_{\pm 0.28} $ & $ 2.08_{\pm 0.27} $ & $ 84.85_{\pm 0.27} $ \\
\addlinespace[0.8pt]
SC & $ 13.75_{\pm 0.51} $ & $ 18.02_{\pm 0.57} $ & $ 14.95_{\pm 0.87} $ & $ 5.15_{\pm 1.12} $ & $ 2.12_{\pm 0.10} $ & $ 16.52_{\pm 0.13} $ & $ 1.26_{\pm 0.08} $ & $ 8.80_{\pm 0.12} $ \\
\rowcolor{gray!10} \textbf{SC + Ours} & $ \mathbf{18.59_{\pm 0.78}} $ & $ \mathbf{12.75_{\pm 0.75}} $ & $ \mathbf{19.23_{\pm 0.76}} $ & $ \mathbf{6.86_{\pm 1.04}} $ & $ \mathbf{3.11_{\pm 0.17}} $ & $ \mathbf{14.78_{\pm 0.19}} $ & $ \mathbf{1.82_{\pm 0.17}} $ & $ \mathbf{10.64_{\pm 0.16}} $ \\
\addlinespace[0.8pt]
Weighted SC & $ 13.91_{\pm 0.54} $ & $ 17.76_{\pm 0.62} $ & $ 15.68_{\pm 0.79} $ & $ 9.41_{\pm 0.86} $ & $ 2.14_{\pm 0.11} $ & $ 16.28_{\pm 0.16} $ & $ 1.58_{\pm 0.07} $ & $ 12.65_{\pm 0.12} $ \\
\rowcolor{gray!10} \textbf{Weighted SC + Ours} & $ \mathbf{18.84_{\pm 0.76}} $ & $ \mathbf{12.25_{\pm 0.79}} $ & $ \mathbf{21.31_{\pm 1.15}} $ & $ \mathbf{6.95_{\pm 1.10}} $ & $ \mathbf{3.29_{\pm 0.24}} $ & $ \mathbf{14.40_{\pm 0.29}} $ & $ \mathbf{2.22_{\pm 0.17}} $ & $ \mathbf{13.80_{\pm 0.26}} $ \\
\addlinespace[0.8pt]
Adaptive SC & $ 13.81_{\pm 0.63} $ & $ 22.30_{\pm 0.96} $ & $ 15.05_{\pm 0.77} $ & $ 6.56_{\pm 0.85} $ & $ 2.12_{\pm 0.10} $ & $ 18.93_{\pm 0.27} $ & $ 1.26_{\pm 0.08} $ & $ 9.65_{\pm 0.12} $ \\
\rowcolor{gray!10} \textbf{Adaptive SC + Ours} & $ \mathbf{17.60_{\pm 1.01}} $ & $ \mathbf{17.05_{\pm 1.31}} $ & $ \mathbf{18.46_{\pm 0.90}} $ & $ \mathbf{8.02_{\pm 1.09}} $ & $ \mathbf{3.10_{\pm 0.19}} $ & $ \mathbf{16.67_{\pm 0.43}} $ & $ \mathbf{1.81_{\pm 0.18}} $ & $ \mathbf{11.62_{\pm 0.26}} $ \\
\addlinespace[0.8pt]
RPC & $ 15.53_{\pm 0.54} $ & $ 16.10_{\pm 0.82} $ & $ 16.07_{\pm 0.94} $ & $ 11.41_{\pm 0.95} $ & $ 2.80_{\pm 0.18} $ & $ 15.64_{\pm 0.15} $ & $ 1.64_{\pm 0.14} $ & $ 14.42_{\pm 0.21} $ \\
\rowcolor{gray!10} \textbf{RPC + Ours} & $ \mathbf{20.41_{\pm 0.69}} $ & $ \mathbf{10.45_{\pm 0.71}} $ & $ \mathbf{21.83_{\pm 0.96}} $ & $ \mathbf{7.69_{\pm 1.24}} $ & $ \mathbf{3.63_{\pm 0.25}} $ & $ \mathbf{13.78_{\pm 0.26}} $ & $ \mathbf{2.33_{\pm 0.17}} $ & $ \mathbf{14.83_{\pm 0.28}} $ \\
\midrule
\rowcolor{gray!20} \multicolumn{9}{c}{\textbf{InternLM2-Math-Plus-7B}} \\
\midrule
PPL & $ 28.61_{\pm 1.33} $ & $ 65.31_{\pm 1.33} $ & $ 26.38_{\pm 1.41} $ & $ 65.61_{\pm 1.41} $ & $ 6.02_{\pm 0.34} $ & $ 87.67_{\pm 0.35} $ & $ 6.50_{\pm 0.41} $ & $ 84.66_{\pm 0.40} $ \\
\addlinespace[0.8pt]
SC & $ 27.89_{\pm 0.43} $ & $ 11.31_{\pm 0.52} $ & $ 27.51_{\pm 0.69} $ & $ 9.38_{\pm 1.14} $ & $ 8.73_{\pm 0.24} $ & $ 13.71_{\pm 0.25} $ & $ 8.02_{\pm 0.20} $ & $ 12.02_{\pm 0.24} $ \\
\rowcolor{gray!10} \textbf{SC + Ours} & $ \mathbf{35.84_{\pm 0.84}} $ & $ \mathbf{9.07_{\pm 0.90}} $ & $ \mathbf{36.35_{\pm 0.50}} $ & $ \mathbf{8.82_{\pm 0.75}} $ & $ \mathbf{9.61_{\pm 0.15}} $ & $ \mathbf{13.08_{\pm 0.15}} $ & $ \mathbf{9.09_{\pm 0.26}} $ & $ \mathbf{11.23_{\pm 0.31}} $ \\
\addlinespace[0.8pt]
Weighted SC & $ 28.38_{\pm 0.33} $ & $ 10.77_{\pm 0.45} $ & $ 28.71_{\pm 0.60} $ & $ 8.11_{\pm 0.85} $ & $ 8.89_{\pm 0.22} $ & $ 13.53_{\pm 0.23} $ & $ 8.40_{\pm 0.25} $ & $ 11.52_{\pm 0.28} $ \\
\rowcolor{gray!10} \textbf{Weighted SC + Ours} & $ \mathbf{36.07_{\pm 0.86}} $ & $ \mathbf{9.20_{\pm 0.84}} $ & $ \mathbf{36.79_{\pm 0.53}} $ & $ \mathbf{8.85_{\pm 0.94}} $ & $ \mathbf{9.75_{\pm 0.23}} $ & $ \mathbf{12.83_{\pm 0.24}} $ & $ \mathbf{9.32_{\pm 0.28}} $ & $ \mathbf{10.90_{\pm 0.29}} $ \\
\addlinespace[0.8pt]
Adaptive SC & $ 27.92_{\pm 0.61} $ & $ 16.96_{\pm 0.77} $ & $ 27.75_{\pm 0.65} $ & $ 13.27_{\pm 0.80} $ & $ 8.71_{\pm 0.25} $ & $ 16.61_{\pm 0.46} $ & $ 8.00_{\pm 0.16} $ & $ 14.51_{\pm 0.26} $ \\
\rowcolor{gray!10} \textbf{Adaptive SC + Ours} & $ \mathbf{34.00_{\pm 0.99}} $ & $ \mathbf{12.35_{\pm 0.82}} $ & $ \mathbf{34.24_{\pm 0.83}} $ & $ \mathbf{10.02_{\pm 0.98}} $ & $ \mathbf{9.43_{\pm 0.32}} $ & $ \mathbf{15.62_{\pm 0.48}} $ & $ \mathbf{9.11_{\pm 0.21}} $ & $ \mathbf{13.11_{\pm 0.41}} $ \\
\addlinespace[0.8pt]
RPC & $ 32.93_{\pm 0.64} $ & $ 8.05_{\pm 0.92} $ & $ 33.01_{\pm 0.50} $ & $ 5.97_{\pm 0.96} $ & $ 9.31_{\pm 0.28} $ & $ 13.52_{\pm 0.25} $ & $ 8.83_{\pm 0.23} $ & $ 11.40_{\pm 0.28} $ \\
\rowcolor{gray!10} \textbf{RPC + Ours} & $ \mathbf{36.99_{\pm 1.07}} $ & $ \mathbf{8.38_{\pm 1.26}} $ & $ \mathbf{38.02_{\pm 0.68}} $ & $ \mathbf{7.97_{\pm 0.91}} $ & $ \mathbf{10.13_{\pm 0.38}} $ & $ \mathbf{12.17_{\pm 0.40}} $ & $ \mathbf{9.30_{\pm 0.22}} $ & $ \mathbf{10.88_{\pm 0.26}} $ \\
\midrule
\rowcolor{gray!20} \multicolumn{9}{c}{\textbf{Deepseek-Math-RL-7B}} \\
\midrule
PPL & $ 24.78_{\pm 1.15} $ & $ 70.85_{\pm 1.18} $ & $ 25.76_{\pm 1.26} $ & $ 68.39_{\pm 1.29} $ & $ 3.94_{\pm 0.37} $ & $ 92.17_{\pm 0.37} $ & $ 4.39_{\pm 0.33} $ & $ 90.33_{\pm 0.34} $ \\
\addlinespace[0.8pt]
SC & $ 36.76_{\pm 0.39} $ & $ 9.46_{\pm 0.81} $ & $ 37.75_{\pm 0.25} $ & $ 8.78_{\pm 1.16} $ & $ 8.95_{\pm 0.26} $ & $ 11.95_{\pm 0.34} $ & $ 9.39_{\pm 0.24} $ & $ 10.23_{\pm 0.26} $ \\
\rowcolor{gray!10} \textbf{SC + Ours} & $ \mathbf{37.22_{\pm 0.55}} $ & $ \mathbf{11.03_{\pm 1.11}} $ & $ \mathbf{38.55_{\pm 0.61}} $ & $ \mathbf{9.01_{\pm 0.70}} $ & $ \mathbf{9.39_{\pm 0.30}} $ & $ \mathbf{11.01_{\pm 0.34}} $ & $ \mathbf{10.15_{\pm 0.33}} $ & $ \mathbf{9.57_{\pm 0.36}} $ \\
\addlinespace[0.8pt]
Weighted SC & $ 36.84_{\pm 0.45} $ & $ 9.40_{\pm 0.81} $ & $ 37.76_{\pm 0.39} $ & $ 8.91_{\pm 0.86} $ & $ 9.04_{\pm 0.29} $ & $ 11.86_{\pm 0.38} $ & $ 9.42_{\pm 0.24} $ & $ 10.24_{\pm 0.24} $ \\
\rowcolor{gray!10} \textbf{Weighted SC + Ours} & $ \mathbf{37.17_{\pm 0.60}} $ & $ \mathbf{11.22_{\pm 1.23}} $ & $ \mathbf{38.56_{\pm 0.63}} $ & $ \mathbf{9.14_{\pm 0.88}} $ & $ \mathbf{9.41_{\pm 0.33}} $ & $ \mathbf{10.93_{\pm 0.37}} $ & $ \mathbf{10.19_{\pm 0.31}} $ & $ \mathbf{9.46_{\pm 0.34}} $ \\
\addlinespace[0.8pt]
Adaptive SC & $ 36.62_{\pm 0.35} $ & $ 12.71_{\pm 1.30} $ & $ 37.63_{\pm 0.42} $ & $ 11.23_{\pm 1.22} $ & $ 8.93_{\pm 0.29} $ & $ 14.23_{\pm 0.42} $ & $ 9.40_{\pm 0.24} $ & $ 12.72_{\pm 0.36} $ \\
\rowcolor{gray!10} \textbf{Adaptive SC + Ours} & $ \mathbf{35.94_{\pm 0.38}} $ & $ \mathbf{12.33_{\pm 0.84}} $ & $ \mathbf{37.03_{\pm 0.96}} $ & $ \mathbf{11.30_{\pm 1.89}} $ & $ \mathbf{9.26_{\pm 0.36}} $ & $ \mathbf{13.09_{\pm 0.55}} $ & $ \mathbf{9.89_{\pm 0.34}} $ & $ \mathbf{11.95_{\pm 0.33}} $ \\
\addlinespace[0.8pt]
RPC & $ 36.92_{\pm 0.40} $ & $ 10.19_{\pm 0.60} $ & $ 38.25_{\pm 0.47} $ & $ 8.96_{\pm 0.81} $ & $ 9.09_{\pm 0.34} $ & $ 11.81_{\pm 0.40} $ & $ 10.04_{\pm 0.39} $ & $ 10.35_{\pm 0.44} $ \\
\rowcolor{gray!10} \textbf{RPC + Ours} & $ \mathbf{37.22_{\pm 0.43}} $ & $ \mathbf{11.63_{\pm 1.45}} $ & $ \mathbf{38.95_{\pm 0.82}} $ & $ \mathbf{8.82_{\pm 1.06}} $ & $ \mathbf{9.35_{\pm 0.29}} $ & $ \mathbf{10.47_{\pm 0.31}} $ & $ \mathbf{10.32_{\pm 0.26}} $ & $ \mathbf{8.82_{\pm 0.30}} $ \\
\bottomrule
\end{tabular}
}
\end{table*}


\subsection{Efficiency Comparison}

\Cref{tab:efficiency_internlm18b_app,tab:efficiency_deepseek_app} compare the sampling budgets of SC, SC + Ours, RPC, and RPC + Ours on MATH, MathOdyssey, and AIME, capped at 64 paths for MATH and 128 otherwise. For each model--dataset setting, the target is the best mean accuracy attained by SC on the evaluated grid. We then find the smallest grid budget at which each other method matches or exceeds that target. The reported savings describe empirical accuracy--budget tradeoffs. 

\Cref{tab:efficiency_internlm18b_app,tab:efficiency_deepseek_app} show that $\mathsf{PGLA}$-enhanced SC and RPC generally reach SC's best observed accuracy with fewer sampled paths. The savings are larger on InternLM2-Math-Plus-1.8B, while DeepSeek-Math-RL-7B also benefits on MathOdyssey and AIME. These results suggest that pairwise evidence improves the tradeoff between accuracy and sampling budget.

The computational cost of the additional aggregation step is analyzed
separately in \Cref{app:computational_overhead}.

\begin{table*}[t]
\centering
\small
\caption{Post-hoc sampling-budget comparison for \textbf{InternLM2-Math-Plus-1.8B}. Best of SC gives its
highest grid mean and smallest attaining $K$. Alternatives use the minimum $K$ matching or
exceeding this unrounded target.  $\Delta$ gives
absolute accuracy change (\%) and sample-count change (\%) relative to SC.}
\label{tab:efficiency_internlm18b_app}
\setlength{\tabcolsep}{4pt}
\begin{tabular*}{\textwidth}{@{\extracolsep{\fill}}lcccccc@{}}
\toprule
\textbf{Method} & \multicolumn{2}{c}{\textbf{MATH}} & \multicolumn{2}{c}{\textbf{MathOdyssey}} & \multicolumn{2}{c}{\textbf{AIME}} \\
\cmidrule(lr){2-3}\cmidrule(lr){4-5}\cmidrule(lr){6-7}
 & Acc. & $K$ & Acc. & $K$ & Acc. & $K$ \\
\midrule
Best of SC & 36.48 $\pm$ 0.14 & 64 & 14.91 $\pm$ 0.95 & 48 & 2.67 $\pm$ 0.21 & 112 \\
\midrule
SC + Ours & 37.76 $\pm$ 0.34 & 16 & 16.53 $\pm$ 0.92 & 8 & 2.76 $\pm$ 0.34 & 32 \\
$\Delta$ vs.\ SC & +1.27 & \textbf{-75.0\%} & +1.62 & \textbf{-83.3\%} & +0.09 & \textbf{-71.4\%} \\
\midrule
RPC & 37.09 $\pm$ 0.20 & 32 & 15.37 $\pm$ 1.45 & 8 & 2.71 $\pm$ 0.39 & 32 \\
$\Delta$ vs.\ SC & +0.61 & \textbf{-50.0\%} & +0.46 & \textbf{-83.3\%} & +0.04 & \textbf{-71.4\%} \\
\midrule
RPC + Ours & 37.67 $\pm$ 0.36 & 16 & 16.20 $\pm$ 0.82 & 8 & 2.86 $\pm$ 0.36 & 40 \\
$\Delta$ vs.\ SC & +1.18 & \textbf{-75.0\%} & +1.28 & \textbf{-83.3\%} & +0.19 & \textbf{-64.3\%} \\
\bottomrule
\end{tabular*}
\end{table*}

\begin{table*}[t]
\centering
\small
\caption{Post-hoc sampling-budget comparison for \textbf{DeepSeek-Math-RL-7B}. Best of SC gives its
highest grid mean and smallest attaining $K$. Alternatives use the minimum $K$ matching or
exceeding this unrounded target.  $\Delta$ gives
absolute accuracy change (\%) and sample-count change (\%) relative to SC.}
\label{tab:efficiency_deepseek_app}
\setlength{\tabcolsep}{4pt}
\begin{tabular*}{\textwidth}{@{\extracolsep{\fill}}lcccccc@{}}
\toprule
\textbf{Method} & \multicolumn{2}{c}{\textbf{MATH}} & \multicolumn{2}{c}{\textbf{MathOdyssey}} & \multicolumn{2}{c}{\textbf{AIME}} \\
\cmidrule(lr){2-3}\cmidrule(lr){4-5}\cmidrule(lr){6-7}
 & Acc. & $K$ & Acc. & $K$ & Acc. & $K$ \\
\midrule
Best of SC & 53.33 $\pm$ 0.08 & 64 & 36.57 $\pm$ 0.56 & 128 & 9.56 $\pm$ 0.24 & 96 \\
\midrule
SC + Ours & 53.52 $\pm$ 0.10 & 64 & 36.76 $\pm$ 1.00 & 40 & 9.80 $\pm$ 0.37 & 64 \\
$\Delta$ vs.\ SC & +0.19 & 0.0\% & +0.19 & \textbf{-68.8\%} & +0.23 & \textbf{-33.3\%} \\
\midrule
RPC & 53.36 $\pm$ 0.11 & 64 & 36.86 $\pm$ 0.57 & 112 & -  & - \\
$\Delta$ vs.\ SC & +0.03 & 0.0\% & +0.29 & \textbf{-12.5\%} & - & - \\
\midrule
RPC + Ours & 53.52 $\pm$ 0.11 & 64 & 36.58 $\pm$ 0.95 & 40 & 9.70 $\pm$ 0.30 & 64 \\
$\Delta$ vs.\ SC & +0.19 & 0.0\% & +0.01 & \textbf{-68.8\%} & +0.14 & \textbf{-33.3\%} \\
\bottomrule
\end{tabular*}
\end{table*}

\subsection{Complete Scaling Results with Varying \texorpdfstring{$K$}{K}}

\Cref{tab:k_scaling_internlm_18b_all,tab:k_scaling_internlm_7b_all} report the complete current-method curves for InternLM2-Math-Plus-1.8B and InternLM2-Math-Plus-7B. We use $K\in\{1,2,4,8,16,32,40,48,64,80,96,112,128\}$, capped at 64 for MATH and 128 for MathOdyssey and AIME by the evaluation protocol. All methods use the same candidate subsets at each budgets. Here \textbf{+ Ours} denotes $\mathsf{PGLA}$ with the voting-weighted comparative advantage $s_u^{\mathrm{Pair}}$ in \Cref{eq:scpa_correction}.

\Cref{tab:k_scaling_internlm_18b_all,tab:k_scaling_internlm_7b_all} show that $\mathsf{PGLA}$ generally improves SC and RPC across sampling budgets. The gains become clearer at moderate and large $K$, especially on MATH and MathOdyssey, indicating that pairwise evidence helps make better use of additional reasoning paths.

\begin{table*}[p]
\centering
\small
\caption{Accuracy across sampling budgets for \textbf{InternLM2-Math-Plus-1.8B} on MATH, MathOdyssey, and
AIME.}
\label{tab:k_scaling_internlm_18b_all}
\renewcommand{\arraystretch}{0.95}
\setlength{\tabcolsep}{4pt}
\begin{tabular*}{\textwidth}{@{\extracolsep{\fill}}rccccc@{}}
\toprule
\multicolumn{6}{c}{\textbf{MATH}} \\
\cmidrule(lr){1-6}
\textbf{$K$} & \textbf{PPL} & \textbf{SC} & \textbf{SC + Ours} & \textbf{RPC} & \textbf{RPC + Ours} \\
\midrule
1 & \textbf{26.94} $\pm$ 0.49 & \textbf{26.94} $\pm$ 0.49 & \textbf{26.94} $\pm$ 0.49 & \textbf{26.94} $\pm$ 0.49 & \textbf{26.94} $\pm$ 0.49 \\
2 & \textbf{29.62} $\pm$ 0.47 & 27.13 $\pm$ 0.44 & \textbf{29.62} $\pm$ 0.47 & \textbf{29.62} $\pm$ 0.47 & \textbf{29.62} $\pm$ 0.47 \\
4 & 31.32 $\pm$ 0.44 & 31.01 $\pm$ 0.36 & 31.32 $\pm$ 0.42 & \textbf{32.36} $\pm$ 0.41 & 31.30 $\pm$ 0.42 \\
8 & 32.53 $\pm$ 0.42 & 33.93 $\pm$ 0.26 & \textbf{34.81} $\pm$ 0.46 & 34.78 $\pm$ 0.32 & 34.34 $\pm$ 0.47 \\
16 & 33.15 $\pm$ 0.42 & 35.49 $\pm$ 0.28 & \textbf{37.76} $\pm$ 0.34 & 36.08 $\pm$ 0.30 & 37.67 $\pm$ 0.36 \\
32 & 33.58 $\pm$ 0.22 & 36.27 $\pm$ 0.22 & 39.04 $\pm$ 0.28 & 37.09 $\pm$ 0.20 & \textbf{39.13} $\pm$ 0.38 \\
40 & 33.43 $\pm$ 0.22 & 36.34 $\pm$ 0.12 & 39.35 $\pm$ 0.25 & 37.40 $\pm$ 0.20 & \textbf{39.50} $\pm$ 0.33 \\
48 & 33.39 $\pm$ 0.28 & 36.37 $\pm$ 0.19 & 39.53 $\pm$ 0.24 & 37.54 $\pm$ 0.21 & \textbf{39.99} $\pm$ 0.28 \\
64 & 33.23 $\pm$ 0.25 & 36.48 $\pm$ 0.14 & 39.77 $\pm$ 0.17 & 37.87 $\pm$ 0.14 & \textbf{40.29} $\pm$ 0.27 \\
\midrule
\multicolumn{6}{c}{\textbf{MathOdyssey}} \\
\cmidrule(lr){1-6}
\textbf{$K$} & \textbf{PPL} & \textbf{SC} & \textbf{SC + Ours} & \textbf{RPC} & \textbf{RPC + Ours} \\
\midrule
1 & \textbf{11.41} $\pm$ 1.23 & \textbf{11.41} $\pm$ 1.23 & \textbf{11.41} $\pm$ 1.23 & \textbf{11.41} $\pm$ 1.23 & \textbf{11.41} $\pm$ 1.23 \\
2 & \textbf{13.26} $\pm$ 1.87 & 11.47 $\pm$ 1.04 & \textbf{13.26} $\pm$ 1.87 & \textbf{13.26} $\pm$ 1.87 & \textbf{13.26} $\pm$ 1.87 \\
4 & 13.91 $\pm$ 1.52 & 13.21 $\pm$ 1.54 & 13.98 $\pm$ 1.53 & 13.91 $\pm$ 1.57 & \textbf{14.04} $\pm$ 1.53 \\
8 & 14.76 $\pm$ 1.26 & 14.43 $\pm$ 0.93 & \textbf{16.53} $\pm$ 0.92 & 15.37 $\pm$ 1.45 & 16.20 $\pm$ 0.82 \\
16 & 15.35 $\pm$ 1.29 & 14.84 $\pm$ 0.65 & \textbf{17.71} $\pm$ 1.08 & 15.30 $\pm$ 0.89 & 17.51 $\pm$ 1.11 \\
32 & 15.94 $\pm$ 1.69 & 14.52 $\pm$ 1.05 & 18.25 $\pm$ 0.93 & 15.45 $\pm$ 1.01 & \textbf{18.64} $\pm$ 1.05 \\
40 & 16.38 $\pm$ 0.76 & 14.70 $\pm$ 0.92 & 18.11 $\pm$ 0.56 & 15.68 $\pm$ 1.07 & \textbf{18.59} $\pm$ 1.44 \\
48 & 16.09 $\pm$ 0.64 & 14.91 $\pm$ 0.95 & 18.33 $\pm$ 0.92 & 15.99 $\pm$ 1.09 & \textbf{19.41} $\pm$ 1.28 \\
64 & 16.71 $\pm$ 1.12 & 14.76 $\pm$ 0.90 & 18.41 $\pm$ 0.69 & 16.22 $\pm$ 0.82 & \textbf{19.64} $\pm$ 0.91 \\
80 & 16.35 $\pm$ 1.35 & 14.38 $\pm$ 0.84 & 18.48 $\pm$ 0.81 & 15.89 $\pm$ 0.52 & \textbf{19.92} $\pm$ 0.87 \\
96 & 16.35 $\pm$ 1.08 & 14.40 $\pm$ 0.76 & 18.23 $\pm$ 0.60 & 16.02 $\pm$ 0.61 & \textbf{19.61} $\pm$ 0.74 \\
112 & 16.58 $\pm$ 1.10 & 14.69 $\pm$ 0.85 & 18.41 $\pm$ 0.67 & 16.32 $\pm$ 0.40 & \textbf{20.36} $\pm$ 0.61 \\
128 & 16.63 $\pm$ 0.83 & 14.57 $\pm$ 0.45 & 18.28 $\pm$ 0.56 & 16.43 $\pm$ 0.60 & \textbf{20.28} $\pm$ 0.48 \\
\midrule
\multicolumn{6}{c}{\textbf{AIME}} \\
\cmidrule(lr){1-6}
\textbf{$K$} & \textbf{PPL} & \textbf{SC} & \textbf{SC + Ours} & \textbf{RPC} & \textbf{RPC + Ours} \\
\midrule
1 & \textbf{1.50} $\pm$ 0.37 & \textbf{1.50} $\pm$ 0.37 & \textbf{1.50} $\pm$ 0.37 & \textbf{1.50} $\pm$ 0.37 & \textbf{1.50} $\pm$ 0.37 \\
2 & \textbf{1.56} $\pm$ 0.28 & 1.41 $\pm$ 0.20 & \textbf{1.56} $\pm$ 0.28 & \textbf{1.56} $\pm$ 0.28 & \textbf{1.56} $\pm$ 0.28 \\
4 & 1.70 $\pm$ 0.35 & 1.68 $\pm$ 0.24 & 1.70 $\pm$ 0.36 & \textbf{1.77} $\pm$ 0.39 & 1.69 $\pm$ 0.36 \\
8 & 1.78 $\pm$ 0.27 & 2.01 $\pm$ 0.35 & 1.99 $\pm$ 0.39 & \textbf{2.14} $\pm$ 0.51 & 1.92 $\pm$ 0.32 \\
16 & 1.88 $\pm$ 0.30 & 2.33 $\pm$ 0.26 & \textbf{2.45} $\pm$ 0.34 & 2.37 $\pm$ 0.36 & 2.37 $\pm$ 0.35 \\
32 & 1.64 $\pm$ 0.24 & 2.40 $\pm$ 0.29 & \textbf{2.76} $\pm$ 0.34 & 2.71 $\pm$ 0.39 & 2.64 $\pm$ 0.31 \\
40 & 1.75 $\pm$ 0.28 & 2.46 $\pm$ 0.29 & \textbf{2.98} $\pm$ 0.38 & 2.77 $\pm$ 0.28 & 2.86 $\pm$ 0.36 \\
48 & 1.80 $\pm$ 0.22 & 2.50 $\pm$ 0.25 & \textbf{3.02} $\pm$ 0.23 & 2.93 $\pm$ 0.28 & 3.00 $\pm$ 0.36 \\
64 & 1.60 $\pm$ 0.26 & 2.56 $\pm$ 0.17 & 3.10 $\pm$ 0.20 & 2.94 $\pm$ 0.25 & \textbf{3.18} $\pm$ 0.29 \\
80 & 1.61 $\pm$ 0.26 & 2.58 $\pm$ 0.16 & 3.30 $\pm$ 0.30 & 3.19 $\pm$ 0.19 & \textbf{3.38} $\pm$ 0.28 \\
96 & 1.67 $\pm$ 0.27 & 2.63 $\pm$ 0.23 & 3.37 $\pm$ 0.15 & 3.20 $\pm$ 0.22 & \textbf{3.45} $\pm$ 0.17 \\
112 & 1.61 $\pm$ 0.27 & 2.67 $\pm$ 0.21 & 3.37 $\pm$ 0.13 & 3.27 $\pm$ 0.18 & \textbf{3.54} $\pm$ 0.24 \\
128 & 1.66 $\pm$ 0.15 & 2.66 $\pm$ 0.20 & 3.37 $\pm$ 0.17 & 3.28 $\pm$ 0.23 & \textbf{3.57} $\pm$ 0.21 \\
\bottomrule
\end{tabular*}
\end{table*}

\begin{table*}[p]
\centering
\small
\caption{Accuracy across sampling budgets for \textbf{InternLM2-Math-Plus-7B} on MATH, MathOdyssey, and
AIME.}
\label{tab:k_scaling_internlm_7b_all}
\renewcommand{\arraystretch}{0.95}
\setlength{\tabcolsep}{4pt}
\begin{tabular*}{\textwidth}{@{\extracolsep{\fill}}rccccc@{}}
\toprule
\multicolumn{6}{c}{\textbf{MATH}} \\
\cmidrule(lr){1-6}
\textbf{$K$} & \textbf{PPL} & \textbf{SC} & \textbf{SC + Ours} & \textbf{RPC} & \textbf{RPC + Ours} \\
\midrule
1 & \textbf{41.69} $\pm$ 0.42 & \textbf{41.69} $\pm$ 0.42 & \textbf{41.69} $\pm$ 0.42 & \textbf{41.69} $\pm$ 0.42 & \textbf{41.69} $\pm$ 0.42 \\
2 & \textbf{43.93} $\pm$ 0.39 & 41.73 $\pm$ 0.22 & \textbf{43.93} $\pm$ 0.39 & \textbf{43.93} $\pm$ 0.39 & \textbf{43.93} $\pm$ 0.39 \\
4 & 45.57 $\pm$ 0.38 & 45.92 $\pm$ 0.26 & 45.51 $\pm$ 0.37 & \textbf{47.09} $\pm$ 0.33 & 45.48 $\pm$ 0.38 \\
8 & 46.49 $\pm$ 0.42 & 48.45 $\pm$ 0.21 & 49.06 $\pm$ 0.51 & \textbf{49.10} $\pm$ 0.18 & 48.54 $\pm$ 0.46 \\
16 & 47.14 $\pm$ 0.22 & 49.76 $\pm$ 0.16 & \textbf{52.07} $\pm$ 0.19 & 50.24 $\pm$ 0.13 & 51.80 $\pm$ 0.17 \\
32 & 47.09 $\pm$ 0.26 & 50.38 $\pm$ 0.26 & \textbf{53.40} $\pm$ 0.28 & 51.17 $\pm$ 0.29 & 53.18 $\pm$ 0.18 \\
40 & 46.96 $\pm$ 0.29 & 50.48 $\pm$ 0.10 & \textbf{53.62} $\pm$ 0.22 & 51.22 $\pm$ 0.14 & 53.31 $\pm$ 0.27 \\
48 & 46.96 $\pm$ 0.26 & 50.51 $\pm$ 0.17 & \textbf{53.70} $\pm$ 0.20 & 51.40 $\pm$ 0.16 & 53.58 $\pm$ 0.21 \\
64 & 47.00 $\pm$ 0.20 & 50.57 $\pm$ 0.18 & \textbf{53.99} $\pm$ 0.18 & 51.94 $\pm$ 0.14 & 53.81 $\pm$ 0.12 \\
\midrule
\multicolumn{6}{c}{\textbf{MathOdyssey}} \\
\cmidrule(lr){1-6}
\textbf{$K$} & \textbf{PPL} & \textbf{SC} & \textbf{SC + Ours} & \textbf{RPC} & \textbf{RPC + Ours} \\
\midrule
1 & \textbf{20.57} $\pm$ 1.13 & \textbf{20.57} $\pm$ 1.13 & \textbf{20.57} $\pm$ 1.13 & \textbf{20.57} $\pm$ 1.13 & \textbf{20.57} $\pm$ 1.13 \\
2 & \textbf{23.42} $\pm$ 1.17 & 20.22 $\pm$ 0.65 & \textbf{23.42} $\pm$ 1.17 & \textbf{23.42} $\pm$ 1.17 & \textbf{23.42} $\pm$ 1.17 \\
4 & 25.40 $\pm$ 0.90 & 23.77 $\pm$ 0.96 & \textbf{25.42} $\pm$ 0.97 & 25.27 $\pm$ 1.43 & 25.37 $\pm$ 1.16 \\
8 & 27.51 $\pm$ 0.84 & 25.60 $\pm$ 0.80 & \textbf{29.36} $\pm$ 1.19 & 27.15 $\pm$ 1.08 & 29.10 $\pm$ 1.12 \\
16 & 27.92 $\pm$ 1.33 & 26.69 $\pm$ 1.09 & \textbf{32.12} $\pm$ 1.01 & 27.94 $\pm$ 1.08 & 31.88 $\pm$ 1.04 \\
32 & 28.33 $\pm$ 0.73 & 27.25 $\pm$ 1.01 & 34.14 $\pm$ 1.30 & 29.23 $\pm$ 1.51 & \textbf{34.29} $\pm$ 1.31 \\
40 & 28.66 $\pm$ 1.21 & 27.58 $\pm$ 0.83 & \textbf{34.73} $\pm$ 1.01 & 29.67 $\pm$ 1.36 & 34.40 $\pm$ 0.90 \\
48 & 28.97 $\pm$ 1.49 & 27.79 $\pm$ 0.74 & 34.81 $\pm$ 1.00 & 29.92 $\pm$ 0.89 & \textbf{35.04} $\pm$ 0.92 \\
64 & 28.59 $\pm$ 1.35 & 27.65 $\pm$ 0.70 & 34.96 $\pm$ 0.69 & 30.93 $\pm$ 1.36 & \textbf{35.60} $\pm$ 1.07 \\
80 & 28.17 $\pm$ 0.96 & 27.76 $\pm$ 0.44 & 35.01 $\pm$ 0.72 & 31.34 $\pm$ 0.77 & \textbf{35.81} $\pm$ 0.81 \\
96 & 28.30 $\pm$ 1.01 & 27.99 $\pm$ 0.51 & 35.14 $\pm$ 0.85 & 31.23 $\pm$ 0.76 & \textbf{36.20} $\pm$ 0.52 \\
112 & 27.61 $\pm$ 0.73 & 28.32 $\pm$ 0.55 & 35.17 $\pm$ 0.57 & 31.54 $\pm$ 0.61 & \textbf{36.56} $\pm$ 0.68 \\
128 & 27.51 $\pm$ 1.18 & 28.30 $\pm$ 0.61 & 35.30 $\pm$ 0.55 & 31.70 $\pm$ 0.71 & \textbf{36.53} $\pm$ 1.06 \\
\midrule
\multicolumn{6}{c}{\textbf{AIME}} \\
\cmidrule(lr){1-6}
\textbf{$K$} & \textbf{PPL} & \textbf{SC} & \textbf{SC + Ours} & \textbf{RPC} & \textbf{RPC + Ours} \\
\midrule
1 & \textbf{5.06} $\pm$ 0.30 & \textbf{5.06} $\pm$ 0.30 & \textbf{5.06} $\pm$ 0.30 & \textbf{5.06} $\pm$ 0.30 & \textbf{5.06} $\pm$ 0.30 \\
2 & \textbf{5.66} $\pm$ 0.54 & 5.19 $\pm$ 0.26 & \textbf{5.66} $\pm$ 0.54 & \textbf{5.66} $\pm$ 0.54 & \textbf{5.66} $\pm$ 0.54 \\
4 & 5.93 $\pm$ 0.35 & 6.23 $\pm$ 0.31 & 5.92 $\pm$ 0.34 & \textbf{6.57} $\pm$ 0.35 & 5.89 $\pm$ 0.37 \\
8 & 6.14 $\pm$ 0.48 & 7.49 $\pm$ 0.67 & 7.15 $\pm$ 0.61 & \textbf{7.86} $\pm$ 0.68 & 6.86 $\pm$ 0.69 \\
16 & 6.21 $\pm$ 0.72 & 8.51 $\pm$ 0.35 & 8.64 $\pm$ 0.35 & \textbf{8.71} $\pm$ 0.43 & 8.48 $\pm$ 0.41 \\
32 & 6.31 $\pm$ 0.60 & 8.89 $\pm$ 0.34 & \textbf{9.44} $\pm$ 0.39 & 9.10 $\pm$ 0.33 & 9.12 $\pm$ 0.33 \\
40 & 6.21 $\pm$ 0.72 & 9.04 $\pm$ 0.26 & \textbf{9.56} $\pm$ 0.41 & 9.37 $\pm$ 0.44 & 9.39 $\pm$ 0.28 \\
48 & 6.37 $\pm$ 0.69 & 9.05 $\pm$ 0.30 & \textbf{9.63} $\pm$ 0.34 & 9.51 $\pm$ 0.44 & 9.60 $\pm$ 0.42 \\
64 & 6.48 $\pm$ 0.43 & 9.24 $\pm$ 0.21 & 9.68 $\pm$ 0.44 & 9.54 $\pm$ 0.33 & \textbf{9.97} $\pm$ 0.35 \\
80 & 6.41 $\pm$ 0.49 & 9.33 $\pm$ 0.22 & 9.66 $\pm$ 0.29 & 9.69 $\pm$ 0.38 & \textbf{9.92} $\pm$ 0.45 \\
96 & 6.28 $\pm$ 0.43 & 9.31 $\pm$ 0.12 & 9.87 $\pm$ 0.32 & 9.86 $\pm$ 0.21 & \textbf{10.05} $\pm$ 0.34 \\
112 & 6.11 $\pm$ 0.44 & 9.39 $\pm$ 0.21 & 9.87 $\pm$ 0.35 & 9.69 $\pm$ 0.15 & \textbf{10.09} $\pm$ 0.38 \\
128 & 5.96 $\pm$ 0.48 & 9.40 $\pm$ 0.21 & 9.95 $\pm$ 0.24 & 9.74 $\pm$ 0.23 & \textbf{10.08} $\pm$ 0.24 \\
\bottomrule
\end{tabular*}
\end{table*}

\subsection{Online Sampling Efficiency}
\label{app:online_sampling_efficiency}

The full-budget results establish the aggregation benefit of $\mathsf{PGLA}$.
We next examine whether this benefit remains when the sampling budget is
determined independently for each query. Following
\Cref{app:online_pgla}, we apply the same stopping rule and hyperparameters to
SC and SC+$\mathsf{PGLA}$ on identical nested path prefixes, with each rule
tracking its own unique winner. Both methods update their predictions every
eight paths and stop after the winner remains unchanged for three consecutive
updates. We impose no additional warm-up threshold, making $K=24$ the
earliest possible stopping point. To isolate the effect of $\mathsf{PGLA}$, its
online variant uses count-based SC support, i.e., $s_{u,j}^{\mathrm{SC}}=|C_{u,j}|$. We fix
this protocol across the evaluated datasets and include Adaptive SC with its original
stopping criterion as a representative online baseline.


\begin{table*}[t]
\centering
\caption{Online aggregation across the two InternLM2-Math-Plus backbones. $\overline K$ is the average number of sampled paths. Gain is
the absolute accuracy change (\%) against Online SC under the same stopping rule, and saving is
relative to $K_{\max}$, which is 64 for MATH and 128 otherwise.}
\label{tab:online_sc_pga_combined}
\small
\setlength{\tabcolsep}{4pt}
\begin{tabular*}{\textwidth}{@{\extracolsep{\fill}}llcccc@{}}
\toprule
\multirow{2}{*}{\textbf{Dataset}} & \multirow{2}{*}{\textbf{Method}} & \multicolumn{2}{c}{\textbf{Accuracy}} & \multicolumn{2}{c}{\textbf{Sampling Cost}} \\
\cmidrule(lr){3-4}\cmidrule(lr){5-6}
& & \textbf{Acc.} & \makecell{\textbf{Gain over}\\\textbf{Online SC}} & $\overline K$ & \textbf{Saving} \\
\midrule
\rowcolor{gray!20}\multicolumn{6}{c}{\textbf{InternLM2-Math-Plus-1.8B}} \\
\multirow{4}{*}{MATH} & SC (Full Budget) & $36.48_{\pm 0.14}$ & -- & 64.0 & -- \\
 & Adaptive SC & $36.49_{\pm 0.16}$ & -- & 36.3 & 43.3\% \\
 & Online SC & $36.38_{\pm 0.19}$ & -- & 33.4 & 47.7\% \\
 & \textbf{Online SC + $\mathsf{PGLA}$} & $\mathbf{39.32}_{\pm 0.31}$ & $\mathbf{+2.95}$ & 31.5 & 50.8\% \\
\addlinespace[2pt]
\multirow{4}{*}{MathOdyssey} & SC (Full Budget) & $14.57_{\pm 0.45}$ & -- & 128.0 & -- \\
 & Adaptive SC & $14.48_{\pm 0.46}$ & -- & 70.0 & 45.3\% \\
 & Online SC & $14.47_{\pm 0.93}$ & -- & 37.4 & 70.8\% \\
 & \textbf{Online SC + $\mathsf{PGLA}$} & $\mathbf{18.20}_{\pm 0.90}$ & $\mathbf{+3.73}$ & 34.5 & 73.1\% \\
\addlinespace[2pt]
\multirow{4}{*}{AIME} & SC (Full Budget) & $2.66_{\pm 0.20}$ & -- & 128.0 & -- \\
 & Adaptive SC & $2.66_{\pm 0.19}$ & -- & 92.6 & 27.7\% \\
 & Online SC & $2.55_{\pm 0.24}$ & -- & 45.4 & 64.5\% \\
 & \textbf{Online SC + $\mathsf{PGLA}$} & $\mathbf{2.94}_{\pm 0.29}$ & $\mathbf{+0.39}$ & 38.3 & 70.1\% \\
\addlinespace[3pt]
\rowcolor{gray!20}\multicolumn{6}{c}{\textbf{InternLM2-Math-Plus-7B}} \\
\multirow{4}{*}{MATH} & SC (Full Budget) & $50.57_{\pm 0.18}$ & -- & 64.0 & -- \\
 & Adaptive SC & $50.55_{\pm 0.22}$ & -- & 27.2 & 57.5\% \\
 & Online SC & $50.51_{\pm 0.20}$ & -- & 30.2 & 52.8\% \\
 & \textbf{Online SC + $\mathsf{PGLA}$} & $\mathbf{53.49}_{\pm 0.22}$ & $\mathbf{+2.98}$ & 29.0 & 54.7\% \\
\addlinespace[2pt]
\multirow{4}{*}{MathOdyssey} & SC (Full Budget) & $28.30_{\pm 0.61}$ & -- & 128.0 & -- \\
 & Adaptive SC & $28.22_{\pm 0.67}$ & -- & 58.8 & 54.1\% \\
 & Online SC & $27.79_{\pm 0.86}$ & -- & 35.3 & 72.4\% \\
 & \textbf{Online SC + $\mathsf{PGLA}$} & $\mathbf{34.19}_{\pm 0.94}$ & $\mathbf{+6.40}$ & 32.0 & 75.0\% \\
\addlinespace[2pt]
\multirow{4}{*}{AIME} & SC (Full Budget) & $9.40_{\pm 0.21}$ & -- & 128.0 & -- \\
 & Adaptive SC & $9.38_{\pm 0.23}$ & -- & 81.6 & 36.3\% \\
 & Online SC & $9.04_{\pm 0.41}$ & -- & 42.0 & 67.2\% \\
 & \textbf{Online SC + $\mathsf{PGLA}$} & $\mathbf{9.50}_{\pm 0.42}$ & $\mathbf{+0.46}$ & 36.3 & 71.7\% \\
\bottomrule
\end{tabular*}
\end{table*}

\textbf{Performance in the Online Setting.}
As shown in \Cref{tab:online_sc_pga_combined}, Online SC+$\mathsf{PGLA}$
consistently outperforms Online SC across the evaluated datasets,
yielding gains of $0.39$--$6.40\%$ under the same stopping
rule. Meanwhile, it reduces the sampling cost by 50.8\%--75.0\% and uses fewer
paths than Online SC in every setting. Despite this reduction, Online
SC+$\mathsf{PGLA}$ remains $0.10$--$5.89\%$ above full-budget SC. These results
show that the comparative evidence introduced by $\mathsf{PGLA}$ improves both the
accuracy and the sampling efficiency of online aggregation.

\clearpage
\subsection{Extension to \texorpdfstring{$32$}{32}B Models}
\label{app:qwen32_extension}

We further evaluate the aggregation methods in \Cref{tab:reasoning}
using Qwen3-32B in reasoning mode on AIME24, AIME25, and HMMT25 (February).
Each benchmark contains 30 problems with 64 sampled responses per problem.
All methods use the same $K=32$ subsets.

\Cref{tab:qwen32_extension} shows that $\mathsf{PGLA}$ matches or improves the corresponding voting baselines on Qwen3-32B, with the most gains on HMMT25. These results suggest that the benefits of pairwise confidence comparisons extend to larger reasoning models.

\begin{table}[!ht]
    \centering
    \caption{Accuracy (\%) of Qwen3-32B in reasoning mode at $K=32$.}
    \label{tab:qwen32_extension}
    \fontsize{7}{8.1}\selectfont
    \renewcommand{\arraystretch}{0.96}
    \setlength{\tabcolsep}{2pt}
    \begin{tabular}{@{\hspace{2pt}}p{\dimexpr0.43\textwidth-4pt\relax}*{3}{>{\centering\arraybackslash}p{\dimexpr0.19\textwidth-4pt\relax}}@{\hspace{2pt}}}
        \toprule
        \textbf{Method} & \textbf{AIME24} & \textbf{AIME25} & \textbf{HMMT25} \\
        \addlinespace[2pt]
        \rowcolor{gray!14} \multicolumn{4}{l}{\strut\hspace{1pt}\textbf{Qwen3-32B Reasoning mode}} \\
        \addlinespace[1.5pt]
        PPL Best-of-$N$ & $75.67_{\pm 3.96}$ & $67.33_{\pm 6.29}$ & $41.00_{\pm 3.96}$ \\
        DeepConf Best-of-$N$ & $84.33_{\pm 3.96}$ & $79.33_{\pm 2.00}$ & $55.33_{\pm 2.67}$ \\
        Self-Certainty Best-of-$N$ & $88.33_{\pm 2.69}$ & $77.00_{\pm 2.33}$ & $62.33_{\pm 3.00}$ \\
        \addlinespace[1.5pt]
        SC & $90.17_{\pm 1.57}$ & $81.33_{\pm 1.63}$ & $59.50_{\pm 2.67}$ \\
        \rowcolor{gray!5} \hspace{7pt}\textbf{+ Ours (DeepConf)} & $\mathbf{90.67}_{\pm 2.00}$ & $\mathbf{81.67}_{\pm 1.67}$ & $63.67_{\pm 2.77}$ \\
        \rowcolor{gray!5} \hspace{7pt}\textbf{+ Ours (Self-Certainty)} & $90.33_{\pm 1.80}$ & $81.33_{\pm 1.63}$ & $\mathbf{65.00}_{\pm 2.24}$ \\
        \addlinespace[1.5pt]
        DeepConf-Weighted SC & $90.33_{\pm 1.80}$ & $81.33_{\pm 1.63}$ & $60.67_{\pm 2.91}$ \\
        \rowcolor{gray!5} \hspace{7pt}\textbf{+ Ours} & $\mathbf{90.67}_{\pm 2.00}$ & $\mathbf{81.67}_{\pm 1.67}$ & $\mathbf{63.67}_{\pm 1.80}$ \\
        \addlinespace[1.5pt]
        Self-Certainty-Weighted SC & $\mathbf{90.33}_{\pm 1.80}$ & $\mathbf{81.33}_{\pm 1.63}$ & $60.67_{\pm 3.27}$ \\
        \rowcolor{gray!5} \hspace{7pt}\textbf{+ Ours} & $\mathbf{90.33}_{\pm 1.80}$ & $\mathbf{81.33}_{\pm 1.63}$ & $\mathbf{65.00}_{\pm 1.67}$ \\
        \addlinespace[1.5pt]
        Self-Certainty Borda & $89.00_{\pm 2.13}$ & $80.33_{\pm 1.00}$ & $64.33_{\pm 3.00}$ \\
        \rowcolor{gray!5} \hspace{7pt}\textbf{+ Ours} & $\mathbf{89.33}_{\pm 2.00}$ & $\mathbf{81.00}_{\pm 1.53}$ & $\mathbf{65.00}_{\pm 2.24}$ \\
        \addlinespace[1.5pt]
        DeepConf-Filter (90\%) & $90.33_{\pm 1.80}$ & $81.33_{\pm 1.63}$ & $64.00_{\pm 3.27}$ \\
        \rowcolor{gray!5} \hspace{7pt}\textbf{+ Ours} & $\mathbf{90.67}_{\pm 2.00}$ & $\mathbf{82.00}_{\pm 1.63}$ & $\mathbf{65.00}_{\pm 2.69}$ \\
        \bottomrule
    \end{tabular}
\end{table}

\clearpage
\subsection{Effect of Comparative Advantage Rules}
\label{app:comparative_rules}

The comparative term in $\mathsf{PGLA}$ is modular: changing its aggregation
rule leaves the voting distribution and KL-regularized update unchanged.
For $U=|\mathcal A|>1$, we compare the default voting-weighted comparative
advantage in \Cref{eq:scpa_correction} with three alternatives:
\begin{equation}
\begin{aligned}
 s_u^{\mathrm{Pair}}&=
 \frac{\sum_{v\ne u}s_v^{\mathcal V}Y_{uv}}{\sum_{v\ne u}s_v^{\mathcal V}},
 &s_u^{\mathrm{Pair},\mathrm{Unif}}&=\frac{1}{U-1}\sum_{v\ne u}Y_{uv},\\
 s_u^{\mathrm{Pair},\mathrm{Copeland}}&=\frac{1}{U-1}\sum_{v\ne u}\operatorname{sgn}(Y_{uv}),
 &s_u^{\mathrm{Pair},\mathrm{Maximin}}&=\min_{v\ne u}Y_{uv}.
\end{aligned}
\label{eq:comparative_rule_variants}
\end{equation}
Uniform averaging takes the mean of signed pairwise scores;
Copeland retains only their signs, with ties contributing zero~\citep{zwicker2016voting};
Maximin retains the smallest signed margin~\citep{fischer2016weighted}.
All four scores lie in $[-1,1]$ and are set to zero for $U=1$.
The weighted score is also zero when its denominator vanishes.
These rules aggregate confidence-based comparisons, and the default weights
apply to opposing answers. Each alternative can replace $s_u^{\mathrm{Pair}}$
in \Cref{eq:scpa_distribution}.

\paragraph{Matched protocol.}
We extend \Cref{tab:main_results} using InternLM2-Math-Plus-1.8B, InternLM2-Math-Plus-7B, and
DeepSeek-Math-RL-7B on MATH ($K=64$), MathOdyssey ($K=128$), and AIME
($K=128$), using internal mean log-probability scores. We evaluate SC,
Weighted SC, Adaptive SC, and RPC. Only the comparative rule changes.

\paragraph{Results.}
\Cref{tab:comparative_rules_internlm18,tab:comparative_rules_internlm7,tab:comparative_rules_deepseek7}
report the results by model. Voting-weighted comparisons attain higher mean accuracy
than uniform averaging in most settings, with macro-average accuracy
of $29.06\%$ versus $28.70\%$, but uniform averaging yields lower macro-average
ECE. Copeland can outperform voting-weighted
comparisons and has slightly higher model-average accuracy on InternLM2-1.8B,
but lower averages on the other models. Maximin is consistently less accurate.
Overall, all four rules improve accuracy over their corresponding baselines in
most evaluated settings, supporting the effectiveness of incorporating comparative
advantage and demonstrating that our general framework accommodates diverse
comparative advantage rules.


\begin{table}[!ht]
\centering
\caption{Comparative-advantage functions on \textbf{InternLM2-Math-Plus-1.8B}. We expand
\Cref{tab:main_results} using the same coefficients (\Cref{app:comparative_rules}).  Voting denotes the default
voting-weighted comparative advantage and \textbf{+ Ours} denotes the scaled $\mathsf{PGLA}$ variants
described in \Cref{app:comparative_rules}.}
\label{tab:comparative_rules_internlm18}
\vskip -3pt\relax
\fontsize{7.8}{8.3}\selectfont
\renewcommand{\arraystretch}{0.99}
\setlength{\tabcolsep}{2pt}
\begin{tabular}{@{\hspace{2pt}}p{\dimexpr\textwidth-324pt\relax}*{6}{>{\centering\arraybackslash}p{40.5pt}}*{2}{>{\centering\arraybackslash}p{22.5pt}}@{\hspace{2pt}}}
\toprule
\multirow{2}{*}{\textbf{Method}} & \multicolumn{2}{c}{\textbf{MATH}} & \multicolumn{2}{c}{\textbf{MathOdyssey}} & \multicolumn{2}{c}{\textbf{AIME}} & \multicolumn{2}{c}{\textbf{Average}} \\
\cmidrule(lr){2-3} \cmidrule(lr){4-5} \cmidrule(lr){6-7} \cmidrule(lr){8-9}
 & Acc. $\uparrow$ & ECE $\downarrow$ & Acc. $\uparrow$ & ECE $\downarrow$ & Acc. $\uparrow$ & ECE $\downarrow$ & Acc. $\uparrow$ & ECE $\downarrow$ \\
\addlinespace[2pt]
\rowcolor{gray!14} \multicolumn{9}{l}{\strut\hspace{1pt}\textbf{InternLM2-Math-Plus-1.8B}} \\
\addlinespace[1.5pt]
PPL & $33.23_{\pm 0.25}$ & $61.58_{\pm 0.25}$ & $16.63_{\pm 0.83}$ & $77.20_{\pm 0.82}$ & $1.66_{\pm 0.15}$ & $92.28_{\pm 0.16}$ & $17.17$ & $77.02$ \\
\addlinespace[2pt]
SC & $36.48_{\pm 0.14}$ & $6.66_{\pm 0.14}$ & $14.57_{\pm 0.45}$ & $18.51_{\pm 0.41}$ & $2.66_{\pm 0.20}$ & $16.84_{\pm 0.16}$ & $17.90$ & $14.00$ \\
\rowcolor{gray!5} \hspace{5pt}+ Ours (voting) & $\mathbf{39.77}_{\pm 0.17}$ & $5.63_{\pm 0.13}$ & $18.28_{\pm 0.56}$ & $13.97_{\pm 0.59}$ & $3.37_{\pm 0.17}$ & $15.51_{\pm 0.15}$ & $20.47$ & $11.70$ \\
\hspace{5pt}+ Ours (Uniform) & $39.24_{\pm 0.14}$ & $\mathbf{4.94}_{\pm 0.13}$ & $17.46_{\pm 0.61}$ & $14.39_{\pm 0.69}$ & $3.25_{\pm 0.14}$ & $\mathbf{15.34}_{\pm 0.11}$ & $19.98$ & $11.55$ \\
\hspace{5pt}+ Ours (Copeland) & $39.65_{\pm 0.17}$ & $5.71_{\pm 0.16}$ & $\mathbf{18.87}_{\pm 0.62}$ & $\mathbf{13.37}_{\pm 0.58}$ & $\mathbf{3.53}_{\pm 0.12}$ & $15.56_{\pm 0.17}$ & $\mathbf{20.68}$ & $\mathbf{11.55}$ \\
\hspace{5pt}+ Ours (Maximin) & $38.17_{\pm 0.19}$ & $6.19_{\pm 0.26}$ & $16.13_{\pm 0.33}$ & $16.35_{\pm 0.37}$ & $2.78_{\pm 0.25}$ & $16.56_{\pm 0.25}$ & $19.03$ & $13.04$ \\
\addlinespace[2pt]
Weighted SC & $36.72_{\pm 0.12}$ & $6.52_{\pm 0.14}$ & $14.86_{\pm 0.47}$ & $18.05_{\pm 0.50}$ & $2.75_{\pm 0.23}$ & $16.65_{\pm 0.19}$ & $18.11$ & $13.74$ \\
\rowcolor{gray!5} \hspace{5pt}+ Ours (voting) & $\mathbf{39.91}_{\pm 0.22}$ & $5.55_{\pm 0.17}$ & $18.56_{\pm 0.56}$ & $13.47_{\pm 0.56}$ & $3.37_{\pm 0.17}$ & $15.34_{\pm 0.15}$ & $20.61$ & $11.45$ \\
\hspace{5pt}+ Ours (Uniform) & $39.40_{\pm 0.17}$ & $\mathbf{4.83}_{\pm 0.19}$ & $17.53_{\pm 0.73}$ & $14.16_{\pm 0.79}$ & $3.32_{\pm 0.19}$ & $\mathbf{15.11}_{\pm 0.15}$ & $20.09$ & $11.37$ \\
\hspace{5pt}+ Ours (Copeland) & $39.75_{\pm 0.19}$ & $5.69_{\pm 0.17}$ & $\mathbf{19.13}_{\pm 0.48}$ & $\mathbf{13.00}_{\pm 0.42}$ & $\mathbf{3.59}_{\pm 0.13}$ & $15.40_{\pm 0.19}$ & $\mathbf{20.82}$ & $\mathbf{11.36}$ \\
\hspace{5pt}+ Ours (Maximin) & $38.28_{\pm 0.15}$ & $6.21_{\pm 0.24}$ & $16.38_{\pm 0.42}$ & $15.95_{\pm 0.39}$ & $2.83_{\pm 0.20}$ & $16.39_{\pm 0.22}$ & $19.16$ & $12.85$ \\
\addlinespace[2pt]
Adaptive SC & $36.48_{\pm 0.16}$ & $10.61_{\pm 0.23}$ & $14.43_{\pm 0.51}$ & $22.69_{\pm 0.73}$ & $2.65_{\pm 0.20}$ & $19.25_{\pm 0.25}$ & $17.85$ & $17.52$ \\
\rowcolor{gray!5} \hspace{5pt}+ Ours (voting) & $\mathbf{39.05}_{\pm 0.23}$ & $9.28_{\pm 0.27}$ & $17.47_{\pm 0.63}$ & $17.96_{\pm 0.60}$ & $3.23_{\pm 0.21}$ & $17.53_{\pm 0.41}$ & $\mathbf{19.92}$ & $14.92$ \\
\hspace{5pt}+ Ours (Uniform) & $38.74_{\pm 0.20}$ & $\mathbf{8.66}_{\pm 0.24}$ & $16.79_{\pm 0.63}$ & $18.22_{\pm 0.84}$ & $3.14_{\pm 0.14}$ & $\mathbf{17.31}_{\pm 0.31}$ & $19.56$ & $\mathbf{14.73}$ \\
\hspace{5pt}+ Ours (Copeland) & $38.21_{\pm 0.18}$ & $10.26_{\pm 0.18}$ & $\mathbf{17.76}_{\pm 0.80}$ & $\mathbf{17.66}_{\pm 0.90}$ & $\mathbf{3.40}_{\pm 0.19}$ & $17.59_{\pm 0.47}$ & $19.79$ & $15.17$ \\
\hspace{5pt}+ Ours (Maximin) & $36.95_{\pm 0.28}$ & $10.77_{\pm 0.31}$ & $16.18_{\pm 0.88}$ & $19.32_{\pm 1.28}$ & $2.75_{\pm 0.28}$ & $18.46_{\pm 0.44}$ & $18.63$ & $16.18$ \\
\addlinespace[2pt]
RPC & $37.87_{\pm 0.14}$ & $6.45_{\pm 0.15}$ & $16.43_{\pm 0.60}$ & $16.25_{\pm 0.76}$ & $3.28_{\pm 0.23}$ & $16.32_{\pm 0.20}$ & $19.19$ & $13.01$ \\
\rowcolor{gray!5} \hspace{5pt}+ Ours (voting) & $\mathbf{40.29}_{\pm 0.27}$ & $5.50_{\pm 0.22}$ & $20.28_{\pm 0.48}$ & $\mathbf{10.78}_{\pm 0.60}$ & $3.57_{\pm 0.21}$ & $14.67_{\pm 0.23}$ & $\mathbf{21.38}$ & $\mathbf{10.32}$ \\
\hspace{5pt}+ Ours (Uniform) & $39.90_{\pm 0.25}$ & $\mathbf{4.83}_{\pm 0.29}$ & $19.00_{\pm 0.48}$ & $11.89_{\pm 0.55}$ & $3.52_{\pm 0.12}$ & $\mathbf{14.63}_{\pm 0.12}$ & $20.80$ & $10.45$ \\
\hspace{5pt}+ Ours (Copeland) & $40.13_{\pm 0.23}$ & $5.87_{\pm 0.20}$ & $\mathbf{20.33}_{\pm 0.62}$ & $11.31_{\pm 0.86}$ & $\mathbf{3.63}_{\pm 0.09}$ & $15.23_{\pm 0.15}$ & $21.37$ & $10.80$ \\
\hspace{5pt}+ Ours (Maximin) & $38.81_{\pm 0.22}$ & $6.82_{\pm 0.27}$ & $17.92_{\pm 0.61}$ & $14.01_{\pm 0.55}$ & $3.09_{\pm 0.22}$ & $16.37_{\pm 0.23}$ & $19.94$ & $12.40$ \\
\addlinespace[2pt]
\bottomrule
\end{tabular}
\end{table}

\begin{table}[!ht]
\centering
\caption{Comparative-advantage functions on \textbf{InternLM2-Math-Plus-7B}. We expand
\Cref{tab:main_results} using the same coefficients (\Cref{app:comparative_rules}). Voting denotes the default
voting-weighted comparative advantage and \textbf{+ Ours} denotes the scaled $\mathsf{PGLA}$ variants
described in \Cref{app:comparative_rules}.}
\label{tab:comparative_rules_internlm7}
\vskip -3pt\relax
\fontsize{7.8}{8.3}\selectfont
\renewcommand{\arraystretch}{0.99}
\setlength{\tabcolsep}{2pt}
\begin{tabular}{@{\hspace{2pt}}p{\dimexpr\textwidth-324pt\relax}*{6}{>{\centering\arraybackslash}p{40.5pt}}*{2}{>{\centering\arraybackslash}p{22.5pt}}@{\hspace{2pt}}}
\toprule
\multirow{2}{*}{\textbf{Method}} & \multicolumn{2}{c}{\textbf{MATH}} & \multicolumn{2}{c}{\textbf{MathOdyssey}} & \multicolumn{2}{c}{\textbf{AIME}} & \multicolumn{2}{c}{\textbf{Average}} \\
\cmidrule(lr){2-3} \cmidrule(lr){4-5} \cmidrule(lr){6-7} \cmidrule(lr){8-9}
 & Acc. $\uparrow$ & ECE $\downarrow$ & Acc. $\uparrow$ & ECE $\downarrow$ & Acc. $\uparrow$ & ECE $\downarrow$ & Acc. $\uparrow$ & ECE $\downarrow$ \\
\addlinespace[2pt]
\rowcolor{gray!14} \multicolumn{9}{l}{\strut\hspace{1pt}\textbf{InternLM2-Math-Plus-7B}} \\
\addlinespace[1.5pt]
PPL & $47.00_{\pm 0.20}$ & $48.98_{\pm 0.20}$ & $27.51_{\pm 1.18}$ & $67.54_{\pm 1.18}$ & $5.96_{\pm 0.48}$ & $88.98_{\pm 0.49}$ & $26.82$ & $68.50$ \\
\addlinespace[2pt]
SC & $50.57_{\pm 0.18}$ & $6.72_{\pm 0.19}$ & $28.30_{\pm 0.61}$ & $12.18_{\pm 0.57}$ & $9.40_{\pm 0.21}$ & $14.35_{\pm 0.23}$ & $29.42$ & $11.08$ \\
\rowcolor{gray!5} \hspace{5pt}+ Ours (voting) & $\mathbf{53.99}_{\pm 0.18}$ & $6.10_{\pm 0.20}$ & $\mathbf{35.30}_{\pm 0.55}$ & $9.69_{\pm 1.02}$ & $9.95_{\pm 0.24}$ & $13.90_{\pm 0.23}$ & $\mathbf{33.08}$ & $9.90$ \\
\hspace{5pt}+ Ours (Uniform) & $53.72_{\pm 0.13}$ & $\mathbf{5.13}_{\pm 0.23}$ & $33.62_{\pm 0.72}$ & $8.68_{\pm 0.75}$ & $10.04_{\pm 0.23}$ & $\mathbf{13.26}_{\pm 0.25}$ & $32.46$ & $\mathbf{9.02}$ \\
\hspace{5pt}+ Ours (Copeland) & $52.98_{\pm 0.23}$ & $6.54_{\pm 0.25}$ & $34.37_{\pm 0.80}$ & $\mathbf{8.52}_{\pm 1.18}$ & $\mathbf{10.21}_{\pm 0.28}$ & $13.67_{\pm 0.30}$ & $32.52$ & $9.58$ \\
\hspace{5pt}+ Ours (Maximin) & $52.56_{\pm 0.17}$ & $6.25_{\pm 0.20}$ & $31.79_{\pm 0.92}$ & $10.16_{\pm 1.01}$ & $9.17_{\pm 0.19}$ & $14.75_{\pm 0.20}$ & $31.17$ & $10.38$ \\
\addlinespace[2pt]
Weighted SC & $50.81_{\pm 0.20}$ & $6.56_{\pm 0.21}$ & $28.82_{\pm 0.67}$ & $11.57_{\pm 0.68}$ & $9.45_{\pm 0.23}$ & $14.27_{\pm 0.25}$ & $29.69$ & $10.80$ \\
\rowcolor{gray!5} \hspace{5pt}+ Ours (voting) & $\mathbf{53.99}_{\pm 0.13}$ & $6.15_{\pm 0.17}$ & $\mathbf{35.78}_{\pm 0.51}$ & $9.37_{\pm 1.16}$ & $9.88_{\pm 0.28}$ & $13.86_{\pm 0.27}$ & $\mathbf{33.22}$ & $9.79$ \\
\hspace{5pt}+ Ours (Uniform) & $53.82_{\pm 0.15}$ & $\mathbf{5.08}_{\pm 0.29}$ & $34.06_{\pm 0.81}$ & $\mathbf{8.46}_{\pm 0.99}$ & $10.01_{\pm 0.24}$ & $\mathbf{13.21}_{\pm 0.24}$ & $32.63$ & $\mathbf{8.92}$ \\
\hspace{5pt}+ Ours (Copeland) & $53.02_{\pm 0.21}$ & $6.57_{\pm 0.23}$ & $34.55_{\pm 0.75}$ & $8.50_{\pm 1.01}$ & $\mathbf{10.29}_{\pm 0.36}$ & $13.55_{\pm 0.40}$ & $32.62$ & $9.54$ \\
\hspace{5pt}+ Ours (Maximin) & $52.67_{\pm 0.17}$ & $6.20_{\pm 0.19}$ & $32.11_{\pm 0.72}$ & $10.01_{\pm 0.98}$ & $9.22_{\pm 0.18}$ & $14.64_{\pm 0.20}$ & $31.33$ & $10.28$ \\
\addlinespace[2pt]
Adaptive SC & $50.54_{\pm 0.22}$ & $11.02_{\pm 0.23}$ & $28.20_{\pm 0.68}$ & $18.17_{\pm 1.02}$ & $9.38_{\pm 0.23}$ & $17.39_{\pm 0.41}$ & $29.37$ & $15.53$ \\
\rowcolor{gray!5} \hspace{5pt}+ Ours (voting) & $\mathbf{52.66}_{\pm 0.26}$ & $10.53_{\pm 0.35}$ & $\mathbf{33.14}_{\pm 0.73}$ & $\mathbf{13.22}_{\pm 1.52}$ & $9.83_{\pm 0.21}$ & $16.45_{\pm 0.33}$ & $\mathbf{31.87}$ & $13.40$ \\
\hspace{5pt}+ Ours (Uniform) & $52.62_{\pm 0.28}$ & $\mathbf{9.60}_{\pm 0.39}$ & $31.93_{\pm 0.46}$ & $13.31_{\pm 0.98}$ & $\mathbf{9.87}_{\pm 0.19}$ & $\mathbf{15.90}_{\pm 0.34}$ & $31.47$ & $\mathbf{12.94}$ \\
\hspace{5pt}+ Ours (Copeland) & $50.73_{\pm 0.26}$ & $12.43_{\pm 0.31}$ & $31.75_{\pm 1.00}$ & $14.22_{\pm 1.21}$ & $9.73_{\pm 0.28}$ & $16.58_{\pm 0.35}$ & $30.74$ & $14.41$ \\
\hspace{5pt}+ Ours (Maximin) & $50.66_{\pm 0.35}$ & $11.71_{\pm 0.36}$ & $30.73_{\pm 0.64}$ & $14.99_{\pm 1.19}$ & $8.57_{\pm 0.27}$ & $17.83_{\pm 0.22}$ & $29.99$ & $14.85$ \\
\addlinespace[2pt]
RPC & $51.94_{\pm 0.14}$ & $6.41_{\pm 0.16}$ & $31.70_{\pm 0.71}$ & $9.88_{\pm 0.80}$ & $9.74_{\pm 0.23}$ & $14.31_{\pm 0.21}$ & $31.13$ & $10.20$ \\
\rowcolor{gray!5} \hspace{5pt}+ Ours (voting) & $\mathbf{53.81}_{\pm 0.12}$ & $6.72_{\pm 0.15}$ & $\mathbf{36.53}_{\pm 1.06}$ & $7.90_{\pm 1.09}$ & $10.08_{\pm 0.24}$ & $13.31_{\pm 0.22}$ & $\mathbf{33.47}$ & $9.31$ \\
\hspace{5pt}+ Ours (Uniform) & $53.80_{\pm 0.10}$ & $\mathbf{5.74}_{\pm 0.26}$ & $35.19_{\pm 0.59}$ & $8.22_{\pm 0.93}$ & $10.14_{\pm 0.32}$ & $\mathbf{12.98}_{\pm 0.33}$ & $33.04$ & $\mathbf{8.98}$ \\
\hspace{5pt}+ Ours (Copeland) & $52.83_{\pm 0.19}$ & $7.36_{\pm 0.21}$ & $35.12_{\pm 1.04}$ & $\mathbf{7.87}_{\pm 1.59}$ & $\mathbf{10.20}_{\pm 0.32}$ & $13.70_{\pm 0.38}$ & $32.72$ & $9.64$ \\
\hspace{5pt}+ Ours (Maximin) & $52.79_{\pm 0.22}$ & $6.97_{\pm 0.23}$ & $33.73_{\pm 0.79}$ & $9.72_{\pm 1.34}$ & $9.68_{\pm 0.21}$ & $14.45_{\pm 0.23}$ & $32.07$ & $10.38$ \\
\addlinespace[2pt]
\bottomrule
\end{tabular}
\end{table}

\begin{table}[!ht]
\centering
\caption{Comparative-advantage functions on \textbf{DeepSeek-Math-RL-7B}. We expand
\Cref{tab:main_results} using the same coefficients (\Cref{app:comparative_rules}). Voting denotes the default
voting-weighted comparative advantage and \textbf{+ Ours} denotes the scaled $\mathsf{PGLA}$ variants
described in \Cref{app:comparative_rules}.}
\label{tab:comparative_rules_deepseek7}
\vskip -3pt\relax
\fontsize{7.8}{8.3}\selectfont
\renewcommand{\arraystretch}{0.99}
\setlength{\tabcolsep}{2pt}
\begin{tabular}{@{\hspace{2pt}}p{\dimexpr\textwidth-324pt\relax}*{6}{>{\centering\arraybackslash}p{40.5pt}}*{2}{>{\centering\arraybackslash}p{22.5pt}}@{\hspace{2pt}}}
\toprule
\multirow{2}{*}{\textbf{Method}} & \multicolumn{2}{c}{\textbf{MATH}} & \multicolumn{2}{c}{\textbf{MathOdyssey}} & \multicolumn{2}{c}{\textbf{AIME}} & \multicolumn{2}{c}{\textbf{Average}} \\
\cmidrule(lr){2-3} \cmidrule(lr){4-5} \cmidrule(lr){6-7} \cmidrule(lr){8-9}
 & Acc. $\uparrow$ & ECE $\downarrow$ & Acc. $\uparrow$ & ECE $\downarrow$ & Acc. $\uparrow$ & ECE $\downarrow$ & Acc. $\uparrow$ & ECE $\downarrow$ \\
\addlinespace[2pt]
\rowcolor{gray!14} \multicolumn{9}{l}{\strut\hspace{1pt}\textbf{DeepSeek-Math-RL-7B}} \\
\addlinespace[1.5pt]
PPL & $42.50_{\pm 0.23}$ & $53.84_{\pm 0.23}$ & $22.34_{\pm 1.00}$ & $73.85_{\pm 0.99}$ & $3.37_{\pm 0.46}$ & $93.38_{\pm 0.46}$ & $22.74$ & $73.69$ \\
\addlinespace[2pt]
SC & $53.33_{\pm 0.08}$ & $\mathbf{6.43}_{\pm 0.11}$ & $36.57_{\pm 0.56}$ & $9.40_{\pm 1.21}$ & $9.42_{\pm 0.23}$ & $12.12_{\pm 0.26}$ & $33.11$ & $9.32$ \\
\rowcolor{gray!5} \hspace{5pt}+ Ours (voting) & $\mathbf{53.52}_{\pm 0.10}$ & $6.86_{\pm 0.09}$ & $37.87_{\pm 0.55}$ & $9.31_{\pm 0.95}$ & $\mathbf{10.02}_{\pm 0.23}$ & $10.88_{\pm 0.25}$ & $\mathbf{33.80}$ & $9.02$ \\
\hspace{5pt}+ Ours (Uniform) & $53.49_{\pm 0.11}$ & $6.57_{\pm 0.09}$ & $37.51_{\pm 0.64}$ & $9.24_{\pm 0.76}$ & $9.99_{\pm 0.22}$ & $\mathbf{10.78}_{\pm 0.20}$ & $33.66$ & $8.87$ \\
\hspace{5pt}+ Ours (Copeland) & $53.30_{\pm 0.16}$ & $7.17_{\pm 0.14}$ & $\mathbf{38.12}_{\pm 0.55}$ & $\mathbf{8.12}_{\pm 0.87}$ & $9.90_{\pm 0.25}$ & $11.20_{\pm 0.23}$ & $33.78$ & $\mathbf{8.83}$ \\
\hspace{5pt}+ Ours (Maximin) & $53.38_{\pm 0.15}$ & $6.56_{\pm 0.15}$ & $37.10_{\pm 0.48}$ & $9.29_{\pm 1.27}$ & $9.26_{\pm 0.25}$ & $11.99_{\pm 0.27}$ & $33.25$ & $9.28$ \\
\addlinespace[2pt]
Weighted SC & $53.35_{\pm 0.09}$ & $\mathbf{6.43}_{\pm 0.10}$ & $36.68_{\pm 0.61}$ & $9.55_{\pm 1.31}$ & $9.43_{\pm 0.31}$ & $12.10_{\pm 0.34}$ & $33.16$ & $9.36$ \\
\rowcolor{gray!5} \hspace{5pt}+ Ours (voting) & $\mathbf{53.53}_{\pm 0.10}$ & $6.84_{\pm 0.09}$ & $37.94_{\pm 0.58}$ & $9.53_{\pm 0.79}$ & $\mathbf{10.00}_{\pm 0.24}$ & $10.84_{\pm 0.25}$ & $\mathbf{33.83}$ & $9.07$ \\
\hspace{5pt}+ Ours (Uniform) & $53.48_{\pm 0.10}$ & $6.59_{\pm 0.07}$ & $37.56_{\pm 0.69}$ & $9.05_{\pm 0.83}$ & $9.98_{\pm 0.21}$ & $\mathbf{10.75}_{\pm 0.17}$ & $33.67$ & $\mathbf{8.80}$ \\
\hspace{5pt}+ Ours (Copeland) & $53.30_{\pm 0.18}$ & $7.17_{\pm 0.17}$ & $\mathbf{38.20}_{\pm 0.60}$ & $\mathbf{8.14}_{\pm 0.92}$ & $9.92_{\pm 0.30}$ & $11.15_{\pm 0.28}$ & $33.81$ & $8.82$ \\
\hspace{5pt}+ Ours (Maximin) & $53.40_{\pm 0.17}$ & $6.54_{\pm 0.18}$ & $37.12_{\pm 0.53}$ & $9.32_{\pm 1.19}$ & $9.31_{\pm 0.30}$ & $11.92_{\pm 0.33}$ & $33.28$ & $9.26$ \\
\addlinespace[2pt]
Adaptive SC & $53.28_{\pm 0.09}$ & $10.01_{\pm 0.14}$ & $\mathbf{36.30}_{\pm 0.58}$ & $13.25_{\pm 1.14}$ & $9.36_{\pm 0.22}$ & $14.88_{\pm 0.41}$ & $32.98$ & $12.71$ \\
\rowcolor{gray!5} \hspace{5pt}+ Ours (voting) & $\mathbf{53.44}_{\pm 0.16}$ & $10.10_{\pm 0.15}$ & $36.27_{\pm 0.80}$ & $\mathbf{12.33}_{\pm 0.87}$ & $\mathbf{9.92}_{\pm 0.23}$ & $13.31_{\pm 0.49}$ & $\mathbf{33.21}$ & $11.91$ \\
\hspace{5pt}+ Ours (Uniform) & $53.41_{\pm 0.15}$ & $9.88_{\pm 0.15}$ & $36.22_{\pm 0.87}$ & $12.47_{\pm 1.62}$ & $9.79_{\pm 0.22}$ & $\mathbf{13.20}_{\pm 0.51}$ & $33.14$ & $\mathbf{11.85}$ \\
\hspace{5pt}+ Ours (Copeland) & $53.09_{\pm 0.17}$ & $10.45_{\pm 0.25}$ & $33.57_{\pm 1.11}$ & $14.89_{\pm 1.22}$ & $9.54_{\pm 0.25}$ & $13.93_{\pm 0.44}$ & $32.07$ & $13.09$ \\
\hspace{5pt}+ Ours (Maximin) & $53.29_{\pm 0.15}$ & $\mathbf{9.85}_{\pm 0.23}$ & $34.58_{\pm 0.86}$ & $13.50_{\pm 0.95}$ & $8.43_{\pm 0.52}$ & $14.93_{\pm 0.66}$ & $32.10$ & $12.76$ \\
\addlinespace[2pt]
RPC & $53.36_{\pm 0.11}$ & $\mathbf{6.47}_{\pm 0.12}$ & $37.12_{\pm 0.61}$ & $9.54_{\pm 1.54}$ & $9.51_{\pm 0.28}$ & $11.94_{\pm 0.32}$ & $33.33$ & $9.32$ \\
\rowcolor{gray!5} \hspace{5pt}+ Ours (voting) & $\mathbf{53.52}_{\pm 0.11}$ & $6.85_{\pm 0.10}$ & $38.10_{\pm 0.61}$ & $8.89_{\pm 0.76}$ & $10.01_{\pm 0.26}$ & $10.44_{\pm 0.29}$ & $\mathbf{33.88}$ & $8.73$ \\
\hspace{5pt}+ Ours (Uniform) & $53.48_{\pm 0.10}$ & $6.60_{\pm 0.09}$ & $\mathbf{38.12}_{\pm 0.65}$ & $8.17_{\pm 0.93}$ & $\mathbf{10.02}_{\pm 0.27}$ & $\mathbf{10.36}_{\pm 0.34}$ & $33.87$ & $\mathbf{8.38}$ \\
\hspace{5pt}+ Ours (Copeland) & $53.31_{\pm 0.18}$ & $7.16_{\pm 0.17}$ & $38.07_{\pm 0.68}$ & $\mathbf{7.62}_{\pm 0.81}$ & $9.99_{\pm 0.36}$ & $10.90_{\pm 0.34}$ & $33.79$ & $8.56$ \\
\hspace{5pt}+ Ours (Maximin) & $53.40_{\pm 0.18}$ & $6.57_{\pm 0.18}$ & $37.48_{\pm 0.73}$ & $9.19_{\pm 1.17}$ & $9.34_{\pm 0.28}$ & $11.85_{\pm 0.33}$ & $33.41$ & $9.20$ \\
\addlinespace[2pt]
\bottomrule
\end{tabular}
\end{table}

\clearpage
\subsection{Computational Overhead}
\label{app:computational_overhead}

\paragraph{Complexity.}
Let $K$ be the number of sampled paths, $U=|\mathcal A|$ the number
of candidate answers, and $n_u=|C_u|$.
Given answer groups and confidence scores, accumulating voting support
requires $O(K+U)$ time. Directly evaluating the pairwise correction in
\Cref{eq:pairwise_flow,eq:scpa_correction} takes $O(K^2)$ time.
The number of unordered cross-answer path pairs is
$\sum_{u<v}n_un_v=(K^2-\sum_u n_u^2)/2\le K(K-1)/2$.
At our experimental budgets $K=32,64,128$, this is at most $496$,
$2{,}016$, and $8{,}128$ scalar comparisons, respectively. The final
score adjustment in \Cref{eq:scpa_score} is computed in closed form
in $O(U)$ time, without iterative optimization.

\paragraph{Impact on inference time.}
Pairwise comparisons reuse the scalar confidence scores of the sampled
paths; they require neither additional reasoning paths nor per-pair
LLM or verifier calls. Even direct evaluation has a reported median
aggregation latency below $0.109$\,ms per query on the real candidate
pools in \Cref{tab:computational_overhead}. For context, the $K=32$
Qwen3-4B/AIME 2024 pools contain approximately $4.77\times10^5$
generated tokens per pool on average. Relative to this GPU-based
autoregressive generation workload, sub-millisecond CPU post-processing
is expected to have a negligible effect on total inference time.
This assessment concerns aggregation given available scores; separate
confidence-model inference is not included, and we do not infer a
measured end-to-end latency percentage from these timings.

\paragraph{Exact acceleration.}
To further reduce aggregation latency as the candidate pool grows,
we implement an exact sorting-and-prefix-sum evaluation of the default
voting-weighted correction. Define
\begin{equation}
    w_i=\frac{s_{a_i}^{\mathcal V}}{n_{a_i}},\qquad
    d_i=\sum_{j:p_j<p_i}w_j-\sum_{j:p_j>p_i}w_j,\qquad
    s_u^{\mathrm{Pair}}=\frac{\sum_{i\in C_u}d_i}
      {n_u\sum_{v\ne u}s_v^{\mathcal V}}.
    \label{eq:pgla_weighted_sort}
\end{equation}
Within-answer signed comparisons cancel because all paths in the same
group have the same comparison weight $w_i$, yielding the exact identity
above. We sort confidence scores once and collect equal scores into
blocks. Prefix and suffix sums of block weights give the total weights
below and above each block, hence every $d_i$; ties contribute zero.
A final group-wise reduction computes the correction and adjusted
scores. A zero denominator gives zero correction, as in
\Cref{app:online_pgla}. This implementation takes
$O(K\log K+U)$ time and $O(K+U)$ working memory, avoiding a dense
pairwise matrix while preserving the original aggregation rule.

\paragraph{Measured implementation cost.}
We compare two different implementations of $\mathsf{PGLA}$ with
identical confidence-weighted voting support on the same device. Direct evaluation constructs a dense path-comparison array;
the accelerated implementation uses \Cref{eq:pgla_weighted_sort}.
We use 300 cached Qwen3-4B/AIME 2024 pools with DeepConf ($K=32$)
and 1,000 InternLM2-Math-Plus-1.8B pools each from MATH ($K=64$)
and MathOdyssey ($K=128$), with inverse-perplexity weights.  At $K=128$, sorting reduces the reported median aggregation time from
$108.3$ to $60.3$\,$\mu$s, a $1.80\times$ speedup
over direct evaluation. For $K=32$ and $64$, the vectorized direct
implementation remains faster because of its lower fixed cost.
Thus sorting improves scalability and the measured latency at the
largest experimental budget, while direct evaluation remains efficient
for smaller pools.

\begin{table}[H]
\centering
\small
\setlength{\tabcolsep}{6pt}
\caption{CPU aggregation latency of two exact $\mathsf{PGLA}$
implementations ($\mu$s per query), measured on real cached candidate
pools. Both use identical inputs and confidence-weighted voting support.}
\label{tab:computational_overhead}
\begin{tabular*}{\textwidth}{@{\extracolsep{\fill}}lrrr@{}}
\toprule
Implementation & $K=32$ & $K=64$ & $K=128$ \\
& AIME 2024 & MATH & MathOdyssey \\
\midrule
Direct pairwise evaluation & 29.7 & 48.9 & 108.3 \\
Exact sorting + prefix sums & 53.4 & 57.4 & 60.3 \\
\bottomrule
\end{tabular*}
\end{table}

\clearpage
\begingroup
\raggedbottom
\subsection{Connection to Rank-Based Borda Aggregation}
\label{app:borda_connection}

Under SC, $\mathsf{PGLA}$ admits an exact rank representation.
Let $n_u=|C_u|$, let $R_i\in[1,K]$ be the ascending confidence rank
of path $i$ (averaged for ties), and define
$\bar R_u=n_u^{-1}\sum_{i\in C_u}R_i$.
Rank-based Borda assigns answer $u$ the score
$T_u^{\mathrm{Borda}}=\sum_{i\in C_u}R_i=n_u\bar R_u$.

\begin{proposition}[Rank representation under SC]
\label{prop:borda_rank_identity}
For the SC base $s_u^{\mathcal V}=n_u$ and at least two observed
answer groups, the correction in \Cref{eq:scpa_correction} satisfies
\begin{equation}
    s_u^{\mathrm{Pair}}
    =\frac{2\bar R_u-K-1}{K-n_u}.
    \label{eq:borda_pair_identity}
\end{equation}
Consequently, SC-$\mathsf{PGLA}$ selects the answer maximizing
$\log n_u+(2\bar R_u-K-1)/(K-n_u)$.
\end{proposition}

\begin{proof}
Under SC, the correction averages all signed cross-group comparisons.
Summing the rank identity
$\sum_j\operatorname{sgn}(p_i-p_j)=2R_i-K-1$
over $i\in C_u$, canceling within-group terms by antisymmetry,
and dividing by $n_u(K-n_u)$ gives \Cref{eq:borda_pair_identity}.
The selection rule follows from \Cref{eq:scpa_score}.
\end{proof}

Proposition~G.1 makes the connection to rank-based voting precise.
Under SC, both Borda and SC-PGLA use the same path-confidence
ordering, but they aggregate it differently. With $\gamma=1$,
Borda sums ranks as $T_u^{(1)}=n_u\bar R_u$, whereas SC-PGLA
selects by
$\log n_u+(2\bar R_u-K-1)/(K-n_u)$.
The latter treats the ordering as a bounded cross-answer
comparative correction while retaining voting support as a
separate term. In particular, an answer $u$ with $n_u<n_v$
can overtake $v$ only if
$s_u^{\mathrm{Pair}}-s_v^{\mathrm{Pair}}>
\log(n_v/n_u)$.
Thus, reversing a larger vote margin requires stronger
comparative evidence. The rank identity in \Cref{eq:borda_pair_identity}. is
specific to SC; \cref{eq:scpa_correction} also combines pairwise comparisons
with confidence-weighted and externally scored voting support.
The KL solution provides the mechanism for this combination,
while the aggregation rule is determined by how the
comparative evidence is constructed and weighted.

To test whether $\mathsf{PGLA}$ can also enhance Borda itself,
we use $s_u^{\mathcal V}=T_u^{\mathrm{Borda}}$ in
\Cref{eq:scpa_correction,eq:scpa_score}, including the opponent weights.
\Cref{tab:borda_qwen,tab:borda_internlm} compare SC and Borda with their
$\mathsf{PGLA}$-adjusted counterparts on the same sampled subsets.

Borda improves over SC on average, showing that confidence ordering
can help correct voting-based selection errors.
More importantly, applying $\mathsf{PGLA}$ directly to Borda raises
macro-average accuracy from $59.17\%$ to $61.21\%$ on Qwen's
AIME24 and HMMT25 comparisons, and from $25.89\%$ to $27.76\%$
across the six InternLM settings.
These paired comparisons support \textbf{explicit answer-level preference modeling}
as a way to further refine rank-based voting on the evaluated settings.

\begin{table}[H]
\centering
\caption{Qwen3 reasoning-mode accuracy on AIME24 and HMMT25 (\%; $K=32$).}
\label{tab:borda_qwen}
\fontsize{7.3}{8.7}\selectfont
\setlength{\tabcolsep}{2.7pt}
\begin{tabular}{@{}l*{5}{c}@{}}
\toprule
& \multicolumn{2}{c}{Qwen3-1.7B} & \multicolumn{2}{c}{Qwen3-4B} & \multirow{2}{*}{\makecell{Avg.\\results}} \\
\cmidrule(lr){2-3}\cmidrule(lr){4-5}
Method & AIME24 & HMMT25 & AIME24 & HMMT25 & \\
\midrule
\multicolumn{6}{l}{\textbf{DeepConf}} \\
SC & $69.50_{\pm 1.98}$ & $27.00_{\pm 1.00}$ & $80.00_{\pm 0.00}$ & $51.50_{\pm 1.57}$ & $57.00$ \\
SC-$\mathsf{PGLA}$  & $\mathbf{75.00}_{\pm 1.67}$ & $30.00_{\pm 2.98}$ & $83.00_{\pm 1.00}$ & $\mathbf{54.00}_{\pm 2.91}$ & $60.50$ \\
Borda & $72.67_{\pm 2.49}$ & $29.00_{\pm 3.35}$ & $82.00_{\pm 1.63}$ & $52.33_{\pm 2.60}$ & $59.00$ \\
Borda+$\mathsf{PGLA}$ & $74.00_{\pm 2.49}$ & $\mathbf{33.17}_{\pm 3.83}$ & $\mathbf{83.67}_{\pm 2.33}$ & $\mathbf{54.00}_{\pm 2.91}$ & $\mathbf{61.21}$ \\
\midrule
\multicolumn{6}{l}{\textbf{Self-Certainty}} \\
SC & $69.50_{\pm 1.98}$ & $27.00_{\pm 1.00}$ & $80.00_{\pm 0.00}$ & $51.50_{\pm 1.57}$ & $57.00$ \\
SC-$\mathsf{PGLA}$ & $\mathbf{74.00}_{\pm 1.33}$ & $30.33_{\pm 2.77}$ & $82.00_{\pm 1.63}$ & $\mathbf{55.67}_{\pm 1.53}$ & $60.50$ \\
Borda & $72.67_{\pm 2.00}$ & $29.00_{\pm 2.60}$ & $80.67_{\pm 1.33}$ & $55.00_{\pm 1.67}$ & $59.33$ \\
Borda+$\mathsf{PGLA}$ & $\mathbf{74.00}_{\pm 2.00}$ & $\mathbf{33.50}_{\pm 4.25}$ & $\mathbf{82.33}_{\pm 2.13}$ & $55.00_{\pm 1.67}$ & $\mathbf{61.21}$ \\
\bottomrule
\end{tabular}
\end{table}

\begin{table}[H]
\centering
\caption{InternLM2-Math-Plus accuracy (\%; PPL ordering), with $K=64$ for MATH and $K=128$ otherwise.}
\label{tab:borda_internlm}
\fontsize{7.3}{8.7}\selectfont
\setlength{\tabcolsep}{2.7pt}
\begin{tabular}{@{}l*{7}{c}@{}}
\toprule
& \multicolumn{3}{c}{InternLM2-Math-Plus-1.8B} & \multicolumn{3}{c}{InternLM2-Math-Plus-7B} & \multirow{2}{*}{\makecell{Avg.\\results}} \\
\cmidrule(lr){2-4}\cmidrule(lr){5-7}
Method & MATH & MathOdyssey & AIME & MATH & MathOdyssey & AIME & \\
\midrule
SC & $36.48_{\pm 0.14}$ & $14.57_{\pm 0.45}$ & $2.66_{\pm 0.20}$ & $50.57_{\pm 0.18}$ & $28.30_{\pm 0.61}$ & $9.40_{\pm 0.21}$ & $23.66$ \\
SC-$\mathsf{PGLA}$  & $39.77_{\pm 0.17}$ & $18.28_{\pm 0.56}$ & $3.37_{\pm 0.17}$ & $\mathbf{53.99}_{\pm 0.18}$ & $35.30_{\pm 0.55}$ & $9.95_{\pm 0.24}$ & $26.77$ \\
Borda & $38.83_{\pm 0.08}$ & $17.12_{\pm 0.62}$ & $3.23_{\pm 0.18}$ & $53.07_{\pm 0.16}$ & $33.14_{\pm 0.59}$ & $9.97_{\pm 0.12}$ & $25.89$ \\
Borda+$\mathsf{PGLA}$ & $\mathbf{40.58}_{\pm 0.21}$ & $\mathbf{21.13}_{\pm 0.57}$ & $\mathbf{3.56}_{\pm 0.20}$ & $53.60_{\pm 0.16}$ & $\mathbf{37.48}_{\pm 0.58}$ & $\mathbf{10.20}_{\pm 0.34}$ & $\mathbf{27.76}$ \\
\bottomrule
\end{tabular}
\end{table}

\par\clearpage
\endgroup

\clearpage

\section{Impact Statement}
\label{app:impact_statement}
$\mathsf{PGLA}$ aims to make LLM reasoning more accurate and efficient by improving how sampled responses are aggregated. The observed gains in mathematical reasoning, knowledge reasoning, and code generation may benefit applications in problem solving and software development, while the online extension can reduce the number of sampled responses needed. These benefits do not eliminate errors or biases in the underlying models or confidence signals. In particular, agreement among sampled responses is not a guarantee of correctness, and stronger capabilities in reasoning and code generation may also facilitate misuse. Deployment should therefore include evaluation in the intended setting, appropriate safeguards, and human oversight for consequential decisions.

\section{AI Use Statement}
\label{app:ai_use_statement}
In this work, we used LLMs to generate the sampled responses evaluated in our test-time scaling experiments, and process reward models to score reasoning steps in the external-reward setting, as described in \Cref{sec:exp_setup}. Generative AI tools assisted with implementation and debugging, checks of mathematical notation and derivations, interpretation and presentation of experimental results, and manuscript drafting, language editing, and \LaTeX{} formatting. AI assistance was also used to create parts of the illustrations, including the voting-game illustration in \Cref{fig:pipeline_society}. The authors reviewed and revised AI-assisted manuscript content and take responsibility for the final text, claims, experimental results, code, and figures.

\end{document}